%% file: main.tex
\documentclass{article}

\PassOptionsToPackage{numbers, sort&compress}{natbib}
\usepackage[main, final]{neurips_2026}

\usepackage[utf8]{inputenc}
\usepackage[T1]{fontenc}
\usepackage{hyperref}
\usepackage{url}
\usepackage{booktabs}
\usepackage{placeins}  %
\usepackage{amsfonts}
\usepackage{amssymb}
\usepackage{amsmath}
\usepackage{nicefrac}
\usepackage{microtype}
\usepackage[table]{xcolor}
\usepackage{makecell}
\usepackage{multirow}
\usepackage{graphicx}
\usepackage{wrapfig}
\usepackage{tablefootnote}
\usepackage{enumitem}

\newcommand{\modelname}{\textbf{ReSMap}}

\title{ReSMap: Recasting Satellite Priors for Robust and Accurate Online HD Map Construction}

\author{%
  Kyungmin Kim$^{1}$\thanks{Equal contribution.} \quad
  Sumin Lee$^{1}$\footnotemark[1] \quad
  Sungoh Jeong$^{1}$ \\
  \textbf{DoHyun Lim$^{2}$} \quad
  \textbf{Seonghyun Park$^{3}$} \quad
  \textbf{Soonmin Hwang$^{4}$}\thanks{Corresponding author.} \\
  $^{1}$Department of Automotive Engineering (Automotive-Computer Convergence) \\
  $^{2}$Department of Computer Science \quad $^{3}$Department of Electrical Engineering \\
  $^{4}$Department of Automotive Engineering \\
  Hanyang University, Seoul, South Korea \\
  {\small\texttt{\{tnrud17, maroona, sungoh0036, ldh991014, dkdldb713, soonminh\}@hanyang.ac.kr}}
}

\begin{document}
\maketitle

\begin{abstract}
Despite recent advances in camera-based online HD map construction, these systems remain susceptible to sensor failures due to their reliance on onboard cameras. 
In contrast, satellite images remain available independently of real-time sensing conditions as they are cached offline. Existing methods utilizing satellite images, however, primarily leverage them under nominal conditions, without explicitly addressing them as redundancy when a sensor degrades. To address this gap, we propose \textbf{ReSMap}, a redundancy-aware camera-satellite fusion framework for robust online HD map construction. Our framework explicitly leverages satellite images as a complementary source to improve robustness against camera failures.
A \textit{Predict-then-Fuse} (PTF) module fuses camera and satellite BEV features by using each branch's own prediction confidence as a per-pixel reliability weight, with no learned router. An \textit{Object-Centric Cooperative Multi-modal Query} (OCMQ) decoder then extends this per-modality separation to the query level: each modality cross-attends only to its own BEV, and per-modality updates are merged through a learned per-token gate.
Our model achieves state-of-the-art performance on nuScenes across all splits and range settings. More importantly, it also sets the new state-of-the-art under camera-failure scenarios, demonstrating that satellite images serve not only as a strong prior but as practical redundancy for robust online HD mapping.
Project page: \url{https://dehsurc.github.io/resmap-page/}
\end{abstract}

\input{section/1_Intro}
\input{section/2_RelatedWork}

\input{section/3_Method}
\input{section/4_Experiments}

\input{section/5_Conclusion}

\begin{ack}
This work was partly supported by Institute of Information \& communications Technology Planning \& Evaluation (IITP) grant funded by the Korea government (MSIT) (No.~RS-2020-II201373, Artificial Intelligence Graduate School Program (Hanyang University)) and the Technology Innovation Program (No.~RS-2025-25448249, E2E Autonomous Driving Reference Data Construction and Core Technology Development) funded by the Ministry of Trade, Industry Resources (MOTIR, Korea). The authors declare no competing interests.
\end{ack}

\bibliographystyle{unsrtnat}
\bibliography{references}

\clearpage
\appendix
\input{section/A_Appendix}

\end{document}

%% file: section/1_Intro.tex
\section{Introduction}
\label{sec:intro}

For safe autonomous driving, an accurate and reliable understanding of the surrounding road geometry and map elements is critical. Online HD map construction aims to predict vectorized road elements around the ego vehicle from real-time camera observations, for regions where an up-to-date HD map is not available. With the advent of query-based vectorized map construction~\cite{maptr}, several methods have been proposed to improve the online HD map construction performance~\cite{vectormapnet, streammapnet, maptracker}. However, relying solely on onboard sensors has two key limitations. 
First, in real-world deployment, they can become partially or completely unavailable due to hardware faults, occlusion, or temporary signal loss.
Second, they provide increasingly sparse and unreliable evidence as distance increases, making distant-region map prediction particularly challenging.

Beyond camera-only designs, several works have explored auxiliary modalities to enhance map prediction. 
Some methods utilize additional LiDAR sensors~\cite{maptrv2,damap,mask2map}, while others propose utilizing pre-existing low-resolution SD maps to understand coarse road topology~\cite{pmapnet,sdtagnet,priordrive}. 
However, LiDAR is still coupled to the onboard sensor stack, while SD maps lack the dense lane-level geometry required for fine-grained HD map prediction. 
Satellite images offer another promising auxiliary modality: they provide top-down visual observations of the surrounding road layout while remaining available independently of real-time sensing conditions through offline caching~\cite{satforhdmap,satmaptr,satmap}.

Nevertheless, existing models using such auxiliary modalities still treat them as supplementary cues for improving accuracy under clean conditions, implicitly relying on the unrealistic assumption that sensor observations remain stable and informative. 
As a result, they neither fully exploit auxiliary modalities as spatial priors under clean conditions nor use them as reliable redundancy when sensor observations are degraded or missing.

In this paper, we recast satellite images as an \emph{offline redundancy modality}. Under normal conditions, they should complement onboard sensing by providing additional spatial evidence where sensor observations are uncertain or incomplete. Under sensor degradation, they must serve as an independent redundancy when onboard observations become unreliable or missing.

Our framework effectively exploits satellite images as both an explicit spatial prior under normal conditions and a redundancy source when sensors degrade. The model consists of two main architectural designs. i) \emph{Predict-then-Fuse (PTF)} fuses camera and satellite BEV features using the prediction confidence estimated from each branch. 
ii) \emph{Object-Centric Cooperative Multi-modal Query (OCMQ) decoder} keeps a single shared object-query pool while computing modality-conditioned updates inside each decoder layer, and merges them through a learned per-token gate. %

Extensive experiments validate the effectiveness of our design. Our model achieves state-of-the-art HD map construction performance across diverse evaluation settings and consistently outperforms existing methods utilizing external modalities under sensor degradation, demonstrating its ability to fully utilize satellite images as both spatial prior and failure-time redundancy.

\paragraph{Contributions.}
Our contributions can be summarized as follows:
\begin{itemize}[leftmargin=1.2em,itemsep=2pt,topsep=2pt]
   \item We formulate geo-aligned satellite images as an offline-cached redundancy modality for onboard camera sensing in online HD map construction.
   \item We introduce two architectural designs: a \emph{Predict-then-Fuse (PTF)} interface for confidence-aware BEV fusion, and an \emph{Object-Centric Cooperative Multi-modal Query (OCMQ)} decoder that preserves complementary map evidence through camera, satellite, and fused query streams.

  \item 
We achieve state-of-the-art mapping performance under clean conditions across all evaluation settings, with the gap over prior work widening as the perception range grows.
  \item
We retain accuracy under camera failure where existing baselines suffer significant drops, by reallocating our model's dependence onto the satellite stream when the camera signal becomes unreliable.
  
\end{itemize}

%% file: section/2_RelatedWork.tex
\section{Related Work}
\label{sec:related}

\paragraph{Online HD map construction.}
Online HD map construction aims to predict vectorized lane dividers, road boundaries, and pedestrian crossings around the ego vehicle directly from onboard cameras. Early dense-BEV-segmentation formulations~\cite{hdmapnet} have been largely superseded by query-based vectorized decoders~\cite{detr,deformabledetr,vectormapnet,maptr,maptrv2,mask2map,mapqr,mapexpert}, with temporal aggregation~\cite{streammapnet,maptracker,memfusionmap,hrmapnet} further improving per-frame consistency. Camera-only systems, however, remain bounded by the cameras' own limits: sparse pixel support at long range and brittle coverage under occlusion. 
Beyond camera-only settings, recent methods have incorporated auxiliary priors from LiDAR~\cite{maptrv2,damap,mask2map}, SD maps~\cite{pmapnet,sdtagnet,priordrive}, and satellite images~\cite{satforhdmap,satmaptr,satmap}. These works demonstrate the benefit of additional spatial cues for HD map construction, but most existing formulations focus on clean conditions rather than exploiting auxiliary priors as redundant evidence under camera failure.

\paragraph{Multi-modal and failure-aware fusion.}
Existing fusion designs in HD map construction and related BEV perception tasks broadly fall into several families: channel concatenation with residual convolution~\cite{satmap,mapfusion}, learned gated refinement~\cite{satmaptr}, masked cross-attention~\cite{satforhdmap,bai2022transfusion}, and learned routers~\cite{mome,chen2023futr3d,sparsefusion}.
SD map-based methods inject topology priors into the decoder via topology-conditioned positional encodings~\cite{pmapnet,sdtagnet,priordrive}.
These designs generally assume that all input modalities are available and reliable at inference time, allowing corrupted or uninformative evidence to leak through residual paths, attention updates, or learned routing weights.
A separate line of failure-aware fusion work addresses reliability through corruption-aware training and reliability-weighted routing, targeting camera-LiDAR or camera-radar failures at the BEV level~\cite{metabev,unibev,park2025rocars} and incomplete-observation recovery within vectorized HD map construction~\cite{safemap,robomap,sefmap}.
Related prediction-aware dense fusion methods~\cite{cao2024pdf,zhuang2021pmf,ye2024sgbev} similarly weight branches by per-branch prediction confidence.
Yet these reliability cues remain confined to fusion and do not reach the decoder.
Existing multi-modal decoder designs fuse modalities before decoding~\cite{unibev,ye2024sgbev}, mix them sequentially within the decoder~\cite{yang2022deepinteraction,palladin2024samfusion}, or use parallel cross-attention followed by simple addition~\cite{chen2023futr3d,wang2025mv2dfusion}, so corruption that survives imperfect BEV-level fusion propagates through the decoder.
In contrast, our approach accounts for modality reliability in both fusion and decoding:
PTF weights camera and satellite BEV features by each branch's own prediction confidence
through a parameter-free per-pixel gate, rather than learning to mix jointly processed
features, and OCMQ extends per-modality isolation into the query decoder.

%% file: section/3_Method.tex
\section{Method}
\label{sec:method}

\begin{figure}[t]
  \centering
  \includegraphics[width=\linewidth]{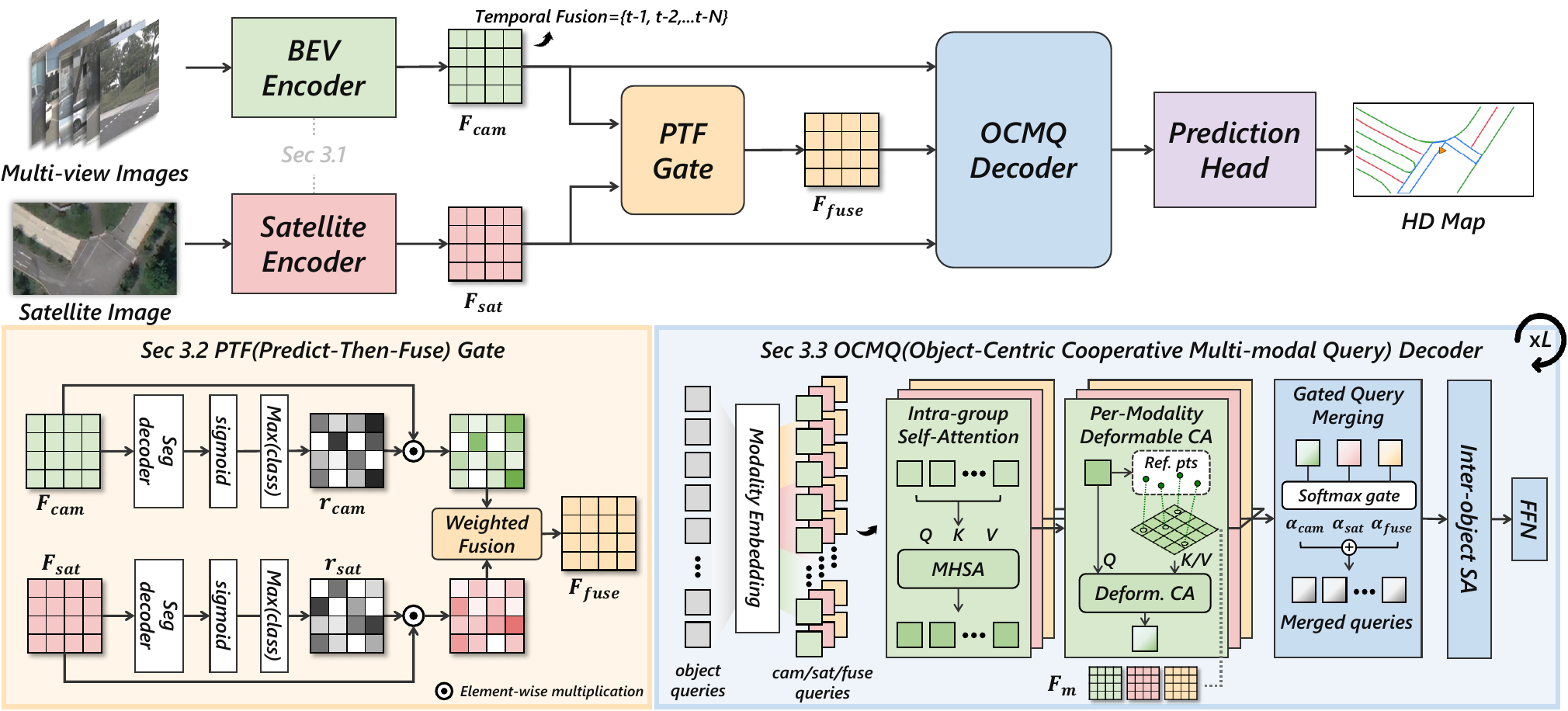}
  \caption{%
  Overview of the model architecture of \modelname{}. Two modality-specific encoders produce camera
  and satellite BEV features, each with a BEV segmentation head that
  outputs per-pixel confidence. The two BEV streams are fused by a
  parameter-free, confidence-weighted gate (PTF). A single object-query set
  then gathers modality-specific updates from the camera, satellite, and
  fused BEV streams, which are recomposed by a learned per-token gate
  (OCMQ) and decoded into the final vector HD map.%
  }
  \label{fig:overview}
\end{figure}

Given a sequence of multi-view camera images and a satellite image centered at the ego vehicle location~\cite{aid4ad}, the model outputs a set of vectorized map instances
\begin{equation}
  \mathcal{M} = \{ (c_n, \mathbf{p}_n) \}_{n=1}^{N},\qquad
  c_n \in \mathcal{C},\quad
  \mathbf{p}_n \in \mathbb{R}^{P \times 2},
  \label{eq:output}
\end{equation}
in the ego-vehicle BEV frame, where $\mathbf{p}_n$ is a polyline of $P$ sampled points and $\mathcal{C}$ covers pedestrian crossings, lane dividers, and road boundaries. Our model has two parts (Fig.~\ref{fig:overview}): a BEV-level fusion module (Sec.~\ref{sec:sc_fusion}) and an instance-level decoder (Sec.~\ref{sec:mm_decoder}). 

\subsection{Modality-Specific BEV Encoders}
\label{sec:bev_encoders}
We use two independent encoders that produce BEV features.
For the camera branch, multi-view images are encoded with BEVFormer~\cite{bevformer}, producing $\mathbf{F}_{\mathrm{cam}} \in \mathbb{R}^{C \times H \times W}$.
We consider two variants that differ only in temporal aggregation: a \emph{non-temporal} variant that encodes each frame independently, and a \emph{temporal} variant that follows MapTracker~\cite{maptracker} by maintaining a short BEV memory queue, warping past BEVs into the current ego-frame, and fusing them with the current BEV through a lightweight memory-fusion block inside the encoder.
The satellite branch uses a ResNet backbone with an FPN-style aggregation head~\cite{fpn} to encode the ego-centered satellite image. The resulting feature map is resized and aligned to the BEV feature grid, producing $\mathbf{F}_{\mathrm{sat}} \in \mathbb{R}^{C \times H \times W}$.
Each branch is paired with a BEV segmentation decoder, providing auxiliary supervision and confidence cues for the fusion gate in Sec.~\ref{sec:sc_fusion}.

\subsection{Predict-then-Fuse with a Self-Confidence Gate}
\label{sec:sc_fusion}

Existing camera-satellite BEV fusion methods~\cite{satmap, satmaptr, satforhdmap} exploit complementary information under clean inputs, but their redundancy under camera failure is implicit: convolutional fusion~\cite{satmap} processes clean and corrupted features with the same learned filters, and cross-attention~\cite{satforhdmap} can transmit corrupted queries through residual paths. Neither explicitly controls how much each modality contributes at each BEV location. This location-wise control matters for robust online HD map construction, since the more reliable modality often varies across segments of a single polyline under occlusion or sensor failure.

\input{figures/qual_fallback_scene0392_v3/qual_fallback_scene0392_v3}

In Predict-then-Fuse (PTF), per-branch BEV segmentation decoders produce modality-specific logits from $\mathbf{F}_{\mathrm{cam}}$ and $\mathbf{F}_{\mathrm{sat}}$, whose maximum class-wise sigmoid confidence we use directly as a per-pixel modality weight without any learnable parameters. The reliability map $r_m$, scalar per pixel and shared across channels, is then defined as
\begin{equation}
  r_m(x) \;=\;
  \max_{c}\;
  \bigl[\sigma\bigl(\mathrm{Seg}_{m}(\mathbf{F}_m)\bigr)\bigr]_{x,c},
  \qquad
  r_m \in [0,1]^{H \times W},\quad
  m \in \{\mathrm{cam},\, \mathrm{sat}\},
  \label{eq:reliability}
\end{equation}
where $\sigma(\cdot)$ denotes the sigmoid, $c$ indexes the segmentation classes, and $x$ indexes spatial locations on the BEV grid.
We stop-gradient $r_m$ before plugging it into the fusion, so it serves as a fixed input that the fusion combines without learning any modality-comparison weights. The fallback $r_{\mathrm{cam}} \to 0 \Rightarrow \mathbf{F}_{\mathrm{fuse}} \to \mathbf{F}_{\mathrm{sat}}$ is then a closed-form property of Eq.~\eqref{eq:fusion}, and $r_m$ stays a reliable map-evidence signal under raw-image failures not seen during training (e.g., zeroed cameras at inference). \emph{Parameter-free} here refers to the fusion rule alone: the segmentation decoders $\mathrm{Seg}_m$ that produce $r_m$ are trained with the BEV segmentation loss (Sec.~\ref{sec:training}), but no learnable weight decides how the two modalities are mixed.
Since each branch's reliability map is the gating signal it derives from its own prediction, we refer to the resulting fusion operator as the \emph{Self-Confidence Gate}, which combines the two modalities as a confidence-weighted average,
\begin{equation}
  \mathbf{F}_{\mathrm{fuse}}(x)
  \;=\;
  \frac{r_{\mathrm{cam}}(x)\, \mathbf{F}_{\mathrm{cam}}(x) \;+\; r_{\mathrm{sat}}(x)\, \mathbf{F}_{\mathrm{sat}}(x)}
       {r_{\mathrm{cam}}(x) \;+\; r_{\mathrm{sat}}(x) \;+\; \epsilon},
  \label{eq:fusion}
\end{equation}
where $\epsilon$ is a small constant for numerical stability.
The fused BEV features are then passed to a dedicated fused segmentation head $\mathrm{Seg}_{\mathrm{fuse}}$.
The Self-Confidence Gate thus treats redundancy as soft evidence reallocation: when camera failures lower the segmentation confidence on map-relevant regions, the camera weight in Eq.~\eqref{eq:fusion} drops and the satellite branch contributes more to the fused feature {(Fig.~\ref{fig:fallback_qualitative_v3}(a))}.

\subsection{Object-Centric Cooperative Multi-modal Query Decoder}
\label{sec:mm_decoder}

\begin{figure*}[t]
\centering
\includegraphics[width=\textwidth]{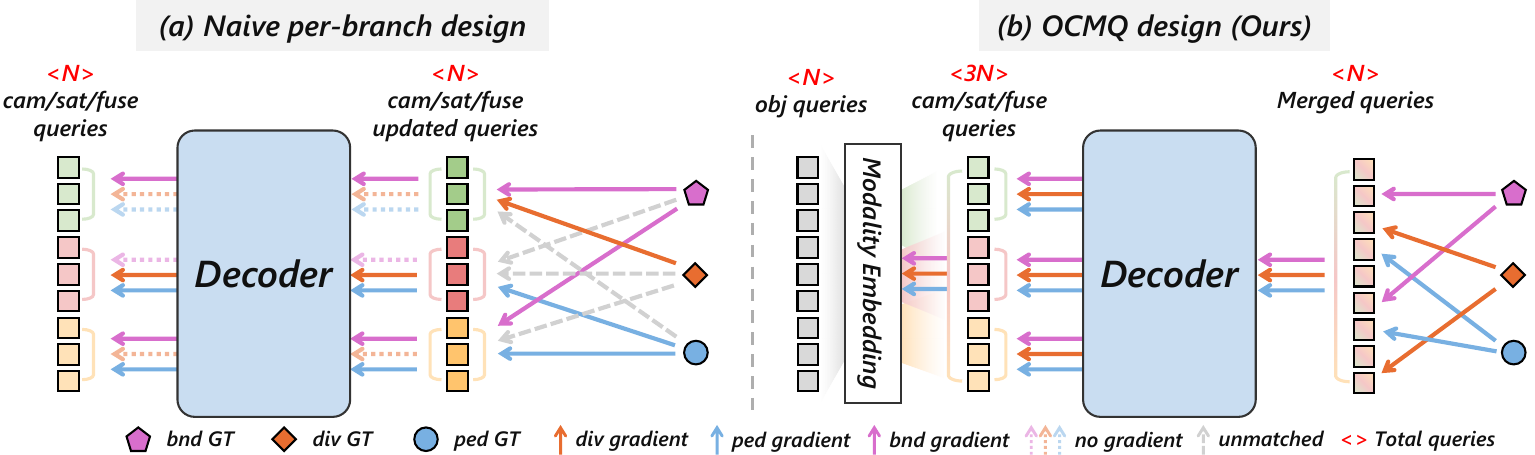}
\caption{Effect of query design on Hungarian matching. \textbf{(a)} With independent cam/sat/fuse query pools, bipartite matching routes each GT to the strongest modality only, causing class--modality entanglement (e.g.\ dividers and boundaries learned only from camera). \textbf{(b)} OCMQ uses a single object-query pool with modality-specific updates merged via a per-token gate, so each query absorbs cam, sat, and fuse gradients jointly and every class receives signal from all modalities.}
\label{fig:query}
\end{figure*}

A standard DETR-based map decoder~\cite{maptr} undermines PTF's explicit redundancy: all queries read from a shared key/value pool that mixes the cam, sat, and fused BEVs, so corrupted camera evidence can propagate to any query through cross-attention. A naive remedy of assigning each query to a single modality removes this direct contamination, but then Hungarian matching biases each per-modality query group toward the classes its modality predicts best (e.g.\ AP$_d$/AP$_b$ to camera, AP$_p$ to fusion in our preliminary study, where the subscripts $p$, $d$, and $b$ denote pedestrian crossings, lane dividers, and road boundaries; see Section~\ref{app:branch}). As a result, the unfailed modality cannot recover the classes the failed one used to dominate.
To resolve this entanglement (Fig.~\ref{fig:query}), we keep a single set of $N$ object queries for set prediction and create modality-specific updates only within each decoder layer. The Hungarian matcher therefore sees only this single unified pool, even though each layer internally splits the queries into modality-conditioned updates. The camera, satellite, and fused updates are computed from separate BEV features ($\mathbf{F}_{\mathrm{cam}}$, $\mathbf{F}_{\mathrm{sat}}$, $\mathbf{F}_{\mathrm{fuse}}$) and then recombined into a unified query set, in four steps. Following MapTracker~\cite{maptracker}, the decoder stacks $L$ such layers with reference points refined at each layer.

\textbf{(1) Modality Embedding and Intra-group Self-Attention.}
We replicate the object query $\mathbf{Q} \in \mathbb{R}^{N \times D}$ three times and add a learnable modality embedding $\mathbf{e}_m \in \mathbb{R}^{D}$ to each copy, producing three modality-conditioned views of the same query set. We then apply a separate self-attention block $\mathrm{IntraSA}_m$ to each view:
\begin{equation}
\begin{aligned}
  \mathbf{Q}_m &= \mathbf{Q} + \mathbf{e}_m,
  \qquad m \in \{\mathrm{cam}, \mathrm{sat}, \mathrm{fuse}\}, \\
  \tilde{\mathbf{Q}}_m
  &= \mathrm{LN}\!\left(
      \mathbf{Q}_m + \mathrm{IntraSA}_m(\mathbf{Q}_m)
    \right).
\end{aligned}
\label{eq:intra_sa}
\end{equation}

\textbf{(2) Per-Modality Deformable Cross-Attention.}
Each modality-conditioned query set $\tilde{\mathbf{Q}}_m$ then attends, via deformable cross-attention~\cite{deformabledetr} with per-modality reference points $\mathbf{R}_m \in [0,1]^{N \times P \times 2}$, only to the BEV tensor of the same modality:
\begin{equation}
  \mathbf{Q}^{u}_{m}
  =
  \mathrm{LN}\!\left(
  \tilde{\mathbf{Q}}_m
  +
  \mathrm{DeformCA}_{m}(\tilde{\mathbf{Q}}_m, \mathbf{R}_m; \mathbf{F}_m)
  \right),
  \qquad
  m \in \{\mathrm{cam}, \mathrm{sat}, \mathrm{fuse}\}.
  \label{eq:modality_ca}
\end{equation}
Per-modality keys/values keep BEV-level corruption from leaking across $\mathbf{Q}^{u}_m$ within this layer, while $\mathbf{F}_{\mathrm{fuse}}$ from Sec.~\ref{sec:sc_fusion} lets $\mathbf{Q}^{u}_{\mathrm{fuse}}$ inherit the calibrated fusion without re-deriving it inside the decoder.

\textbf{(3) Gated Query Merging.}
The three updated queries are merged back into a single representation, the \emph{merged query} $\mathbf{Q}'$, through a per-token modality gate:
\begin{equation}
  \alpha_m
  =
  \mathrm{softmax}_{m}\!\left(W\mathbf{Q}^{u}_{m}\right),
  \qquad
  \mathbf{Q}'
  =
  \mathrm{LN}\!\left(
  \sum_{m}
  \alpha_m \mathbf{Q}^{u}_{m}
  \right).
  \label{eq:gated_collapse}
\end{equation}
Here, $W \in \mathbb{R}^{1 \times D}$ scores each modality-specific update and $\alpha_m$ weights its contribution per token. We omit a residual to the previous-layer query $\mathbf{Q}$: adding it would let down-weighted modality information re-enter through the residual path and accumulate over layers, so $\mathbf{Q}'$ depends on a corrupted modality only through its own $\alpha_m$. {Fig.~\ref{fig:fallback_qualitative_v3}(b) shows this query-level routing.}

\textbf{(4) Inter-Object Self-Attention and Refinement.}
The merged queries finally exchange information across objects through a standard self-attention block followed by an FFN:
\begin{equation}
  \mathbf{Q}'' \;=\; \mathrm{LN}\!\bigl(\mathbf{Q}' + \mathrm{InterSA}(\mathbf{Q}')\bigr), \qquad
  \mathbf{Q}_{\mathrm{out}} \;=\; \mathrm{LN}\!\bigl(\mathrm{FFN}(\mathbf{Q}'') + \mathbf{Q}''\bigr).
  \label{eq:post_sa}
\end{equation}
A classification head and a regression head are then applied to $\mathbf{Q}_{\mathrm{out}}$ to produce the per-instance prediction in Eq.~\eqref{eq:output},
\begin{equation}
  \hat{c}_n \;=\; \mathrm{Cls}(\mathbf{Q}_{\mathrm{out}})_n, \qquad
  \hat{\mathbf{p}}_n \;=\; \mathrm{Reg}(\mathbf{Q}_{\mathrm{out}})_n,
  \label{eq:pred_heads}
\end{equation}
where $\mathrm{Cls}$ outputs class logits over $\mathcal{C}$ and $\mathrm{Reg}$ outputs the $P$ polyline points. Steps (1)--(4) constitute one of the $L$ stacked decoder layers, and following standard DETR-based map decoders~\cite{maptr}, both heads are applied at every layer for auxiliary Hungarian-matched supervision.

\subsection{Training Objective}
\label{sec:training}

Our model has two training objectives, per-branch BEV segmentation and set-prediction vector decoding, combined into a single loss
\begin{equation}
  \mathcal{L}
  \;=\;
  \underbrace{\vphantom{\sum_{\ell=1}^{L}}\sum_{m \in \{\mathrm{cam},\mathrm{sat},\mathrm{fuse}\}} \lambda_m^{\mathrm{seg}}\, \mathcal{L}_m^{\mathrm{seg}}}_{\text{BEV segmentation}}
  \;+\;
  \underbrace{\vphantom{\sum_{m \in \{\mathrm{cam},\mathrm{sat},\mathrm{fuse}\}}}\sum_{\ell=1}^{L}\!\bigl(\lambda^{\mathrm{cls}} \mathcal{L}^{\mathrm{cls}}_{\ell} + \lambda^{\mathrm{reg}} \mathcal{L}^{\mathrm{reg}}_{\ell}\bigr)}_{\text{Vector set prediction}},
  \label{eq:total_loss}
\end{equation}
where each $\mathcal{L}_m^{\mathrm{seg}}$ is a sum of focal and dice losses~\cite{focal_loss,dice_loss} applied to BEV segmentation decoder $\mathrm{Seg}_m$, and $\mathcal{L}^{\mathrm{cls}}_{\ell}$ and $\mathcal{L}^{\mathrm{reg}}_{\ell}$ are the focal and smooth-L1 losses applied at each decoder layer $\ell$ to the Hungarian-matched~\cite{hungarian} predictions of the OCMQ decoder.

%% file: figures/qual_fallback_scene0392_v3/qual_fallback_scene0392_v3.tex
\begin{figure*}[t]
\centering
\newlength{\fbgateW}
\setlength{\fbgateW}{0.27\textwidth}        %
\newlength{\fbcellh}
\setlength{\fbcellh}{0.5\fbgateW}            %
\newlength{\fbcbarW}
\setlength{\fbcbarW}{0.035\textwidth}        %
\newlength{\fbpanelH}
\setlength{\fbpanelH}{\dimexpr 2\fbcellh+5pt\relax}  %
\newlength{\fbpanelW}
\setlength{\fbpanelW}{3.0\fbpanelH}         %

\newcommand{\fbrowlbl}[2]{%
  \parbox[c][\fbcellh][c]{0.5em}{\centering\rotatebox{90}{%
    \footnotesize\textsf{\shortstack[c]{#1\\[1pt]#2}}}}}
\newcommand{\fbgatecell}[1]{%
  \parbox[c][\fbcellh][c]{\fbgateW}{\centering%
    \includegraphics[width=\fbgateW]{#1}}}

\setlength{\tabcolsep}{2pt}
\renewcommand{\arraystretch}{1.0}
\begin{tabular}{c@{\hskip 4pt}cc@{\hskip 2pt}c@{\hskip 4pt}c}
 & {\footnotesize\textsf{Clean}}
 & {\footnotesize\textsf{All-camera drop}}
 &
 & {\footnotesize\textsf{Per-query camera gate}} \\
\fbrowlbl{$\bar r_{\mathrm{cam}}$}{} &
\fbgatecell{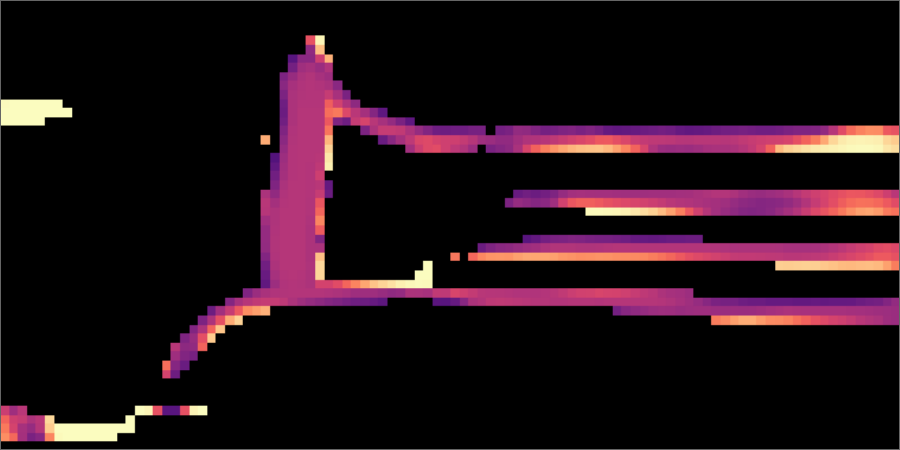} &
\fbgatecell{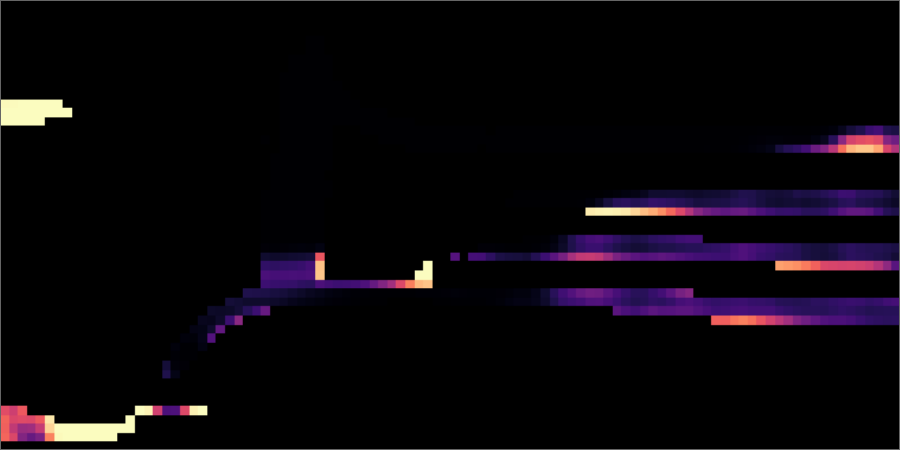} &
\multirow{2}{*}[10pt]{%
  \includegraphics[height=\fbpanelH,width=\fbcbarW,keepaspectratio]%
    {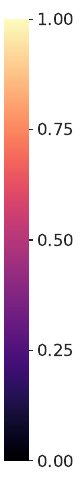}} &
\multirow{2}{*}[20pt]{%
  \begin{tabular}{@{}c@{\hskip -10pt}c@{}}
    \includegraphics[height=\fbpanelH,width=\fbpanelW,keepaspectratio]%
      {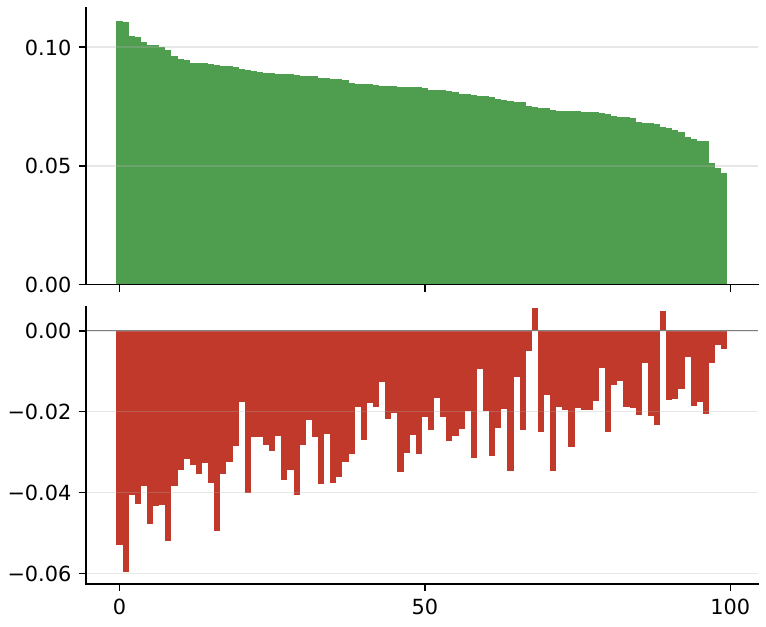} &
    \raisebox{55pt}{
    \begin{tabular}{@{}c@{}}
      \parbox[c][0.5\fbpanelH][c]{1.6em}{\centering%
        \rotatebox{270}{\footnotesize$\alpha_{\mathrm{cam}}$}}\\
      \parbox[c][0.5\fbpanelH][c]{1.6em}{\centering%
        \rotatebox{270}{\footnotesize$\Delta\alpha_{\mathrm{cam}}$}}
    \end{tabular}} 
  \end{tabular}} \\[25pt]
\fbrowlbl{$\bar r_{\mathrm{sat}}$}{} &
\fbgatecell{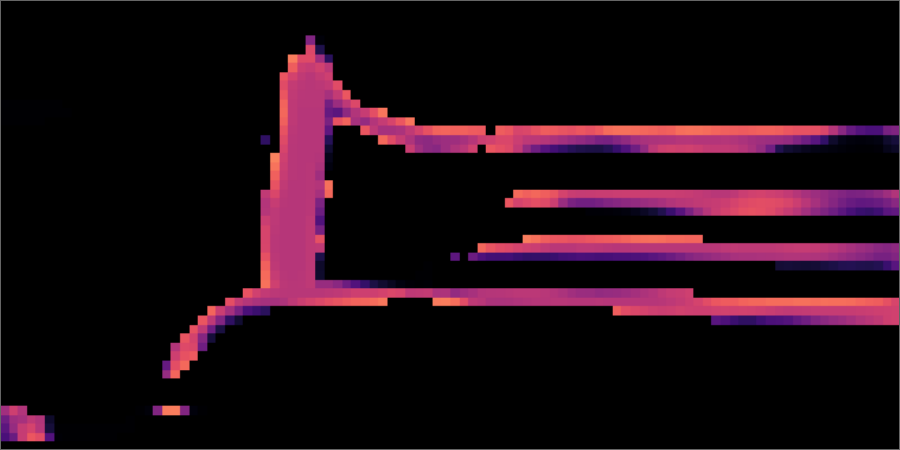} &
\fbgatecell{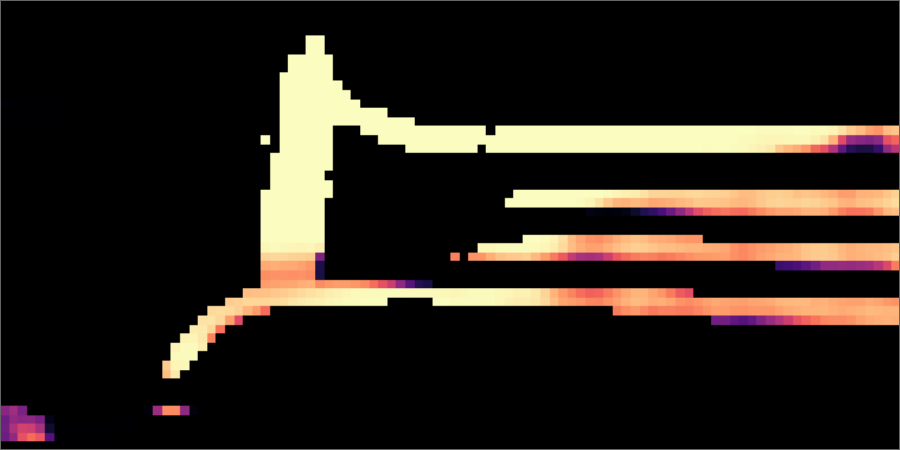} &
 & \\[2pt]
 & \multicolumn{2}{c}{\footnotesize\textsf{(a) Per-pixel BEV gate}}
 &
 & {\footnotesize\textsf{(b) Per-query gate}} \\
\end{tabular}

\vspace{3pt}
\caption{%
\textbf{Per-pixel and per-query gate response under camera dropout.}
\textbf{(a)} Camera and satellite reliability normalized as in Eq.~\eqref{eq:fusion},
$\bar r_m = r_m/(r_{\mathrm{cam}}+r_{\mathrm{sat}})$, over active cells
$\mathcal{A}=\{x:r_{\mathrm{cam}}^{\mathrm{clean}}(x)\ge 0.35\}$. Removing all cameras collapses $\bar r_{\mathrm{cam}}$ while
$\bar r_{\mathrm{sat}}$ rises to cover the same regions.
\textbf{(b)} The same routing is visible at the query level: sorting
queries by $\alpha_{\mathrm{cam}}^{\mathrm{clean}}$ (top) and plotting
the matched change $\Delta\alpha_{\mathrm{cam}}$ under all-camera drop
(bottom) shows a negative shift for $98\%$ of queries, with the largest
drops on those most camera-reliant in the clean run.}
\label{fig:fallback_qualitative_v3}
\end{figure*}

%% file: section/4_Experiments.tex
\section{Experiments}
\label{sec:experiments}

\input{tables/mapping_performance}

\input{tables/robustness_performance_new}
\subsection{Experimental Setup}
\label{sec:setup}

\paragraph{Implementation Details.} We train our model with the three-stage curriculum described in Section~\ref{app:impl} (BEV segmentation, vector decoding, and joint optimization), using AdamW on $4$ NVIDIA RTX 4090 GPUs and a ResNet-50 image backbone. Full training schedules, learning rates, and per-loss weights are deferred to Section~\ref{app:impl}.

\paragraph{Dataset.} We trained and evaluated our model on nuScenes~\cite{nuscenes} using the geographically non-overlapping split~\cite{streammapnet} as the main benchmark. We use ego-aligned AID4AD~\cite{aid4ad} satellite images, at up to $0.15$\,m per pixel, cropped to the corresponding BEV region of interest (ROI). Results on Argoverse~2~\cite{argoverse2} with satellite tiles from Google Maps are reported in Section~\ref{app:av2}.

\paragraph{Metrics.} We evaluate two ROIs, $60\,\mathrm{m}\times30\,\mathrm{m}$ and $100\,\mathrm{m}\times50\,\mathrm{m}$. We report average precision (AP) for pedestrian crossings (AP$_p$), lane dividers (AP$_d$), and road boundaries (AP$_b$), averaged over Chamfer thresholds $\{0.5, 1.0, 1.5\}\,$m and $\{1.0, 1.5, 2.0\}\,$m for the $60\,\mathrm{m}\times30\,\mathrm{m}$ and $100\,\mathrm{m}\times50\,\mathrm{m}$ ROIs, respectively. Robustness is measured by dropping the front camera, the three front cameras, and all six cameras at inference time, with each dropped view replaced by a blackout (all-zero) image.

\paragraph{Baselines.} For each modality category in Tab.~\ref{tab:mapping_performance}, we include representative baselines: camera-only MapTracker~\cite{maptracker}, camera-LiDAR MapTRv2~\cite{maptrv2} and DAMap~\cite{damap}, camera-SD map SDTagNet~\cite{sdtagnet}, and camera-satellite SatforHDMap~\cite{satforhdmap}. To match our input modality, SatforHDMap is retrained with the same AID4AD~\cite{aid4ad} satellite images, and SDTagNet uses OpenStreetMap~\cite{OpenStreetMap} as its SD map source. All baselines except those marked $^{*}$ in Tab.~\ref{tab:mapping_performance} are retrained under the same evaluation conditions, with each model's training schedule following its original paper. None of the retrained baselines applies modality dropout or feature-noise augmentation in its released training configuration, and we retrain them as released; our model is the only one trained with BEV-feature dropout (Section~\ref{app:impl}). Section~\ref{app:aug} retrains SatforHDMap, SDTagNet, and DAMap with the same augmentation and our model without it.

\subsection{Quantitative Results}
\label{sec:quantitative}
\paragraph{Performance under Clean Conditions.}

Tab.~\ref{tab:mapping_performance} reports online HD map construction performance under clean inputs.
At $60\,\mathrm{m}\times30\,\mathrm{m}$ (top block), our model reaches \textbf{54.9} mAP with temporal modeling and $47.3$ mAP without, the highest across all listed methods despite being trained for only $24$ epochs, surpassing camera-only (MapTracker~\cite{maptracker}, $+14.6$ mAP), LiDAR-fused (DAMap~\cite{damap}, $+15.5$ mAP), and SD+HD-map-fused (PriorDrive~\cite{priordrive}, $+12.4$ mAP) baselines.
At $100\,\mathrm{m}\times50\,\mathrm{m}$ (bottom block), where camera evidence becomes sparser and less reliable at distance, the advantage becomes more pronounced: our model reaches \textbf{54.7} mAP, again the best across all methods, and its temporal variant attains the highest AP for all three map classes.
The larger performance gap over camera-only (MapTracker, $+18.5$ mAP), LiDAR-fused (DAMap, $+24.7$ mAP), and SD-fused (SDTagNet~\cite{sdtagnet}, $+32.0$ mAP) baselines suggests that satellite images are especially useful when the BEV region extends beyond dense onboard visual support. Despite the additional satellite branch, our model achieves competitive FPS --- comparable to camera-only baselines and faster than LiDAR/SD-fused models (Tab.~\ref{tab:mapping_performance}).

We additionally evaluate our model on the original split of nuScenes in Section~\ref{app:oldsplit}, where it also achieves state-of-the-art HD map construction performance.

\input{figures/qual_scene0053/qual_scene0053}

\paragraph{Satellite Image as Redundancy.}
Tab.~\ref{tab:robustness_performance_new} reports robustness under camera dropout. Across Front-1, Front-3, and All-camera drop, our non-temporal model degrades by only \textbf{-6.6\%}, \textbf{-9.9\%}, and \textbf{-15.2\%} from clean inputs, the smallest at every drop level. The most severe case isolates the redundancy effect: under All-camera drop, our model retains \textbf{40.1} mAP, on par with the clean-input mAP of the camera-only MapTracker ($40.3$) and above that of every other baseline in Tab.~\ref{tab:robustness_performance_new}, while the camera-only MapTracker~\cite{maptracker} collapses by $-99.8\%$, the LiDAR-fused DAMap~\cite{damap} by $-25.9\%$, the SD-fused SDTagNet~\cite{sdtagnet} by $-84.5\%$, and the prior camera-satellite SatforHDMap~\cite{satforhdmap} by $-47.0\%$. The non-temporal variant outperforms its temporal counterpart under severe dropout because temporal modeling is applied only to the camera BEV features, so corrupted camera evidence propagates through the BEV history. We further analyze this effect in Section~\ref{app:temporal_robustness}.

We further evaluate our model under the MapBench~\cite{mapbench} sensor-corruption protocol on both splits of nuScenes in Section~\ref{app:mapbench}, where it also achieves state-of-the-art robustness.

\subsection{Qualitative Results}
\label{sec:qualitative}
Given multi-view images at a given timestep and the corresponding ego-aligned satellite image as input, Fig.~\ref{fig:qual} qualitatively compares vectorized HD map predictions at an intersection.
At $60\,\mathrm{m}\times30\,\mathrm{m}$, SDTagNet adds spurious crossings, SatforHDMap breaks the crossings into fragments, and MapTracker misses the lane dividers on the far side of the intersection. Our model recovers all three pedestrian crossings, the traffic island ahead of the ego vehicle, and the far-side dividers.
The gap widens under Front-3 drop. MapTracker loses every map element in front of the ego vehicle, SDTagNet produces cluttered boundaries and spurious crossings, and SatforHDMap recovers only fragments of the crossings. In contrast, our model keeps the intersection nearly unchanged by using satellite images as a redundancy cue for missing camera observations.
At $100\,\mathrm{m}\times50\,\mathrm{m}$, MapTracker and SDTagNet miss most of the distant crossings, and SatforHDMap recovers them only partially. Our model preserves the crossing layout at distance and remains more consistent with the ground truth. Additional qualitative comparisons are provided in Section~\ref{app:qual}.

\subsection{Ablation Study}
\paragraph{Evaluating BEV Fusion Method.}
Tab.~\ref{tab:ocmq_ablation}(a) compares PTF against two published camera-satellite BEV fusion designs, Cross-Attention~\cite{satforhdmap} and ConvFuser~\cite{satmap}, under our common training setup, with only the fusion module swapped. We also compare with CIT+DDF~\cite{mapfusion}, a fusion module for online mapping originally proposed for camera-LiDAR fusion, by replacing the LiDAR encoder with our satellite encoder.
Cross-Attention collapses to $0.9$ mIoU under All-camera drop ($-98.0\%$) because residual paths transmit corrupted camera queries directly to the decoder. ConvFuser and CIT+DDF reach competitive clean mIoU ($52.6$ and $52.3$) but still degrade by $-31.7\%$ and $-41.9\%$, since neither explicitly suppresses the unreliable modality. PTF achieves the highest clean mIoU (\textbf{53.5}) and the smallest drops ($-4.9\%$ Front-1, $-24.7\%$ All-camera): the per-pixel reliability gate, derived from each branch's own segmentation head, naturally shifts mass to the satellite branch when camera evidence weakens.

\paragraph{Evaluating Decoder Design.}
Tab.~\ref{tab:ocmq_ablation}(b) ablates OCMQ by additively turning on each component of Sec.~\ref{sec:mm_decoder}, with PTF held fixed so that $\mathbf{F}_{\mathrm{cam}}, \mathbf{F}_{\mathrm{sat}}, \mathbf{F}_{\mathrm{fuse}}$ are identical across rows.

The baseline (row~1) uses a single shared query pool that sees the three BEVs jointly as one token sequence; corrupted camera evidence reaches every query, giving $52.8$ clean mAP and large drops ($-12.5\%$ / $-51.3\%$). Adding \emph{per-modality deformable CA} (Eq.~\eqref{eq:modality_ca}) restores BEV-level isolation at the keys/values, lifting clean to $53.6$ and Front-1 to $-9.3\%$, but All-camera barely moves ($-51.3\%$). Further adding \emph{modality embedding + IntraSA} (Eq.~\eqref{eq:intra_sa}) conditions queries per modality, improving All-camera to $-48.5\%$, yet clean ($53.2$) and Front-1 ($-9.4\%$) regress: in both rows the uniform mean still pulls clean tokens toward the weakest modality at the merge step.

Replacing the mean with the learned \emph{gated query merging} (Eq.~\eqref{eq:gated_collapse}) completes the OCMQ block and is best in every column: \textbf{54.9} clean, \textbf{50.3} ($-8.4\%$) Front-1, \textbf{32.9} ($-40.1\%$) All-camera.
Gated merging allows the decoder to rely less on the camera-conditioned query when the camera input is corrupted. In contrast, mean merging averages all modality queries with fixed weights regardless of corruption, allowing corrupted camera information to affect the final query.

\input{tables/ocmq_ablation}

%% file: tables/mapping_performance.tex
\begin{table*}[t]
\centering
\normalsize
\caption{Online HD map construction performance on the geographically non-overlapping split of nuScenes~\cite{streammapnet}; FPS measured on a single RTX 4090.
\emph{Mod.}: C, L, SD, HD, SA denote Camera, LiDAR, SD map, prior HD map, and Satellite images;
\emph{Temp.}: temporal memory or historical-prediction priors.
$^{*}$~from the original paper;
$^{\dagger}$~Consistent-GT supervision~\cite{maptracker}.}
\label{tab:mapping_performance}
\setlength{\tabcolsep}{3.5pt}
\renewcommand{\arraystretch}{1.0}
\begin{tabular}{c l c c c c c c c c c}
\toprule
Range & Method & Venue & Mod. & Temp. & Ep. & AP$_p$ & AP$_d$ & AP$_b$ & mAP & FPS \\
\midrule

\multirow{8}{*}{\raisebox{-0.8em}{\rotatebox[origin=c]{90}{\textbf{60 m × 30 m}}}}
& MapTracker$^{\dagger}$~\cite{maptracker}   & ECCV'24     & C       & \checkmark  & 58  & 45.9 & 30.0 & 45.1 & 40.3 & 13.5 \\

\cmidrule(lr){2-11}
& MapTRv2~\cite{maptrv2}                     & IJCV'24     & C+L     & $\times$    & 110 & 18.8 & 29.5 & 61.7 & 36.6 & 7.3 \\
& DAMap~\cite{damap}                         & ICCV'25     & C+L     & $\times$    & 110 & 19.2 & 32.8 & \textbf{66.1} & 39.4 & 7.2 \\

\cmidrule(lr){2-11}
& SDTagNet~\cite{sdtagnet}                   & NeurIPS'25  & C+SD    & $\times$    & 110 & 11.7 & 25.3 & 40.5 & 25.8 & 9.6 \\
& PriorDrive$^{*}$~\cite{priordrive}         & AAAI'26     & C+SD+HD & \checkmark  & 110 & 44.7 & 35.5 & 47.3 & 42.5 & 8.2 \\

\cmidrule(lr){2-11}
& SatforHDMap~\cite{satforhdmap}             & ICRA'24     & C+SA    & $\times$    & 110 & 44.1 & 12.1 & 24.8 & 27.0 & 14.1 \\

\cmidrule(lr){2-11}
& \textbf{ReSMap (Ours)}                              & -           & C+SA    & $\times$    & 24  & 63.1 & 30.0 & 48.9 & 47.3 & 16.1 \\
& \textbf{ReSMap (Ours)}                              & -           & C+SA    & \checkmark  & 24  & \textbf{69.5} & \textbf{36.6} & 58.7 & \textbf{54.9} & 10.2 \\

\midrule

\multirow{8}{*}{\raisebox{-0.8em}{\rotatebox[origin=c]{90}{\textbf{100 m × 50 m}}}}
& MapTracker$^{\dagger}$~\cite{maptracker}   & ECCV'24     & C       & \checkmark  & 58  & 45.9 & 24.3 & 38.4 & 36.2 & 13.4 \\

\cmidrule(lr){2-11}
& MapTRv2~\cite{maptrv2}                     & IJCV'24     & C+L     & $\times$    & 24  & 16.6 & 27.6 & 45.0 & 29.8 & 7.6 \\
& DAMap~\cite{damap}                         & ICCV'25     & C+L     & $\times$    & 24  & 13.7 & 27.6 & 48.7 & 30.0 & 7.2 \\

\cmidrule(lr){2-11}
& SDTagNet~\cite{sdtagnet}                   & NeurIPS'25  & C+SD    & $\times$    & 30  & 11.9 & 25.5 & 30.8 & 22.7 & 8.4 \\

\cmidrule(lr){2-11}
& SatforHDMap~\cite{satforhdmap}             & ICRA'24     & C+SA    & $\times$    & 24  & 40.0 & 9.9 & 10.4 & 20.1 & 14.2 \\

\cmidrule(lr){2-11}
& \textbf{ReSMap (Ours)}                              & -           & C+SA    & $\times$    & 24  & 74.5 & 30.2 & 44.9 & 49.9 & 15.9 \\
& \textbf{ReSMap (Ours)}                              & -           & C+SA    & \checkmark  & 24  & \textbf{77.6} & \textbf{35.0} & \textbf{51.3} & \textbf{54.7} & 10.2 \\

\bottomrule
\end{tabular}
\end{table*}

%% file: tables/robustness_performance_new.tex
\newcommand{\drop}[1]{{\tiny\textcolor{red}{(#1\%)}}}
\newcommand{\bdrop}[1]{{\tiny\textcolor{red}{\textbf{(#1\%)}}}}

\begin{table*}[t]
\centering
\normalsize
\caption{Robustness on the geographically non-overlapping split of nuScenes~\cite{streammapnet}, at $60\,\mathrm{m}\times30\,\mathrm{m}$.
\emph{Front-1}, \emph{Front-3}, and \emph{All} zero out the front camera, the
three front cameras, and all six cameras at inference, respectively. Each cell shows
mAP and the relative change from Clean,
$(\mathrm{Drop}-\mathrm{Clean})/\mathrm{Clean}\times100\%$, in red.
Symbols as in Tab.~\ref{tab:mapping_performance}.}
\label{tab:robustness_performance_new}
\setlength{\tabcolsep}{4pt}
\renewcommand{\arraystretch}{1.0}
\begin{tabular}{l c c c c c c c}
\toprule
& & & & & \multicolumn{3}{c}{Camera Drop} \\
\cmidrule(lr){6-8}
Method & Mod. & Temp. & Ep. & Clean & Front-1 & Front-3 & All \\
\midrule
MapTracker$^{\dagger}$~\cite{maptracker} & C & \checkmark & 58 & 40.3 & 23.0 \drop{-42.9} & 10.8 \drop{-73.2} & 0.1 \drop{-99.8} \\
\midrule
MapTRv2~\cite{maptrv2} & C+L & $\times$ & 110 & 36.6 & 32.2 \drop{-12.0} & 29.5 \drop{-19.4} & 23.8 \drop{-35.0} \\
DAMap~\cite{damap} & C+L & $\times$ & 110 & 39.4 & 35.8 \drop{-9.1} & 34.2 \drop{-13.2} & 29.2 \drop{-25.9} \\
\midrule
SDTagNet~\cite{sdtagnet} & C+SD & $\times$ & 110 & 25.8 & 19.8 \drop{-23.3} & 14.5 \drop{-43.8} & 4.0 \drop{-84.5} \\
\midrule
SatforHDMap~\cite{satforhdmap} & C+SA & $\times$ & 110 & 27.0 & 21.5 \drop{-20.4} & 21.3 \drop{-21.1} & 14.3 \drop{-47.0} \\
\midrule
\textbf{ReSMap (Ours)} & C+SA & $\times$ & 24 & 47.3 & 44.2 \bdrop{-6.6} & 42.6 \bdrop{-9.9} & \textbf{40.1} \bdrop{-15.2} \\
\textbf{ReSMap (Ours)} & C+SA & \checkmark & 24 & \textbf{54.9} & \textbf{50.3} \drop{-8.4} & \textbf{45.1} \drop{-17.9} & 32.9 \drop{-40.1} \\
\bottomrule
\end{tabular}
\vspace{-2mm}
\end{table*}

%% file: figures/qual_scene0053/qual_scene0053.tex
\begin{figure*}[t]
\centering

\definecolor{clsped}{HTML}{1F77B4}
\definecolor{clsdiv}{HTML}{D62728}
\definecolor{clsbnd}{HTML}{2CA02C}

\setlength{\tabcolsep}{0pt}
\renewcommand{\arraystretch}{1.0}
\begin{tabular}{@{}c@{\hskip 2pt}c@{\hskip 2pt}c@{\hskip 6pt}c@{}}
\multicolumn{3}{c}{\footnotesize\textsf{Multi-view Images}} &
\multicolumn{1}{c}{\footnotesize\textsf{Satellite Image}} \\[2pt]
\includegraphics[height=0.115\textwidth]{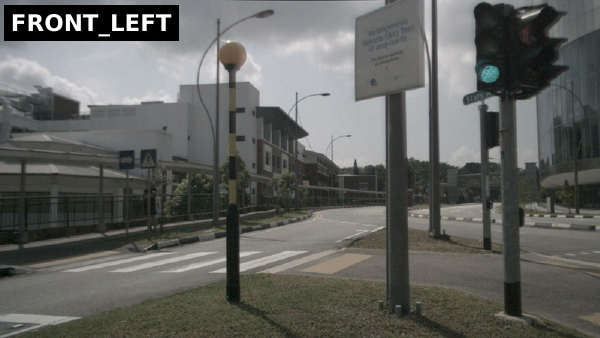} &
\includegraphics[height=0.115\textwidth]{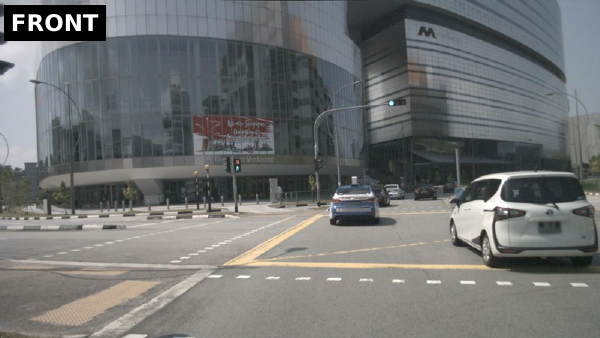} &
\includegraphics[height=0.115\textwidth]{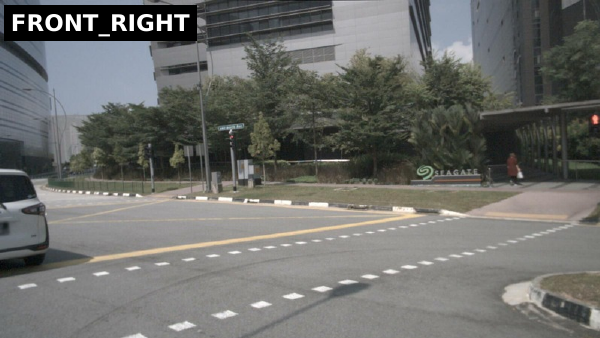} &
\includegraphics[height=0.115\textwidth]{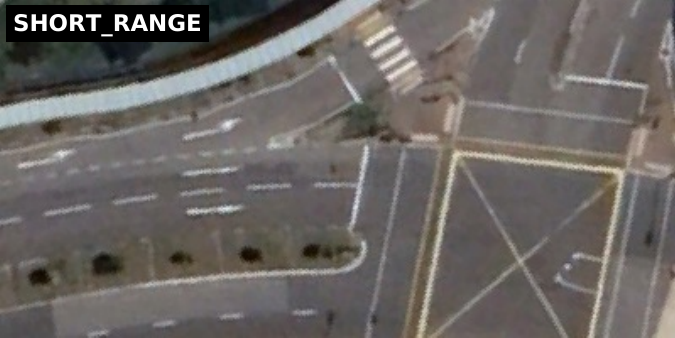} \\
\includegraphics[height=0.115\textwidth]{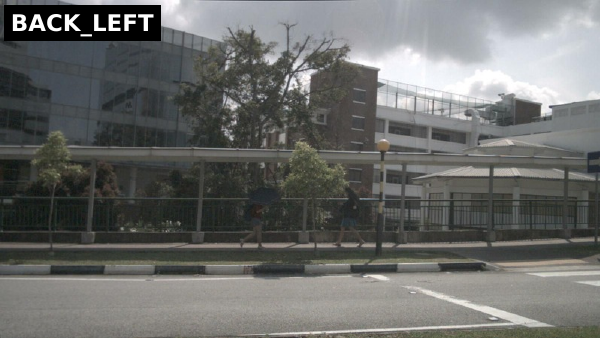} &
\includegraphics[height=0.115\textwidth]{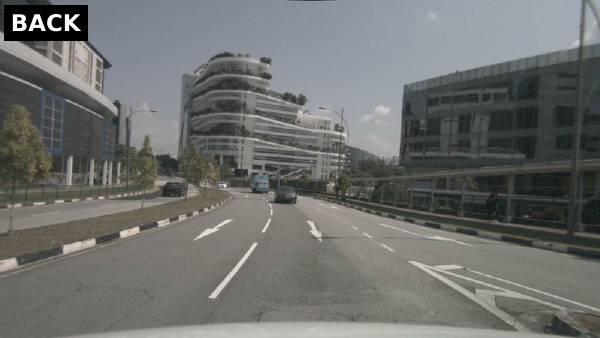} &
\includegraphics[height=0.115\textwidth]{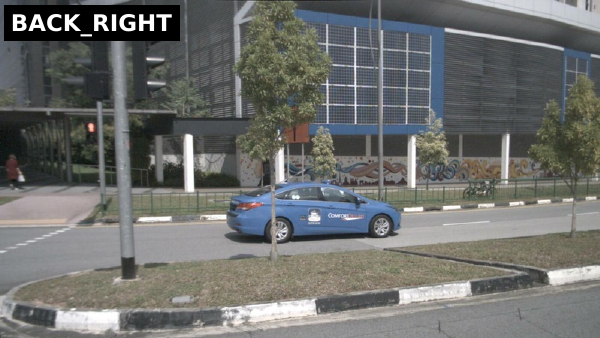} &
\includegraphics[height=0.115\textwidth]{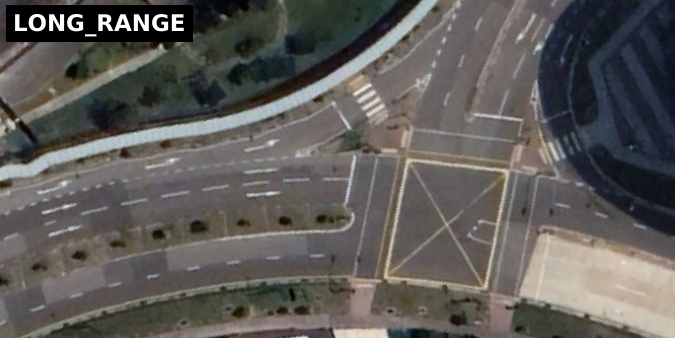} \\
\end{tabular}

\vspace{8pt}

\newlength{\bevh}
\setlength{\bevh}{0.092\textwidth}    %
\newcommand{\rowlbl}[1]{\parbox[c][\bevh][c]{1.5em}{\centering\rotatebox{90}{\scriptsize\textit{#1}}}}
\newcommand{\bevcell}[1]{\parbox[c][\bevh][c]{0.18\textwidth}{\centering\fbox{\includegraphics[width=0.09\textwidth, angle=-90]{#1}}}}

\setlength{\fboxsep}{0pt}
\setlength{\fboxrule}{0.4pt}
\setlength{\tabcolsep}{2pt}
\renewcommand{\arraystretch}{1.0}
\begin{tabular}{c@{\hskip 4pt}ccccc}
 & {\footnotesize\textsf{Ground Truth}} & {\footnotesize\textbf{\textsf{\modelname{}(Ours)}}} & {\footnotesize\textsf{MapTracker~\cite{maptracker}}} & {\footnotesize\textsf{SDTagNet~\cite{sdtagnet}}} & {\footnotesize\textsf{SatforHDMap~\cite{satforhdmap}}} \\
\rowlbl{60\,m$\times$30\,m} &
\bevcell{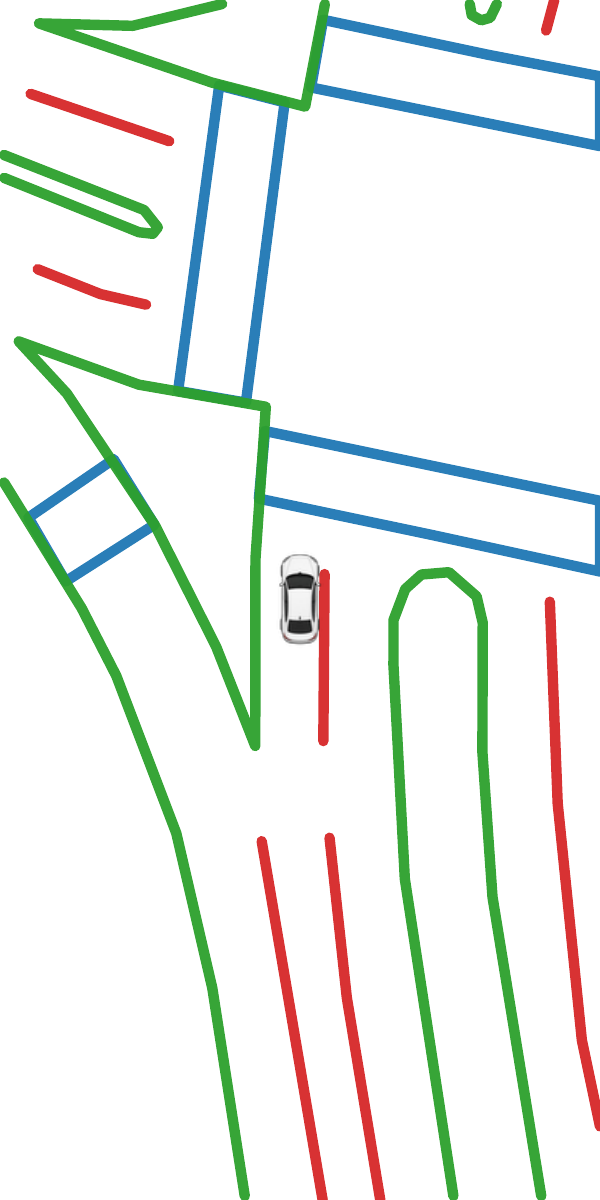} &
\bevcell{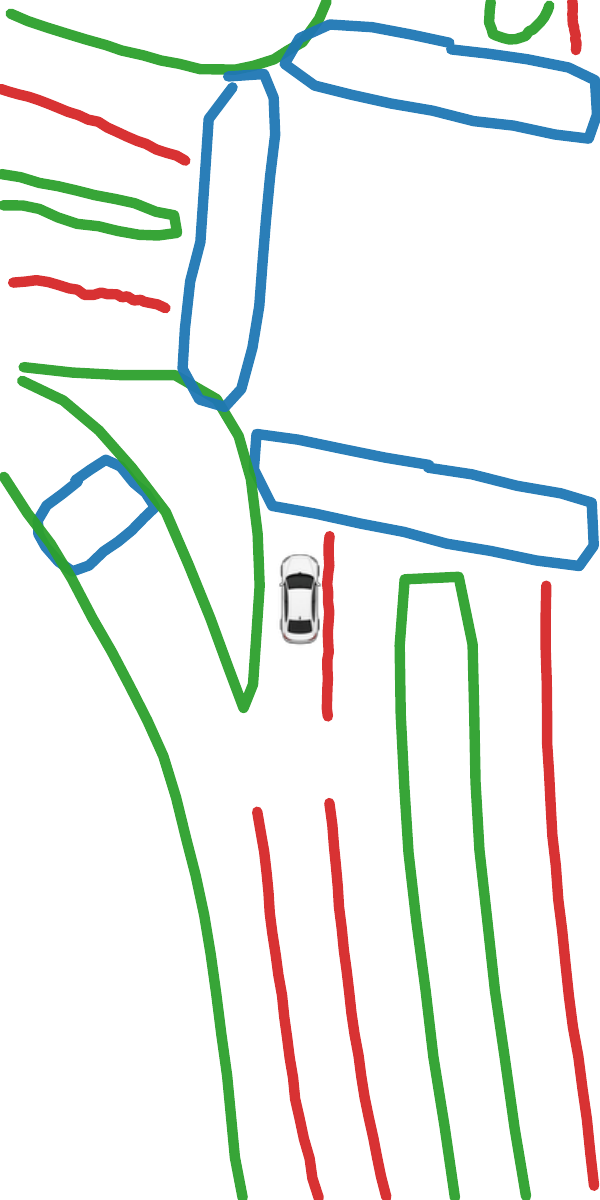} &
\bevcell{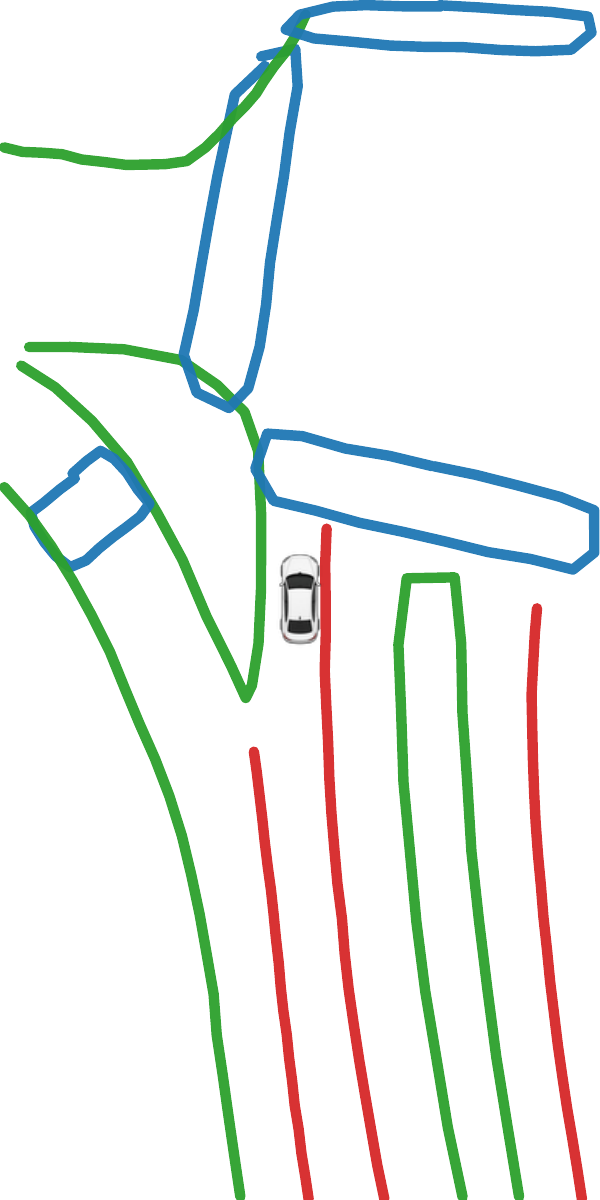} &
\bevcell{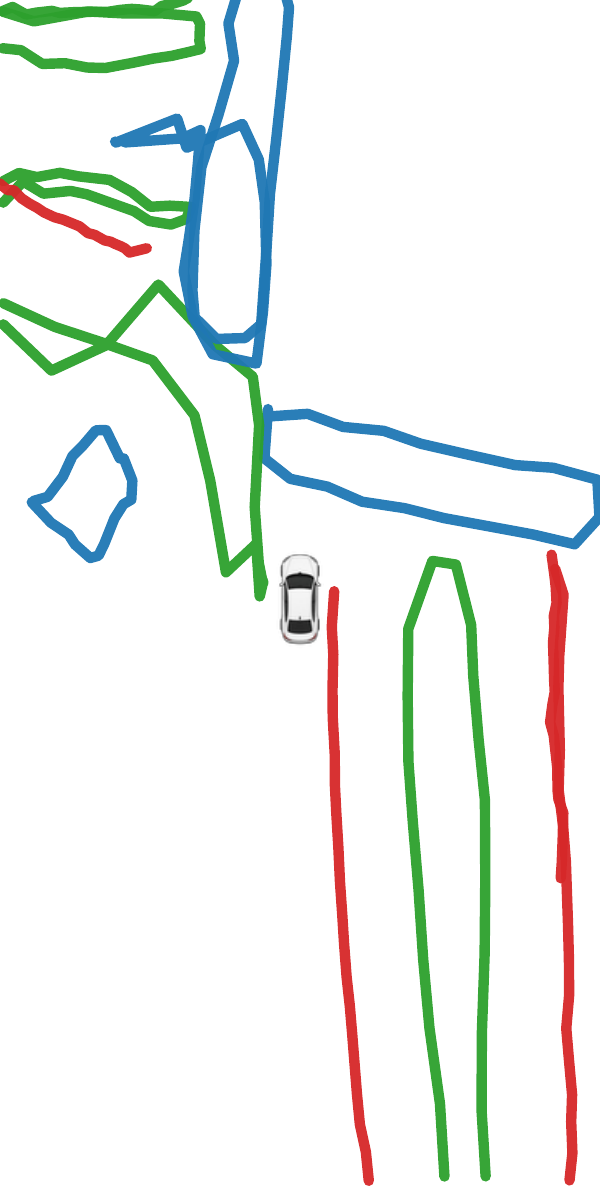} &
\bevcell{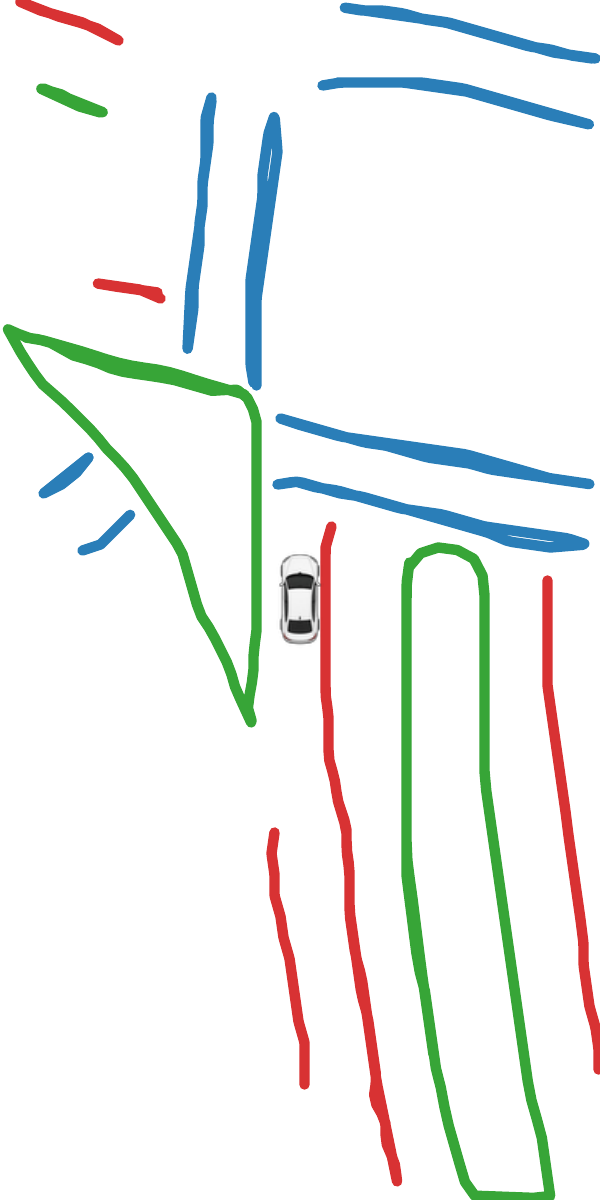} \\
\noalign{\vskip 3pt}
\rowlbl{Front-3 drop} &
\bevcell{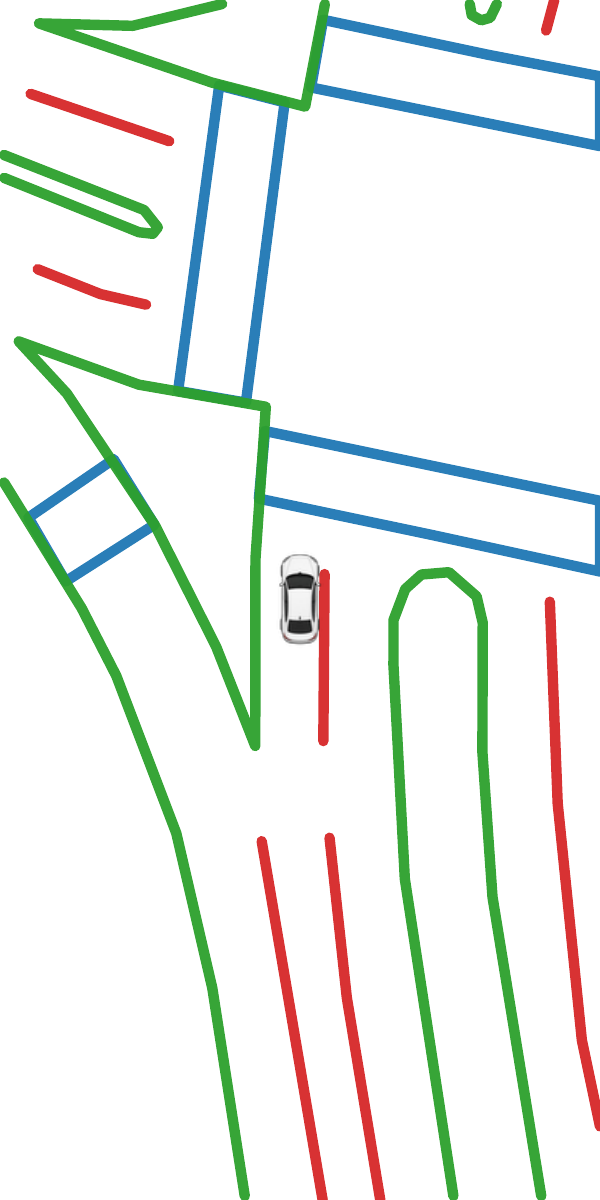} &
\bevcell{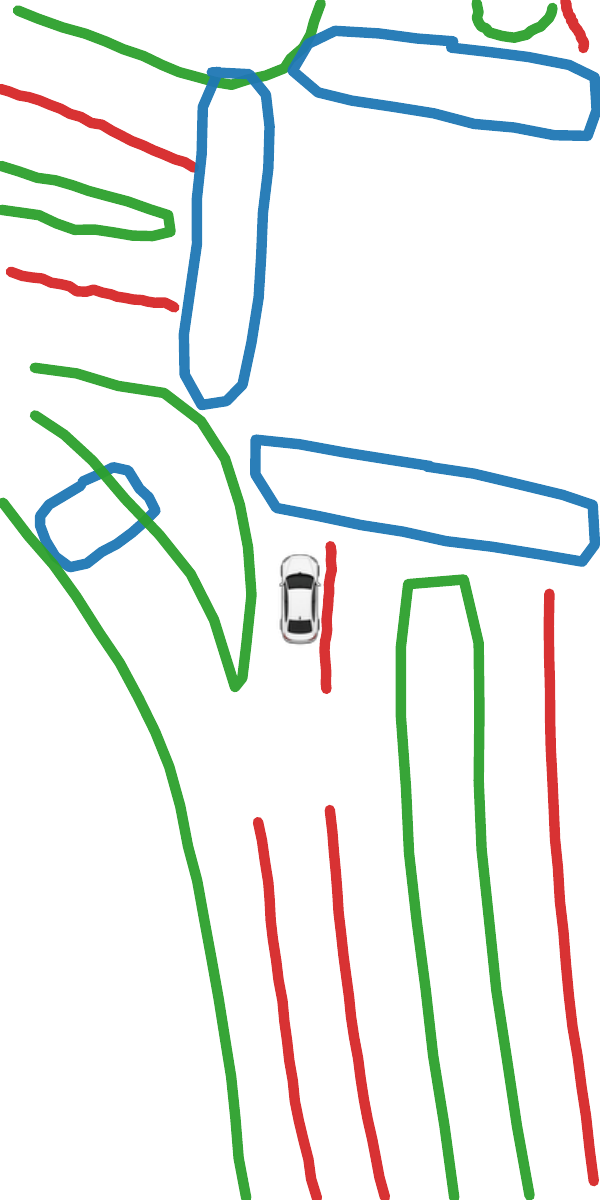} &
\bevcell{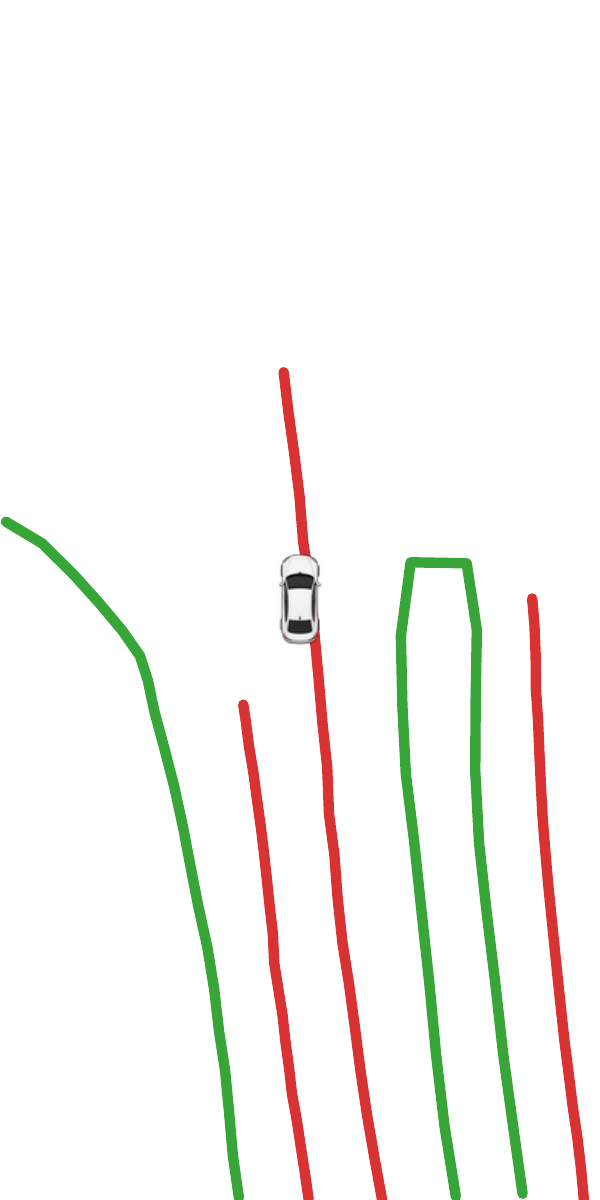} &
\bevcell{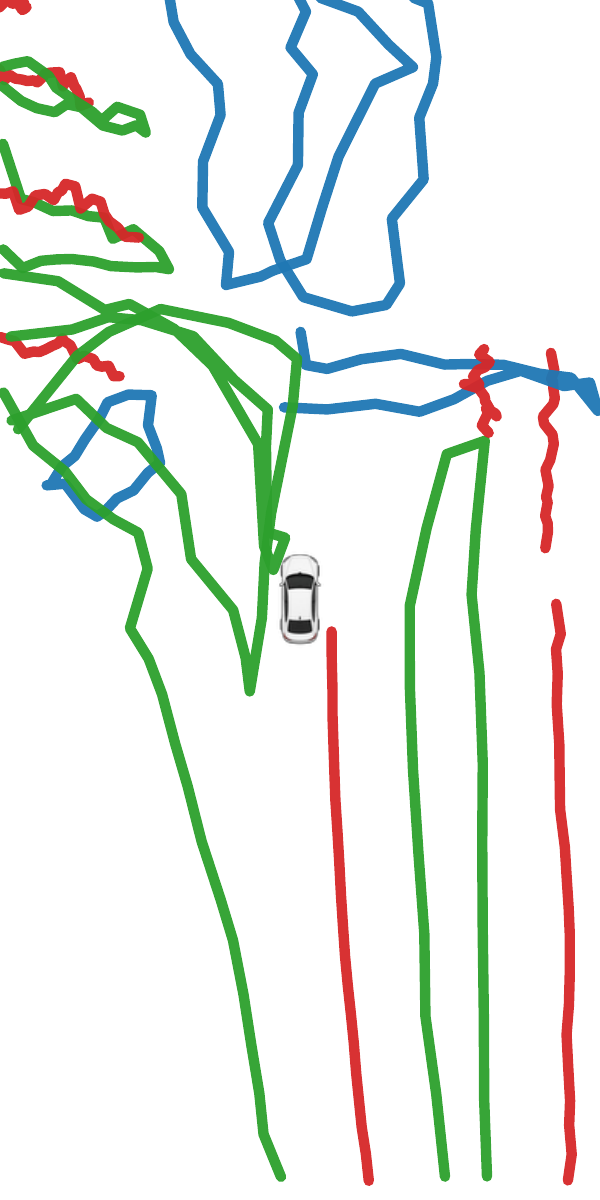} &
\bevcell{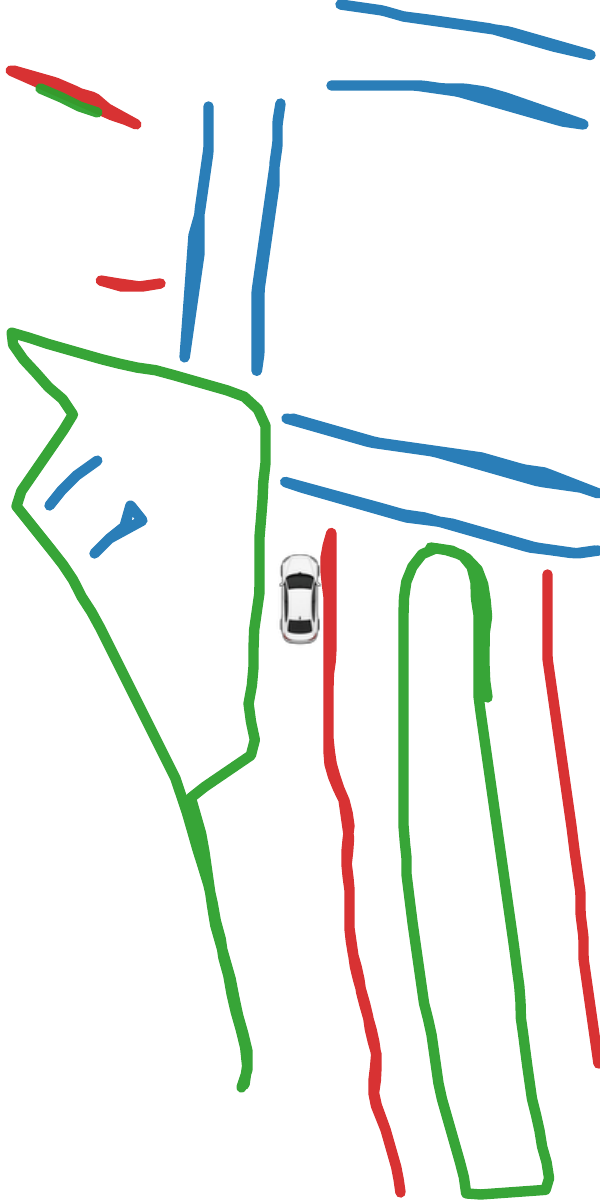} \\
\noalign{\vskip 3pt}
\rowlbl{100\,m$\times$50\,m} &
\bevcell{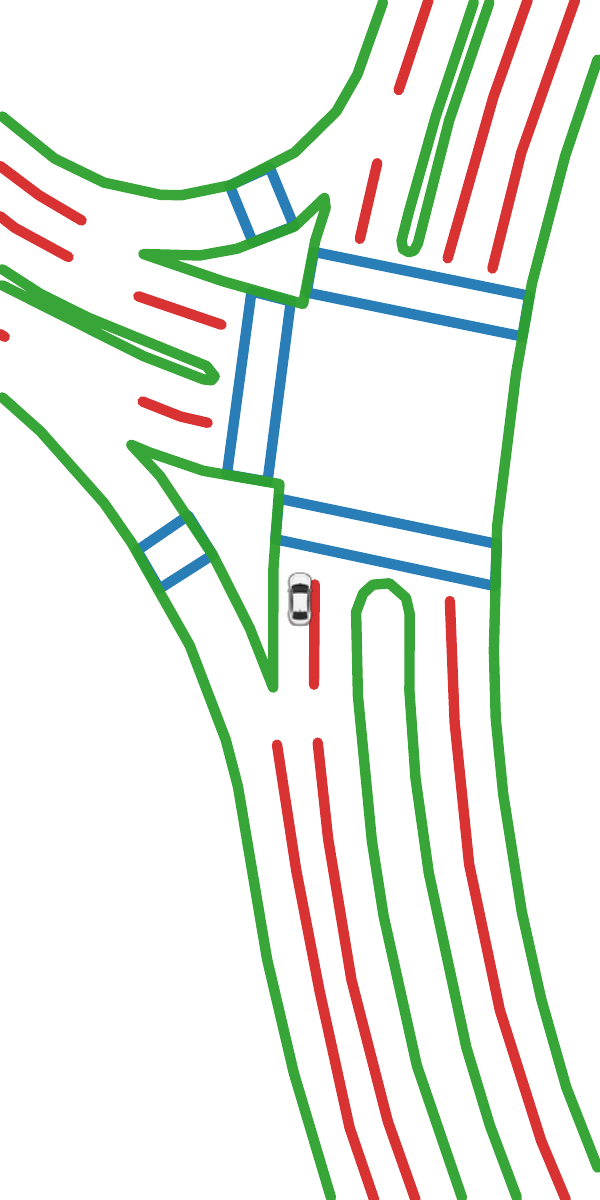} &
\bevcell{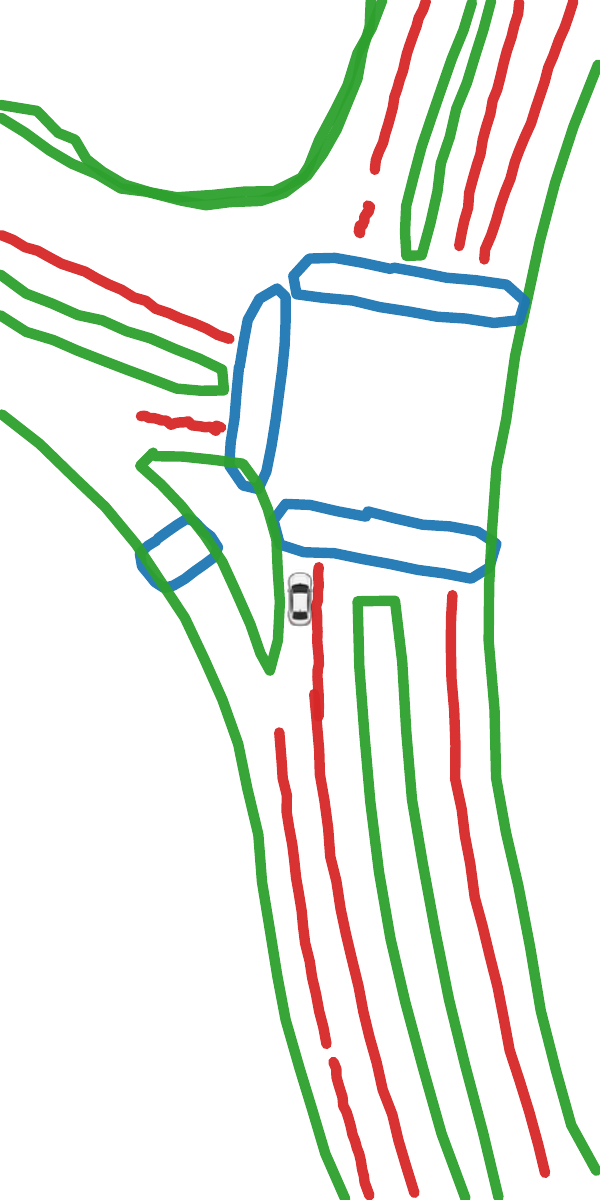} &
\bevcell{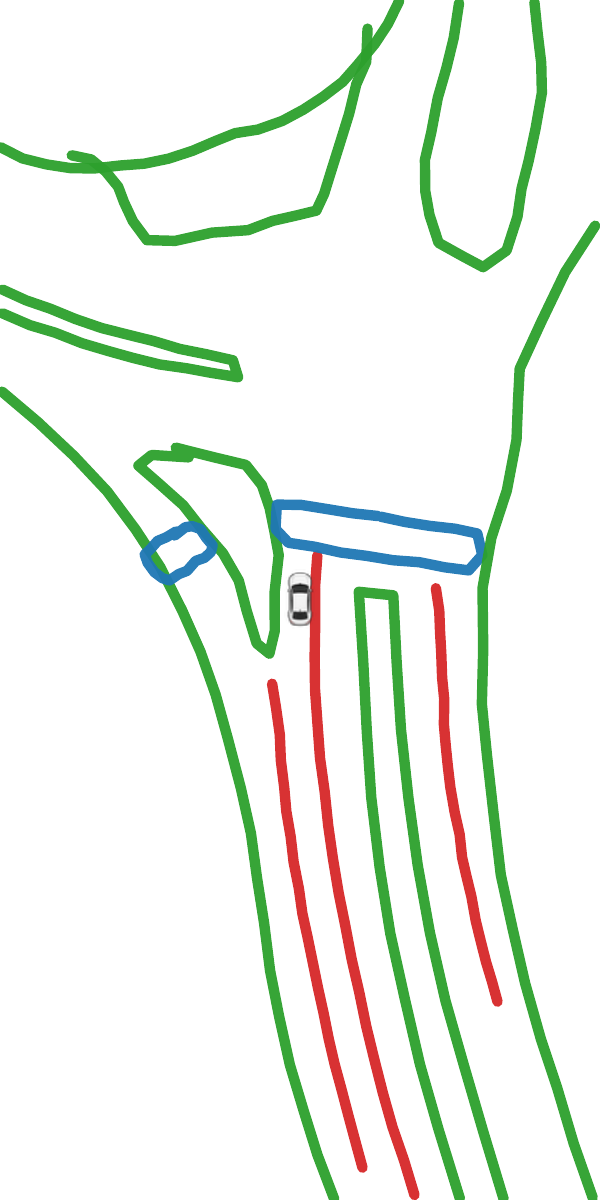} &
\bevcell{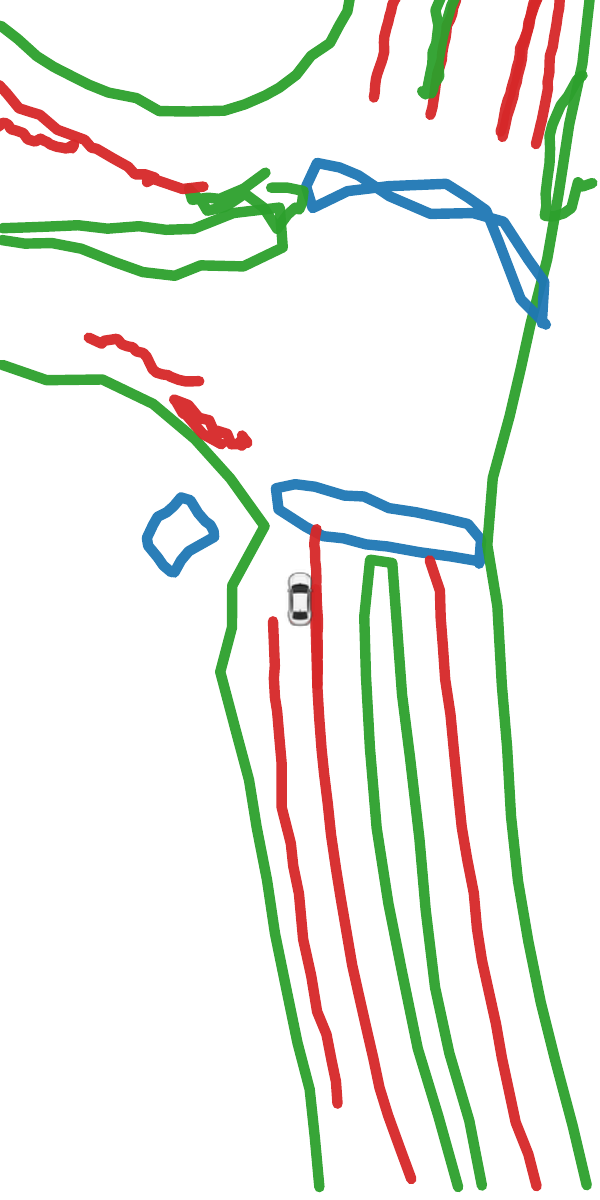} &
\bevcell{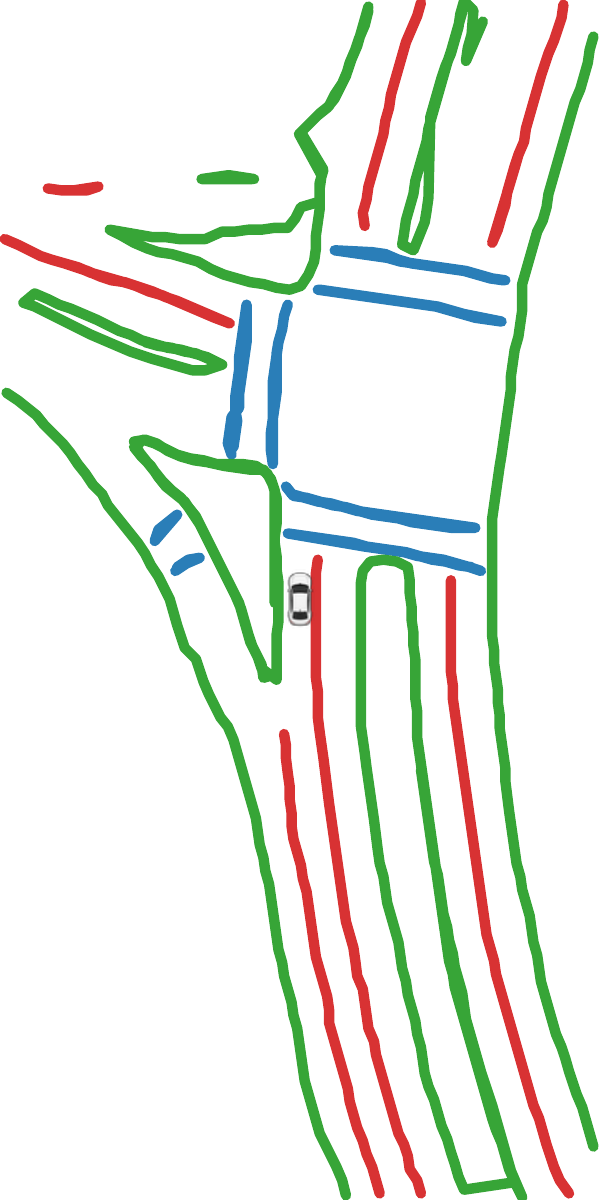} \\
\end{tabular}

\vspace{5pt}
{\footnotesize
\textcolor{clsped}{\rule[-1pt]{8pt}{8pt}}~ped.\,crossing\quad
\textcolor{clsdiv}{\rule[-1pt]{8pt}{8pt}}~divider\quad
\textcolor{clsbnd}{\rule[-1pt]{8pt}{8pt}}~boundary%
}

\caption{%
Qualitative comparison of four methods. Top: multi-view cameras and aerial
satellite images for the two evaluation ROIs ($60\,\mathrm{m}\times30\,\mathrm{m}$
and $100\,\mathrm{m}\times50\,\mathrm{m}$). Bottom: predicted vector maps under
\emph{Clean}, \emph{Front-3} at
$60\,\mathrm{m}\times30\,\mathrm{m}$, and \emph{Clean} at
$100\,\mathrm{m}\times50\,\mathrm{m}$ (forward axis horizontal, ego facing
right).%
}
\label{fig:qual}
\end{figure*}

%% file: tables/ocmq_ablation.tex
\begin{table}[t]
\centering
\caption{Ablation study on the geographically non-overlapping split of nuScenes~\cite{streammapnet}.
Left: BEV fusion method ablation with the decoder fixed to full OCMQ.
Right: decoder design ablation with the BEV fusion method fixed to PTF.
Robustness columns report mIoU (left) and mAP (right) under front-camera and all-camera dropout; mIoU is the BEV segmentation mIoU of $\mathrm{Seg}_{\mathrm{fuse}}$ on $\mathbf{F}_{\mathrm{fuse}}$.}
\label{tab:ocmq_ablation}
\scriptsize
\renewcommand{\arraystretch}{1.08}
\newlength{\abltableheight}
\setlength{\abltableheight}{16mm}

\begin{minipage}[t]{0.42\linewidth}
\centering
{\small\textbf{(a) BEV fusion method (Sec.~\ref{sec:sc_fusion})}}\\[2pt]
\setlength{\tabcolsep}{2.2pt}
\renewcommand{\arraystretch}{1.2425}
\resizebox{\linewidth}{!}{%
\begin{tabular}{@{}l c c c@{}}
\toprule
\multicolumn{1}{c}{\multirow{2}{*}{Fusion}}
& \multicolumn{3}{c}{mIoU$\uparrow$} \\
\cmidrule(lr){2-4}
& Clean & Front-1 & All \\
\midrule
Cross-Attn~\cite{satforhdmap}
    & 45.5
    & 34.2 {\tiny\textcolor{red}{(-24.8\%)}}
    & \phantom{0}0.9 {\tiny\textcolor{red}{(-98.0\%)}} \\
ConvFuser~\cite{satmap}
    & 52.6
    & 48.6 {\tiny\textcolor{red}{(-7.6\%)}}
    & 35.9 {\tiny\textcolor{red}{(-31.7\%)}} \\
CIT+DDF~\cite{mapfusion}
    & 52.3
    & 47.9 {\tiny\textcolor{red}{(-8.4\%)}}
    & 30.4 {\tiny\textcolor{red}{(-41.9\%)}} \\
\rowcolor{gray!10}
\textbf{PTF}
    & \textbf{53.5}
    & \textbf{50.9} {\tiny\textcolor{red}{\textbf{(-4.9\%)}}}
    & \textbf{40.3} {\tiny\textcolor{red}{\textbf{(-24.7\%)}}} \\
\bottomrule
\end{tabular}%
}
\end{minipage}
\hfill
\begin{minipage}[t]{0.56\linewidth}
\centering
{\small\textbf{(b) Decoder design (Sec.~\ref{sec:mm_decoder})}}\\[2pt]
\setlength{\tabcolsep}{2.0pt}
\resizebox{\linewidth}{\abltableheight}{%
\begin{tabular}{@{}c c c c c c@{}}
\toprule
\multicolumn{3}{c}{Decoder design} & \multicolumn{3}{c}{mAP$\uparrow$} \\
\cmidrule(lr){1-3}
\cmidrule(lr){4-6}
\makecell{Per-mod.\\CA}
& \makecell{Mod. emb.\\+ IntraSA}
& \makecell{Gated\\merging}
& Clean
& Front-1
& All \\
\midrule
-- & -- & --
    & 52.8
    & 46.2 {\tiny\textcolor{red}{(-12.5\%)}}
    & 25.7 {\tiny\textcolor{red}{(-51.3\%)}} \\
$\checkmark$ & -- & mean
    & 53.6
    & 48.6 {\tiny\textcolor{red}{(-9.3\%)}}
    & 26.1 {\tiny\textcolor{red}{(-51.3\%)}} \\
$\checkmark$ & $\checkmark$ & mean
    & 53.2
    & 48.2 {\tiny\textcolor{red}{(-9.4\%)}}
    & 27.4 {\tiny\textcolor{red}{(-48.5\%)}} \\
\rowcolor{gray!10}
$\checkmark$ & $\checkmark$ & $\checkmark$
    & \textbf{54.9}
    & \textbf{50.3} {\tiny\textcolor{red}{\textbf{(-8.4\%)}}}
    & \textbf{32.9} {\tiny\textcolor{red}{\textbf{(-40.1\%)}}} \\
\bottomrule
\end{tabular}%
}
\end{minipage}

\vspace{-3pt}
\end{table}

%% file: section/5_Conclusion.tex
\section{Conclusion}
\label{sec:conclusion}

We present \modelname{}, a camera-satellite fusion framework for online HD map construction that treats satellite images not only as a spatial prior but also as a failure-decoupled redundancy modality. The parameter-free Predict-then-Fuse (PTF) interface mixes camera and satellite BEV features under each branch's own segmentation confidence. The Object-Centric Cooperative Multi-modal Query (OCMQ) decoder preserves modality-specific information until the prediction stage and adaptively recombines per-modality query updates through a learned per-token gate. Our model achieves state-of-the-art online HD map construction performance under clean conditions across $60\,\mathrm{m}\times30\,\mathrm{m}$ and $100\,\mathrm{m}\times50\,\mathrm{m}$ ROIs and under camera failure, showing that our model effectively utilizes satellite images as a strong spatial prior and a practical redundancy when cameras fail.

\paragraph{Limitations.}
Our method assumes that the satellite image is accurately localized with respect to the ego vehicle. Infrequent GPS/INS errors can cause spatial misalignment, weakening the correspondence between the camera and satellite streams; Section~\ref{sec:appendix_localization_noise} quantifies this case. Falling back on satellite images also assumes that the cached tiles are reasonably current: re-striping or construction between capture and deployment weakens the prior. Another open question is how the model behaves when the satellite view itself carries little usable evidence---under shadows, tunnels, dense tree canopy, or tall buildings that occlude the top-down view. Public driving datasets, however, do not contain such scenes paired with high-quality satellite images, so we leave this regime to future work.

%% file: section/A_Appendix.tex
\section{Technical Appendices and Supplementary Material}
\raggedbottom
\label{app:appendix}

\begin{center}
\small
\renewcommand{\arraystretch}{1.1}
\begin{tabular}{@{}l l l@{}}
\toprule
& Content & Main text \\
\midrule
\ref{app:impl} & Implementation details & \S\ref{sec:setup} \\
\ref{app:oldsplit} & Results on the original split of nuScenes & \S\ref{sec:quantitative} \\
\ref{app:av2} & Results on Argoverse~2 & \S\ref{sec:setup} \\
\ref{app:mapbench} & MapBench sensor-corruption robustness & \S\ref{sec:quantitative} \\
\ref{app:aug} & BEV-feature dropout versus architecture & \S\ref{sec:setup} \\
\ref{sec:appendix_localization_noise} & Robustness to satellite localization noise & \S\ref{sec:conclusion} \\
\ref{app:branch} & Per-branch decoder: a preliminary study & \S\ref{sec:mm_decoder} \\
\ref{app:temporal_robustness} & Why non-temporal outperforms temporal under severe camera failure & \S\ref{sec:quantitative} \\
\ref{app:qual} & Additional qualitative results & \S\ref{sec:qualitative} \\
\bottomrule
\end{tabular}
\end{center}

\subsection{Implementation Details}
\label{app:impl}

\paragraph{Training and optimization.}
We train our model with a three-stage curriculum on $4$ NVIDIA RTX
4090 48GB GPUs with batch size of $6$, using AdamW with weight
decay $0.01$ and a cosine schedule with a linear warm-up
(ratio $1/3$; $1000$ iterations in Stages~1--2 and $600$ in Stage~3).
In Stage~1, we pre-train the BEV pipeline (camera/satellite
encoders and segmentation head) end-to-end under the
segmentation loss for $6$ epochs with base lr $3.5{\times}10^{-4}$.
In Stage~2, we freeze all BEV-side modules and train only the OCMQ
head with the matching loss for $6$ epochs at base lr
$1.4{\times}10^{-4}$.
In Stage~3, we unfreeze the full network and fine-tune all
components jointly under the combined objective for another $12$
epochs with the same lr.
A complete three-stage training run takes approximately $30$ hours.
Up to $20$ past frames are propagated
through the temporal memory following MapTracker (disabled for the
non-temporal version).

\paragraph{Architecture hyper-parameters.}
The camera branch employs a ResNet-50 backbone with DCNv2 in the last two stages and an FPN producing $3$ feature levels at $C{=}256$, lifted to a $100\!\times\!50$ BEV feature grid by a $2$-layer BEVFormer encoder
($4$ pillar points, $8$ deformable sampling points per level).
The satellite branch uses a ResNet-50 backbone whose multi-scale
features are projected to the same BEV grid.
The PTF reliability gate is a lightweight max-sigmoid module that
computes a per-cell confidence map from each branch's segmentation output and produces
the fused BEV feature as a confidence-weighted sum.
The OCMQ decoder has $L{=}6$ transformer layers (embedding
$512$, $8$ heads, FFN $1024$, dropout $0.1$) and $N{=}100$ object
queries; each query regresses a polyline of $20$ points in $2$D BEV
coordinates.

\paragraph{Data augmentation.}
Multi-view images are resized to $480{\times}800$. During training we apply \emph{BEV-feature dropout}, which perturbs the BEV feature maps after the encoders rather than the sensor inputs. Each iteration applies at most one of four mutually exclusive cases: $\mathbf{F}_{\mathrm{cam}}$ is zeroed ($p{=}0.10$), perturbed with Gaussian noise ($p{=}0.10$), or zeroed inside a random rectangular region covering a quarter of the BEV grid ($p{=}0.05$), or $\mathbf{F}_{\mathrm{sat}}$ is zeroed ($p{=}0.05$); the remaining $70\%$ of iterations are clean. The camera-failure protocol used for evaluation is a different operation: it zeroes the raw multi-view images before the backbone, so the model is never trained on the corruption it is evaluated under. Two parts of the model are outside the reach of this augmentation: the PTF gate has no parameters and $r_m$ enters Eq.~\eqref{eq:fusion} under stop-gradient, and the OCMQ decoder receives unmasked BEV features at every training step. In Stages 2 and 3, one of the camera, satellite and fused evidence streams is additionally excluded from the OCMQ gate per training scene ($p{=}0.3/0.3/0.1$), which keeps the gate from collapsing onto one stream.

\paragraph{Loss weights.}
The total objective in Eq.~\eqref{eq:total_loss} is instantiated
with classification weight $\lambda^{cls}{=}5.0$
(focal loss, $\gamma{=}2.0$, $\alpha{=}0.25$), regression weight
$\lambda^{reg}{=}50.0$ (smooth-L1 on $20$-point polylines,
$\beta{=}0.01$), and BEV-segmentation weights
$\lambda^{seg}_{focal}{=}10.0$ and $\lambda^{seg}_{dice}{=}1.0$ for
the auxiliary semantic head. The same Hungarian cost
($\lambda^{cls}_{cost}{=}5.0$, $\lambda^{reg}_{cost}{=}50.0$,
permutation-invariant) is used for query--GT matching.

\FloatBarrier
\subsection{Results on the Original Split}
\label{app:oldsplit}

The original split of nuScenes follows the standard partition,
where the training and validation sets contain geographically overlapping regions.
As a result, many road structures seen during training can reappear during evaluation,
potentially overestimating generalization performance~\cite{streammapnet}. For comparability with prior work,
we nevertheless report additional results on the original split under the same
evaluation protocol.

Tab.~\ref{tab:mapping_performance_old} reports performance under the original split at
$60\,\mathrm{m}\times30\,\mathrm{m}$. Our model achieves the best mAP in both temporal and
non-temporal settings, reaching $84.9$ mAP and $82.8$ mAP, respectively.
Notably, our model without temporal memory already outperforms PriorDrive~\cite{priordrive} by
$+2.3$ mAP, while PriorDrive uses an SD map together with a noisy prior HD map and is trained for
$110$ epochs. With temporal memory, our model further increases the margin to $+4.4$ mAP.
Across both the original split and the geographically non-overlapping split
(Tab.~\ref{tab:mapping_performance}), our model consistently outperforms other baselines. This indicates that
the improvement is not tied to a particular data split.

\input{tables/mapping_performance_old}

\FloatBarrier
\subsection{Results on Argoverse 2}
\label{app:av2}

To test whether the results carry over to a second dataset and a second imagery source, we train and evaluate on the geographically non-overlapping split~\cite{streammapnet} of Argoverse~2~\cite{argoverse2} at $60\,\mathrm{m}\times30\,\mathrm{m}$. Since AID4AD covers nuScenes only, we use satellite tiles from Google Maps, cropped per sample to the same ego-aligned BEV region as on nuScenes. Unlike the curated AID4AD imagery, these tiles are present-day retrievals several years apart from the recorded drives, and they are used without ICP-based alignment refinement, so they keep the staleness and localization error a deployed system would see. Argoverse~2 has seven ring cameras: Front-1 zeroes the front-center camera, Front-3 also the front-left and front-right cameras, and All zeroes all seven. The baselines are retrained under the same protocol as on nuScenes.

Tab.~\ref{tab:av2} reports the results. Our model obtains the highest mAP under every condition. Under all-camera drop the non-temporal variant retains $46.9$ mAP and the temporal one $30.9$, while the camera-only MapTracker collapses to $0.2$, SDTagNet falls to $18.6$, and SatforHDMap to $21.0$. Both the clean-condition advantage and the failure-time behavior thus carry over to a second dataset and a less curated imagery source.

\input{tables/av2_performance}

\FloatBarrier
\subsection{Additional Camera-Corruption Robustness (MapBench)}
\label{app:mapbench}

We additionally evaluate our model on the original split of nuScenes under the sensor-corruption protocol proposed in MapBench~\cite{mapbench}, which aggregates camera-drop and image-degradation cases into two summary metrics: the mean Resilience Rate mRR$\uparrow$ (clean-normalized retention across all corruptions) and the mean Corruption Error mCE$\downarrow$ (error normalized by that of a reference model, MapTR~\cite{maptr} as in MapBench).
Tab.~\ref{tab:robustness_performance_old} compares our model with the camera-only and multi-modal baselines under the MapBench protocol, evaluated by us except where marked, and additionally reports mAP under Front-1, Front-3, and All-camera drop together with the relative change from clean.
Our model with temporal modeling retains $82.0$ mAP under Front-1 drop ($-3.4\%$ from clean), $77.9$ mAP under Front-3 drop ($-8.2\%$), and $62.5$ mAP under All-camera drop ($-26.4\%$), smaller relative drops than every baseline.
Camera-only baselines such as MapTracker collapse rapidly ($-46.0\%$ at Front-1, $-77.7\%$ at Front-3), and even the LiDAR-fused DAMap loses more than half of its clean mAP under All-camera drop ($-51.2\%$).
These numbers confirm that the redundancy gains observed on the geographically non-overlapping split (Tab.~\ref{tab:robustness_performance_new}) are not an artifact of the split, and that satellite priors continue to act as a failure-decoupled modality on the original split of nuScenes.

\input{tables/robustness_performance_old}

We also run the MapBench protocol on the geographically non-overlapping split (Tab.~\ref{tab:mapbench_geo}), with the same $24$ corruptions (eight types at three severities) and MapTracker as the mCE reference. Our model has the highest mRR ($90.8\%$) and the lowest mCE ($0.712$) among the compared methods, so its robustness to image degradation also holds on the split where training and validation do not share road layouts.

\input{tables/mapbench_geo}

\FloatBarrier
\subsection{BEV-Feature Dropout versus Architecture}
\label{app:analyses}
\label{app:aug}
Tab.~\ref{tab:aug} retrains our model without BEV-feature dropout, and SatforHDMap, SDTagNet, and DAMap with it, applied identically to their own BEV features. Removing the augmentation from our model leaves its failure behavior essentially unchanged: $35.5$ against $36.0$ mAP under all-camera drop ($-31.2\%$ against $-32.1\%$), at a clean cost of $1.4$ mAP. The failure-time reallocation is therefore architectural, and the augmentation contributes a modest clean-accuracy gain. Conversely, the augmentation shifts SDTagNet by at most $1.5$ mAP, since its SD prior lacks lane-level geometry to fall back on. It improves SatforHDMap's failure behavior modestly ($14.3$ to $16.7$ under all-camera drop), but SatforHDMap remains far below our unaugmented model ($35.5$), from a clean accuracy ($27.6$) below even the camera-only baseline. For DAMap, whose fallback is onboard LiDAR, the augmentation raises all-camera-drop mAP from $29.2$ to $30.7$ at a clean cost of $1.1$ mAP, which remains below our unaugmented model ($35.5$). That the satellite-prior baseline benefits most from failure exposure supports the premise of this work: satellite imagery is the prior best suited as a fallback, and the architectural question is how well a model exploits it. The fusion ablation in Tab.~\ref{tab:ocmq_ablation}(a) shows the same within our architecture, since every learnable fusion variant there was trained with this augmentation.

\begin{table}[!htbp]
\centering
\caption{BEV-feature dropout on and off. $^{\dagger}$Our model before hyperparameter tuning.}
\label{tab:aug}
\small
\setlength{\tabcolsep}{4pt}
\begin{tabular}{l c c c c c}
\toprule
Method & BEV-feature dropout & Clean & Front-1 & Front-3 & All \\
\midrule
SatforHDMap~\cite{satforhdmap} & no & 27.0 & 21.5 & 21.3 & 14.3 \\
SatforHDMap~\cite{satforhdmap} & yes & 27.6 & 24.9 & 21.0 & 16.7 \\
\midrule
SDTagNet~\cite{sdtagnet} & no & 25.8 & 19.8 & 14.5 & 4.0 \\
SDTagNet~\cite{sdtagnet} & yes & 26.3 & 20.1 & 13.0 & 4.3 \\
\midrule
DAMap~\cite{damap} & no & 39.4 & 35.8 & 34.2 & 29.2 \\
DAMap~\cite{damap} & yes & 38.3 & 35.2 & 34.5 & 30.7 \\
\midrule
ReSMap (temporal)$^{\dagger}$ & no & 51.6 & 48.2 & 43.6 & 35.5 \\
ReSMap (temporal)$^{\dagger}$ & yes & 53.0 & 50.0 & 44.4 & 36.0 \\
\bottomrule
\end{tabular}
\end{table}

\FloatBarrier
\subsection{Results on Localization Noise}
\label{sec:appendix_localization_noise}

Satellite images can be misaligned or temporally outdated due to GPS/INS drift,
signal loss, or latency in the localization stack. To assess robustness under
such errors, we apply two corruptions to the satellite image at inference, each with probability $p{=}0.2$ per sample on the
geographically non-overlapping split of nuScenes~\cite{streammapnet}:
(i) temporal swap, replacing the $t$-frame tile with the $t{\pm}1$ tile to
simulate a delayed localization update, and (ii) geometric noise, warping the
tile by $\mathcal{N}(0,\sigma_t{=}1\text{m})$ shift and/or
$\mathcal{N}(0,\sigma_r{=}5^\circ)$ rotation (Tab.~\ref{tab:satmap_corruption_robust}).

\input{tables/satellite_image_noise_performance}

\FloatBarrier
\subsection{Per-Branch Decoder: A Preliminary Study}
\label{app:branch}

To motivate the gated query merging in Eq.~\eqref{eq:gated_collapse}, we examine an alternative design in which each modality maintains its own pool of object queries and is matched independently against the ground truth.
Concretely, this baseline (which we call the \emph{per-branch decoder}) keeps three (Camera, Satellite, Fused) modality-specific query groups of size 50 each, totaling 150 queries that all act as final prediction candidates. Within each decoder layer, queries first attend within their own modality group through a block-diagonal masked self-attention, then perform per-modality deformable cross-attention exactly as in OCMQ (Sec.~\ref{sec:mm_decoder}), and finally exchange information across all 150 queries through a full self-attention. Crucially, no gated query merging is applied: the three groups remain disjoint until the final prediction layer, where Hungarian matching is performed over all 150 candidates, so every ground-truth instance is assigned to whichever modality already explains it best. The gated query merging of Eq.~\eqref{eq:gated_collapse} replaces this disjoint design by routing the modality-conditioned updates back into a single object-query pool before matching.
Tab.~\ref{tab:decoder_branch_perf}(a) reports per-branch mAP under clean inputs.
Each branch specializes to a different class: the camera branch is strongest on lane dividers and road boundaries, the satellite branch on pedestrian crossings, and the fused branch dominates the overall mAP almost entirely through AP$_p$.
This class--modality entanglement is a direct consequence of Hungarian matching applied to disjoint query pools: when each pool only minimizes its own loss, training assigns ground-truth instances to whichever pool already explains them best, leaving the other pools unsupervised on those classes.

\input{tables/decoder_branch_perf}

Tab.~\ref{tab:decoder_branch_perf}(b) shows that this entanglement becomes brittle once the dominant modality fails.
The same model trained with independent pools loses $15.9$ mAP when the three front cameras are dropped, and the largest single-class drop ($\mathrm{AP}_b\!:\!-28.9$) is on the boundary class that the camera branch had specialized in.
Because no other branch was supervised on $\mathrm{AP}_b$, the satellite branch cannot compensate at inference, even though the per-pixel PTF gate already reallocates its evidence weight to satellite features.
This observation directly motivates the single shared object-query pool used by OCMQ: by routing all ground-truth instances through one pool and merging per-modality updates only via the per-token gate of Eq.~\eqref{eq:gated_collapse}, every class receives gradients from all three modalities, and no class is left unsupervised when its dominant modality is degraded.

\FloatBarrier
\subsection{Why Non-Temporal Outperforms Temporal under Severe Camera Failure}
\label{app:temporal_robustness}

The robustness results in Tab.~\ref{tab:robustness_performance_new} show that the non-temporal variant of our model outperforms the temporal one under severe camera dropout, even though the temporal variant is stronger on clean inputs. To diagnose this gap, we run both variants on the entire validation set of the geographically non-overlapping nuScenes split under three input regimes (clean, front-1 drop, all-camera drop). All BEV-pixel statistics are restricted to the per-frame active cells $\mathcal{A}_t = \{x : r_{\mathrm{cam}}^{\mathrm{clean}}(x,t) \geq 0.35\}$ (Fig.~\ref{fig:fallback_qualitative_v3}(a) convention), so background regions where the segmentation head saturates near its prior do not dilute the signal. Within $\mathcal{A}_t$ we probe four intermediate signals: the two PTF reliability scores $r_{\mathrm{cam}}, r_{\mathrm{sat}}$ defined in Eq.~\eqref{eq:reliability}, the L2 norm of the fused BEV $\|\mathbf{F}_{\mathrm{fuse}}\|$ over channels, and an auxiliary probe $r_{\mathrm{fuse}}$ that is not part of PTF: we feed $\mathbf{F}_{\mathrm{fuse}}$ through the already-trained $\mathrm{Seg}_{\mathrm{fuse}}$ head and take its max-class sigmoid as a per-pixel indicator of how informative the fused BEV remains. Tab.~\ref{tab:temporal_robustness_alpha} summarizes the result.

\input{tables/temporal_robustness_alpha}

\textbf{Camera-stage signal: $r_{\mathrm{cam}}$ collapses far more in the temporal model.} Under all-camera drop, the temporal $r_{\mathrm{cam}}$ falls from $0.751$ to $0.166$ ($-77.9\%$ relative), while the non-temporal $r_{\mathrm{cam}}$ falls from $0.686$ to $0.278$ ($-59.5\%$). Both variants register the corruption, but the temporal one loses additional reliability because corrupted camera evidence enters the BEV memory queue and is fused with the current BEV through the encoder's memory-fusion blocks: not just the current frame's camera BEV but its entire short-term history is corrupted, lowering the camera segmentation confidence even on cells where map evidence should be present. $r_{\mathrm{sat}}$ is essentially identical across all six conditions ($0.473$ vs.\ $0.456$), confirming that the satellite branch behaves the same in both variants and that the differences below are not driven by satellite-side artifacts.

\textbf{Fuse-stage signal: the temporal model's fused stream degrades more.} The two columns describing the fused BEV, $r_{\mathrm{fuse}}$ and $\|\mathbf{F}_{\mathrm{fuse}}\|$, give a single, consistent reading. Under all-camera drop, the temporal $r_{\mathrm{fuse}}$ falls from $0.678$ to $0.437$ ($-35.6\%$), while the non-temporal $r_{\mathrm{fuse}}$ falls only from $0.545$ to $0.445$ ($-18.3\%$); on the same active cells, the temporal $\|\mathbf{F}_{\mathrm{fuse}}\|$ contracts by $-19.1\%$ ($11.59 \to 9.37$) versus only $-6.2\%$ for the non-temporal variant ($9.78 \to 9.17$). Relative to its own clean state, the temporal fused stream therefore loses roughly two to three times more on both probes, although after all-camera drop both variants end at similar absolute values ($0.437$ vs.\ $0.445$; $9.37$ vs.\ $9.17$): the temporal model relies more on camera evidence on clean inputs and loses that share when the cameras fail.

Together, these signals support the claim made in Sec.~\ref{sec:experiments}: temporal modeling, applied only to the camera BEV, lets corrupted camera evidence accumulate in the BEV history. This is most visible at the camera stage, where the temporal $r_{\mathrm{cam}}$ ends below the non-temporal one under both Front-1 and all-camera drop ($0.497$ vs.\ $0.553$; $0.166$ vs.\ $0.278$) despite starting above it on clean inputs. Because the temporal model also relies more on camera evidence, its fused stream loses more, and the gate in Eq.~\eqref{eq:gated_collapse} cannot restore camera evidence that $\mathbf{F}_{\mathrm{fuse}}$ no longer carries. The non-temporal variant isolates the corruption to a single frame: $r_{\mathrm{fuse}}$ stays much closer to its clean value, and $\|\mathbf{F}_{\mathrm{fuse}}\|$ contracts only modestly, which is precisely the failure-decoupling behavior the PTF--OCMQ pipeline is designed to provide.

\FloatBarrier
\subsection{Additional Qualitative Results}
\label{app:qual}

We provide five additional qualitative comparisons spanning all four nuScenes
locations (Figs.~\ref{fig:qual_appendix_scene0767_4353}--\ref{fig:qual_appendix_scene0962_5338}). Following the visualization in Fig.~\ref{fig:qual},
each figure shows multi-view camera images and satellite images on the top, and
ground-truth / method predictions on the bottom. Compared with
Fig.~\ref{fig:qual}, we additionally include DAMap~\cite{damap}, a
Camera+LiDAR fusion-based model, as a baseline. The bottom rows report predicted
vector maps under three input settings: the standard $60{\times}30\,\mathrm{m}$
ROI, Front-3 drop, and the extended $100{\times}50\,\mathrm{m}$ ROI. The last
example highlights a representative failure case, where occlusion and visual
ambiguity in the satellite image caused by nearby buildings lead to missed
map elements in the extended-ROI setting.

\input{figures/qual_appendix_6frames/scene-0767_4353/qual_scene-0767_4353_6col}
\clearpage
\input{figures/qual_appendix_6frames/scene-1056_4e93423b/qual_scene-1056_4e93423b_6col}
\clearpage
\input{figures/qual_appendix_6frames/scene-0572_2771/qual_scene-0572_2771_6col}
\clearpage
\input{figures/qual_appendix_6frames/scene-0795_4690/qual_scene-0795_4690_6col}
\clearpage
\input{figures/qual_appendix_6frames/scene-0962_5338/qual_scene-0962_5338_6col}
\clearpage

%% file: tables/mapping_performance_old.tex
\begin{table}[!htbp]
\centering
\normalsize

\caption{Online HD map construction performance on the original split of nuScenes, $60\,\mathrm{m}\times30\,\mathrm{m}$.
\emph{Mod.}: C, L, SD, HD, SA denote Camera, LiDAR, SD map, prior HD map, and Satellite images;
\emph{Temp.}: temporal memory or historical-prediction priors.
$^{*}$~from the original paper;
$^{\dagger}$~Consistent-GT supervision~\cite{maptracker}.
PriorDrive uses an SD map together with a noisy prior HD map and online local maps from earlier traversals.}

\label{tab:mapping_performance_old}
\setlength{\tabcolsep}{4pt}
\renewcommand{\arraystretch}{1.0}
\begin{tabular*}{\textwidth}{@{\extracolsep{\fill}} l c c c c c c c c}
\toprule
Method & Venue & Mod. & Temp. & Ep. & AP$_p$ & AP$_d$ & AP$_b$ & mAP \\
\midrule

InteractionMap$^{*}$~\cite{interactionmap} & CVPR'25     & C       & \checkmark  & 110 & 75.2 & 75.0 & 76.1 & 75.5 \\
MapTracker$^{\dagger}$~\cite{maptracker}   & ECCV'24     & C       & \checkmark  & 70  & 80.0 & 74.1 & 74.1 & 76.1 \\

\cmidrule(lr){1-9}
MapTRv2~\cite{maptrv2}                     & IJCV'24     & C+L     & $\times$    & 24  & 65.6 & 66.5 & 74.8 & 69.0 \\
Mask2Map$^{*}$~\cite{mask2map}             & ECCV'24     & C+L     & $\times$    & 24  & 76.5 & 76.6 & \textbf{82.1} & 78.4 \\
DAMap~\cite{damap}                         & ICCV'25     & C+L     & $\times$    & 24  & 67.7 & 72.4 & 77.1 & 72.4 \\

\cmidrule(lr){1-9}
SDTagNet~\cite{sdtagnet}                   & NeurIPS'25  & C+SD    & $\times$    & 30  & 61.5 & 63.1 & 61.4 & 62.0 \\
PriorDrive$^{*}$~\cite{priordrive}         & AAAI'26     & C+SD+HD & \checkmark  & 110 & 78.0 & \textbf{82.0} & 81.6 & 80.5 \\

\cmidrule(lr){1-9}
SatforHDMap~\cite{satforhdmap}             & ICRA'24     & C+SA    & $\times$    & 24  & 54.9 & 50.3 & 60.5 & 55.2 \\
SatMap$^{*}$~\cite{satmap}                 & arXiv'26    & C+SA    & $\times$    & 110 & 73.9 & 72.2 & 72.2 & 72.7 \\
SATMapTR$^{*}$~\cite{satmaptr}             & arXiv'25    & C+SA    & $\times$    & 24  & 73.9 & 72.3 & 75.1 & 73.8 \\

\cmidrule(lr){1-9}
\textbf{ReSMap (Ours)}                     & -           & C+SA    & $\times$    & 24  & 91.2 & 78.8             & 78.5             & 82.8 \\
\textbf{ReSMap (Ours)}                     & -           & C+SA    & \checkmark  & 24  & \textbf{91.5}    & 81.5 & 81.8 & \textbf{84.9} \\
\bottomrule
\end{tabular*}
\end{table}

%% file: tables/av2_performance.tex
\begin{table}[!htbp]
\centering
\caption{Results on the geographically non-overlapping split of Argoverse~2~\cite{argoverse2}, at $60\,\mathrm{m}\times30\,\mathrm{m}$, with satellite tiles from Google Maps. Left: clean inputs. Right: camera dropout, where \emph{Front-1}, \emph{Front-3}, and \emph{All} zero out the front-center camera, the three front cameras, and all seven ring cameras at inference. Each drop cell shows mAP and the relative change from Clean in red. $^{\dagger}$~Consistent-GT supervision~\cite{maptracker}.}
\label{tab:av2}
\setlength{\tabcolsep}{3.8pt}
\renewcommand{\arraystretch}{1.0}
\begin{tabular}{l c c c c c c c c c}
\toprule
& & & & & & & \multicolumn{3}{c}{Camera Drop} \\
\cmidrule(lr){8-10}
Method & Mod. & Temp. & AP$_p$ & AP$_d$ & AP$_b$ & Clean & Front-1 & Front-3 & All \\
\midrule
MapTracker$^{\dagger}$~\cite{maptracker} & C & \checkmark & 70.0 & 75.1 & 69.0 & 71.3 & 64.4 \drop{-9.7} & 30.0 \drop{-57.9} & 0.2 \drop{-99.7} \\
SDTagNet~\cite{sdtagnet} & C+SD & $\times$ & 64.3 & 70.8 & 68.4 & 67.8 & 64.6 \bdrop{-4.7} & 52.3 \drop{-22.9} & 18.6 \drop{-72.6} \\
SatforHDMap~\cite{satforhdmap} & C+SA & $\times$ & 17.6 & 40.5 & 37.3 & 31.8 & 30.0 \drop{-5.7} & 29.3 \bdrop{-7.9} & 21.0 \drop{-34.0} \\
\textbf{ReSMap (Ours)} & C+SA & $\times$ & 69.2 & 67.3 & 65.6 & 67.4 & 63.5 \drop{-5.8} & 57.1 \drop{-15.3} & \textbf{46.9} \bdrop{-30.4} \\
\textbf{ReSMap (Ours)} & C+SA & \checkmark & \textbf{77.3} & \textbf{75.8} & \textbf{74.3} & \textbf{75.8} & \textbf{71.2} \drop{-6.1} & \textbf{58.5} \drop{-22.8} & 30.9 \drop{-59.2} \\
\bottomrule
\end{tabular}
\end{table}

%% file: tables/robustness_performance_old.tex
\begin{table}[!htbp]
\centering
\caption{Robustness on the original split of nuScenes, at $60\,\mathrm{m}\times30\,\mathrm{m}$.
\emph{Front-1}, \emph{Front-3}, and \emph{All-camera} zero out the front camera, the
three front cameras, and all six cameras at inference, respectively. We also report
MapBench~\cite{mapbench} mRR and mCE (in \%), with MapTR~\cite{maptr} as the mCE reference following MapBench. Each camera-drop cell shows
mAP and the relative change from Clean,
$(\mathrm{Drop}-\mathrm{Clean})/\mathrm{Clean}\times100\%$, in red. SafeMap~\cite{safemap} only reports single-view drop.
\emph{Temp.}: temporal memory or historical-prediction priors.
$^{*}$~from the original paper;
$^{\dagger}$~Consistent-GT supervision~\cite{maptracker}.}
\label{tab:robustness_performance_old}
\footnotesize
\setlength{\tabcolsep}{3pt}
\renewcommand{\arraystretch}{1.05}
\resizebox{\textwidth}{!}{%
\begin{tabular}{l c c c c c c c c c}
\toprule
& & & & & \multicolumn{3}{c}{Camera Drop} & \multicolumn{2}{c}{Corruption} \\
\cmidrule(lr){6-8} \cmidrule(lr){9-10}
Method & Mod. & Temp. & Ep. & Clean & Front-1 & Front-3 & All & mRR $\uparrow$ & mCE $\downarrow$ \\
\midrule
MapTracker$^{\dagger}$~\cite{maptracker}  & C     & \checkmark  & 70 & 76.1 
& 41.1 {\tiny\textcolor{red}{(-46.0\%)}} 
& 17.0 {\tiny\textcolor{red}{(-77.7\%)}} 
& 0.1  {\tiny\textcolor{red}{(-99.9\%)}} 
& 69.1 & 61.2 \\
SafeMap$^{*}$~\cite{safemap}              & C     & $\times$    & 24 & 52.5 & 42.4 {\tiny\textcolor{red}{(-19.2\%)}} & -- & -- & 51.2 & 90.6 \\
\midrule
MapTRv2~\cite{maptrv2}                    & C+L   & $\times$    & 24 & 69.0 & 62.3 {\tiny\textcolor{red}{(-9.7\%)}} & 55.5 {\tiny\textcolor{red}{(-19.6\%)}} & 41.3 {\tiny\textcolor{red}{(-40.1\%)}} & 80.7 & 55.4 \\
DAMap~\cite{damap}                        & C+L   & $\times$    & 24 & 72.4 & 62.4 {\tiny\textcolor{red}{(-13.8\%)}} & 54.8 {\tiny\textcolor{red}{(-24.3\%)}} & 35.3 {\tiny\textcolor{red}{(-51.2\%)}} & 79.3 & 55.8 \\
\midrule
SDTagNet~\cite{sdtagnet}                  & C+SD  & $\times$    & 30 & 62.0 & 47.1 {\tiny\textcolor{red}{(-24.0\%)}} & 38.7 {\tiny\textcolor{red}{(-37.6\%)}} & 14.5 {\tiny\textcolor{red}{(-76.6\%)}} & 68.6 & 75.6 \\
\midrule
SatforHDMap~\cite{satforhdmap}            & C+SA  & $\times$    & 24 & 55.2 & 48.7 {\tiny\textcolor{red}{(-11.8\%)}} & 48.2 {\tiny\textcolor{red}{(-12.7\%)}} & 31.8 {\tiny\textcolor{red}{(-42.4\%)}} & 92.1 & 66.8 \\
\midrule
\textbf{ReSMap (Ours)} & C+SA & $\times$ & 24 

& 82.8 

& 81.2 {\tiny\textcolor{red}{(-1.9\%)}}

& \textbf{80.3 {\tiny\textcolor{red}{(-3.0\%)}} }

& \textbf{78.7 {\tiny\textcolor{red}{(-5.0\%)}} }

& \textbf{97.8} & 27.7 \\

\textbf{ReSMap (Ours)}                             & C+SA  & \checkmark  & 24 & 84.9 & \textbf{82.0} {\tiny\textcolor{red}{\textbf{(-3.4\%)}}} & 77.9 {\tiny\textcolor{red}{(-8.2\%)}} & 62.5 {\tiny\textcolor{red}{(-26.4\%)}} & 94.4 & \textbf{26.4} \\
\bottomrule
\end{tabular}%
}
\end{table}

%% file: tables/mapbench_geo.tex
\begin{table}[!htbp]
\centering
\caption{MapBench~\cite{mapbench} sensor-corruption robustness on the geographically non-overlapping split of nuScenes~\cite{streammapnet}, at $60\,\mathrm{m}\times30\,\mathrm{m}$: mean Resilience Rate (mRR, clean-normalized retention over the $24$ corruptions) and mean Corruption Error (mCE, relative to MapTracker). ReSMap is the temporal model. $^{\dagger}$~Consistent-GT supervision~\cite{maptracker}.}
\label{tab:mapbench_geo}
\small
\setlength{\tabcolsep}{6pt}
\begin{tabular}{l c c c}
\toprule
Method & Mod. & mRR $\uparrow$ & mCE $\downarrow$ \\
\midrule
MapTracker$^{\dagger}$~\cite{maptracker} & C & 65.7 & 1.000 \\
\midrule
MapTRv2~\cite{maptrv2} & C+L & 87.3 & 0.927 \\
DAMap~\cite{damap} & C+L & 85.5 & 0.920 \\
\midrule
SDTagNet~\cite{sdtagnet} & C+SD & 66.9 & 1.132 \\
\midrule
SatforHDMap~\cite{satforhdmap} & C+SA & 90.4 & 1.052 \\
\midrule
\textbf{ReSMap (Ours)} & C+SA & \textbf{90.8} & \textbf{0.712} \\
\bottomrule
\end{tabular}
\end{table}

%% file: tables/satellite_image_noise_performance.tex
\begin{table}[!htbp]
\centering
\normalsize
\caption{Robustness to satellite image corruption on the geographically
non-overlapping split of nuScenes~\cite{streammapnet}, at
$60\,\mathrm{m}\times30\,\mathrm{m}$, with corruption probability $p{=}0.2$.
Temporal swap: feeding the $t{\pm}1$ tile in place of the $t$-frame tile.
Geometric noise: shifting the tile by a Gaussian offset
($\sigma_t{=}1\text{m}$) and/or rotating it
($\sigma_r{=}5^\circ$); for each corrupted sample we randomly pick one of
\{shift, rotate, both\}.
Each cell shows mAP and the relative change from Clean,
$(\mathrm{Corrupt}-\mathrm{Clean})/\mathrm{Clean}\times100\%$, in red.}
\label{tab:satmap_corruption_robust}
\setlength{\tabcolsep}{4pt}
\renewcommand{\arraystretch}{1.0}
\begin{tabular}{l c c c c c c}
\toprule
& & & & & \multicolumn{2}{c}{Satellite Corruption mAP} \\
\cmidrule(lr){6-7}
Method & Mod. & Temp. & Ep. & Clean & Temporal swap & Geometric noise \\
\midrule
\textbf{ReSMap (Ours)} & C+SA & $\times$       & 24 & 47.3          & 43.1 \drop{-8.9} & 43.4 \drop{-8.2} \\
\textbf{ReSMap (Ours)} & C+SA & \checkmark     & 24 & \textbf{54.9} & \textbf{51.6} \bdrop{-6.0} & \textbf{51.9} \bdrop{-5.5} \\
\bottomrule
\end{tabular}
\vspace{-2mm}
\end{table}

%% file: tables/decoder_branch_perf.tex
\begin{table}[!htbp]
\centering
\caption{Per-branch prediction on clean inputs and final prediction under Front-3 drop (three front cameras zeroed).}
\label{tab:decoder_branch_perf}
\normalsize
\setlength{\tabcolsep}{6pt}
\renewcommand{\arraystretch}{1.2}
\begin{tabular}{p{0.30\linewidth} l c c c c}
\toprule
Setting & Branch / Input & AP$_p$ & AP$_d$ & AP$_b$ & mAP \\
\midrule
\multirow{3}{=}{\emph{(a) Per-branch prediction on clean inputs}}
& Camera    & 11.0          & \textbf{17.2} & \textbf{38.6} & 22.3          \\
& Satellite & 17.8          & 2.7           & 9.7           & 10.1          \\
& Fused     & \textbf{58.0} & 11.5          & 4.6           & \textbf{24.7} \\
\midrule
\multirow{2}{=}{\emph{(b) Final prediction under camera dropout}}
& Clean        & 67.8 & 28.1 & 51.0 & 49.0 \\
& Front-3 drop  
& \makecell{58.0{\tiny\textcolor{red}{(-14.5\%)}}}
& \makecell{19.2{\tiny\textcolor{red}{(-31.7\%)}}}
& \makecell{22.1{\tiny\textcolor{red}{(-56.7\%)}}}
& \makecell{33.1{\tiny\textcolor{red}{(-32.4\%)}}} \\
\bottomrule
\end{tabular}
\end{table}

%% file: tables/temporal_robustness_alpha.tex
\begin{table}[!htbp]
\centering
\caption{Diagnosing temporal vs.\ non-temporal robustness on the geographically non-overlapping split validation set (5981 frames). All BEV-pixel statistics are restricted to the active cells $\mathcal{A}_t = \{x : r_{\mathrm{cam}}^{\mathrm{clean}}(x,t) \geq 0.35\}$. $r_{\mathrm{cam}}, r_{\mathrm{sat}}$ follow Eq.~\eqref{eq:reliability}; $r_{\mathrm{fuse}}$ is an auxiliary probe ($\mathrm{Seg}_{\mathrm{fuse}}$ on $\mathbf{F}_{\mathrm{fuse}}$); $\|\mathbf{F}_{\mathrm{fuse}}\|$ is the channel-wise L2 norm of $\mathbf{F}_{\mathrm{fuse}}$. Parentheses are absolute changes from the clean row.}
\label{tab:temporal_robustness_alpha}
\small
\setlength{\tabcolsep}{8pt}
\renewcommand{\arraystretch}{1.2}
\begin{tabular}{@{}l l c c c c@{}}
\toprule
& Setting
& $r_{\mathrm{cam}}$ & $r_{\mathrm{sat}}$ & $r_{\mathrm{fuse}}$
& $\|\mathbf{F}_{\mathrm{fuse}}\|$ \\
\midrule
\multirow{3}{*}{Temporal}
    & Clean
    & 0.751 & 0.473 & 0.678 & 11.59 \\
    & Front-1 drop
    & 0.497~{\tiny\textcolor{red}{$(-0.254)$}}
    & 0.473
    & 0.575~{\tiny\textcolor{red}{$(-0.103)$}}
    & 10.67~{\tiny\textcolor{red}{$(-0.92)$}} \\
    & All-cam drop
    & \textbf{0.166}~{\tiny\textcolor{red}{$(-0.585)$}}
    & 0.473
    & \textbf{0.437}~{\tiny\textcolor{red}{$(-0.242)$}}
    & \textbf{9.37}~{\tiny\textcolor{red}{$(-2.21)$}} \\
\midrule
\multirow{3}{*}{Non-temporal}
    & Clean
    & 0.686 & 0.456 & 0.545 & 9.78 \\
    & Front-1 drop
    & 0.553~{\tiny\textcolor{red}{$(-0.133)$}}
    & 0.456
    & 0.507~{\tiny\textcolor{red}{$(-0.038)$}}
    & 9.56~{\tiny\textcolor{red}{$(-0.22)$}} \\
    & All-cam drop
    & 0.278~{\tiny\textcolor{red}{$(-0.408)$}}
    & 0.456
    & 0.445~{\tiny\textcolor{red}{$(-0.100)$}}
    & 9.17~{\tiny\textcolor{red}{$(-0.61)$}} \\
\bottomrule
\end{tabular}
\end{table}

%% file: figures/qual_appendix_6frames/scene-0767_4353/qual_scene-0767_4353_6col.tex
\begin{figure*}[!h]
\centering

\definecolor{clsped}{HTML}{1F77B4}
\definecolor{clsdiv}{HTML}{D62728}
\definecolor{clsbnd}{HTML}{2CA02C}

\newlength{\camhsixA}    \setlength{\camhsixA}{0.115\textwidth}
\newlength{\sathsixA}    \setlength{\sathsixA}{\dimexpr 2\camhsixA + 2pt\relax}
\newlength{\bevhsixA}    \setlength{\bevhsixA}{0.215\textwidth}
\newlength{\bevimgwsixA} \setlength{\bevimgwsixA}{0.105\textwidth}
\newlength{\bevcellwsixA}\setlength{\bevcellwsixA}{0.13\textwidth}

\setlength{\tabcolsep}{0pt}
\renewcommand{\arraystretch}{1.0}
\noindent
\begin{minipage}[c]{0.63\textwidth}
\centering
{\small\textsf{Multi-view Images}}\\[2pt]
\begin{tabular}{@{}c@{\hskip 2pt}c@{\hskip 2pt}c@{}}
\includegraphics[height=\camhsixA]{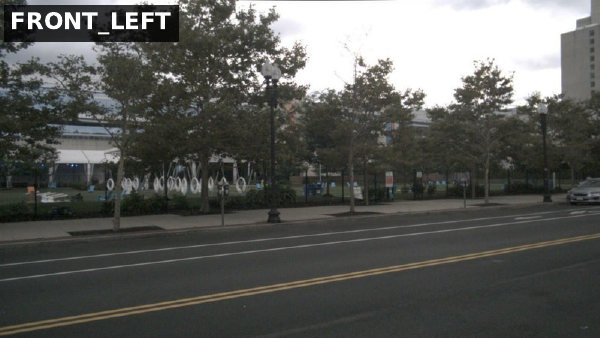} &
\includegraphics[height=\camhsixA]{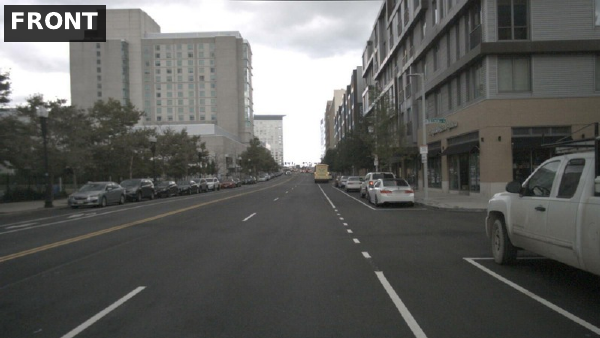} &
\includegraphics[height=\camhsixA]{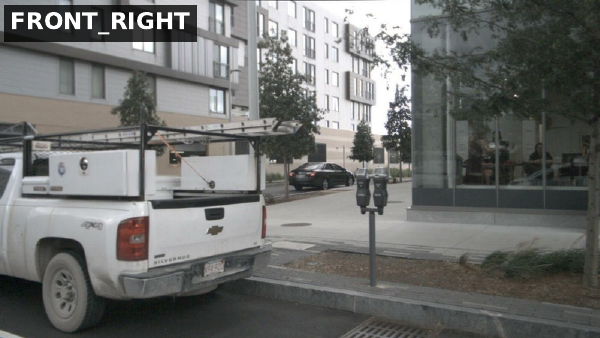} \\
\includegraphics[height=\camhsixA]{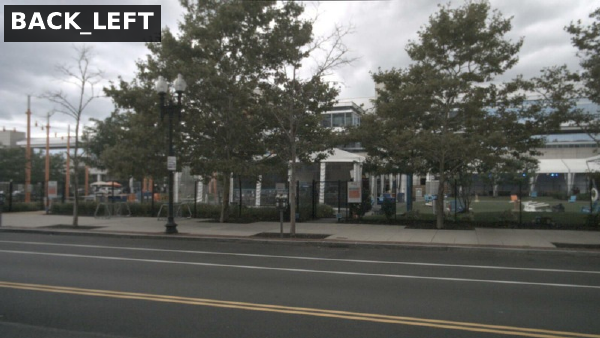} &
\includegraphics[height=\camhsixA]{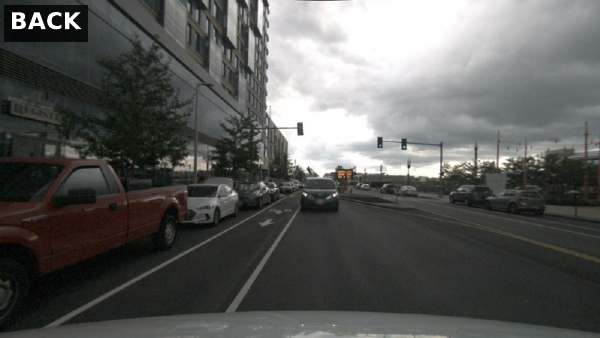} &
\includegraphics[height=\camhsixA]{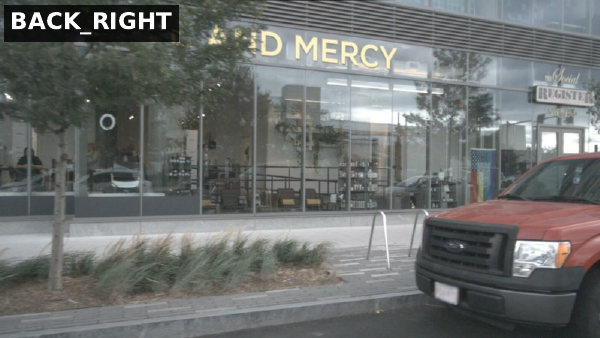} \\
\end{tabular}
\end{minipage}%
\hspace{4pt}%
\begin{minipage}[c]{0.25\textwidth}
\centering
{\small\textsf{Satellite Image}}\\[2pt]
\raisebox{4pt}{%
\mbox{%
\includegraphics[height=\sathsixA]{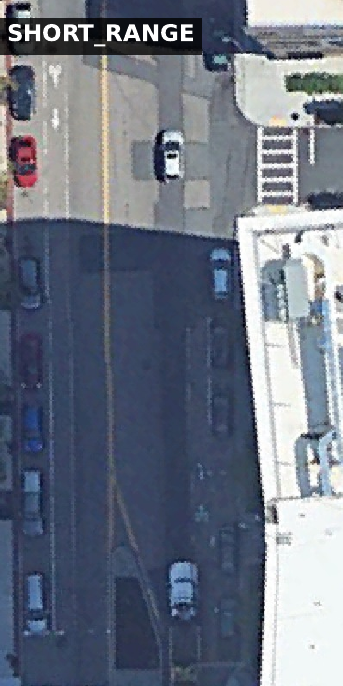}\hspace{3pt}%
\includegraphics[height=\sathsixA]{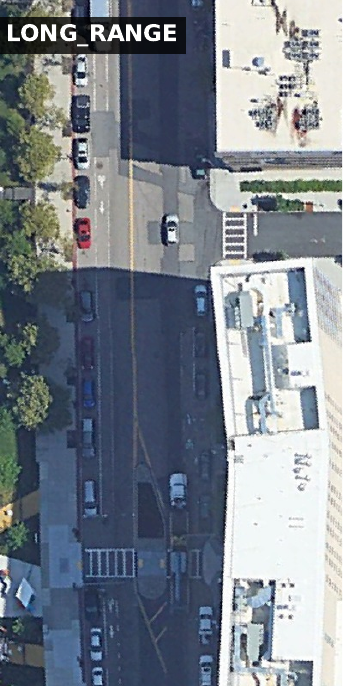}%
}
}
\end{minipage}

\vspace{8pt}

\newcommand{\rowlblsixA}[1]{\parbox[c][\bevhsixA][c]{4.5em}{\centering\footnotesize\mbox{#1}}}
\newcommand{\bevcellsixA}[1]{\parbox[c][\bevhsixA][c]{\bevcellwsixA}{\centering\fbox{\includegraphics[width=\bevimgwsixA]{#1}}}}

\setlength{\fboxsep}{0pt}
\setlength{\fboxrule}{0.4pt}
\setlength{\tabcolsep}{1pt}
\renewcommand{\arraystretch}{1.0}
\begin{tabular}{c@{\hskip 2pt}cccccc}
 & {\footnotesize\textsf{Ground Truth}} & {\footnotesize\textbf{\textsf{ReSMap(Ours)}}} & {\footnotesize\textsf{MapTracker}} & {\footnotesize\textsf{DAMap}} & {\footnotesize\textsf{SDTagNet}} & {\footnotesize\textsf{SatforHDMap}} \\
\rowlblsixA{60\,m$\times$30\,m} &
\bevcellsixA{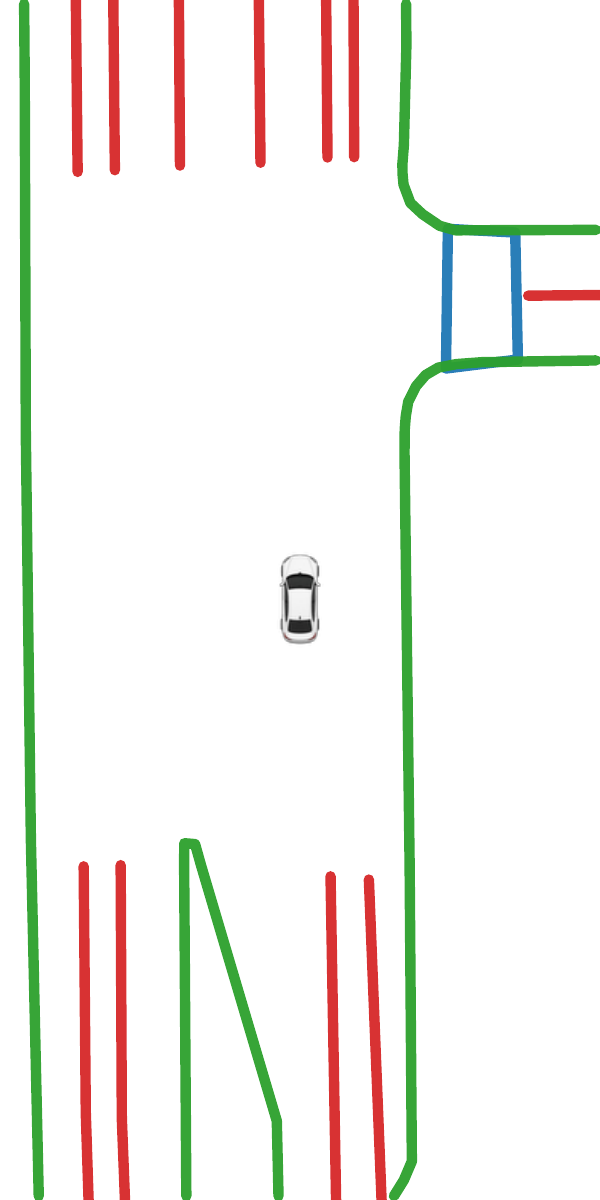} &
\bevcellsixA{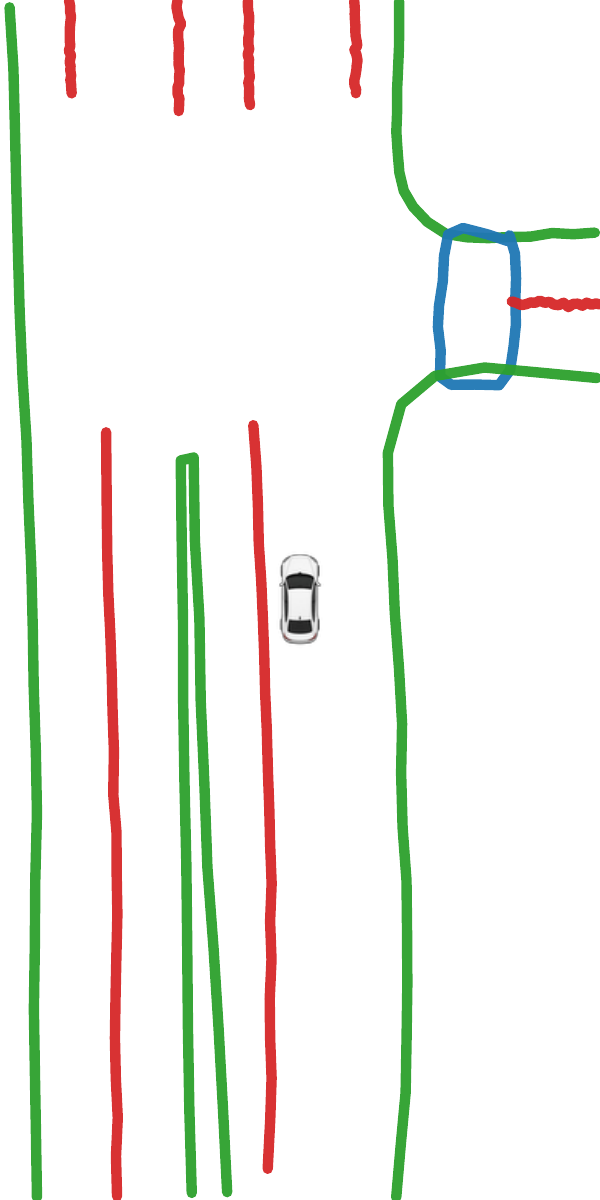} &
\bevcellsixA{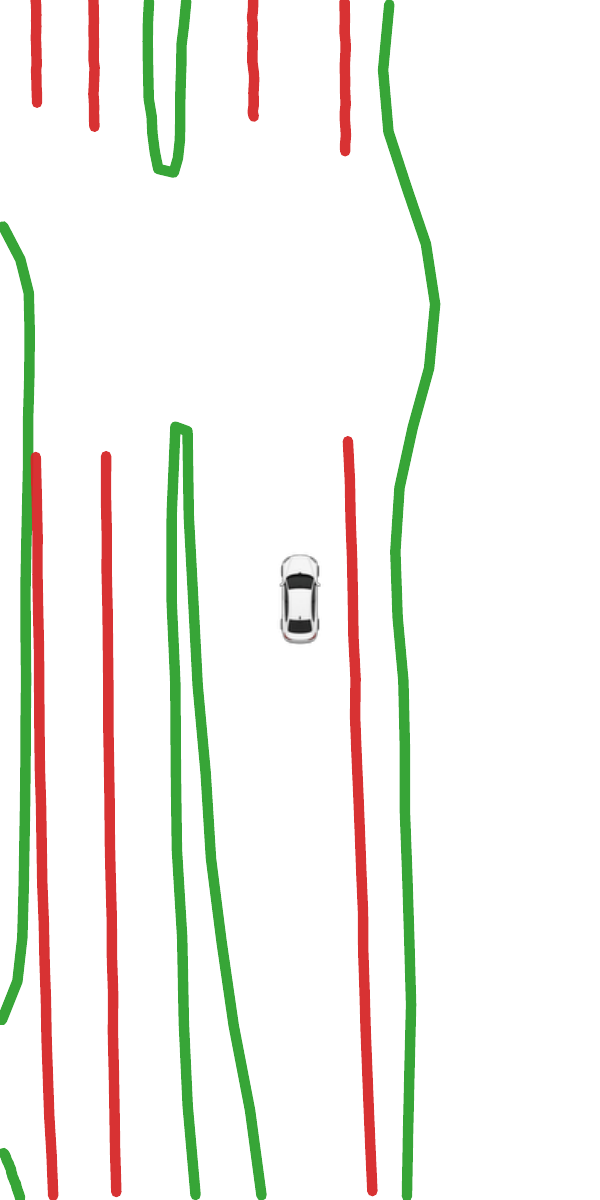} &
\bevcellsixA{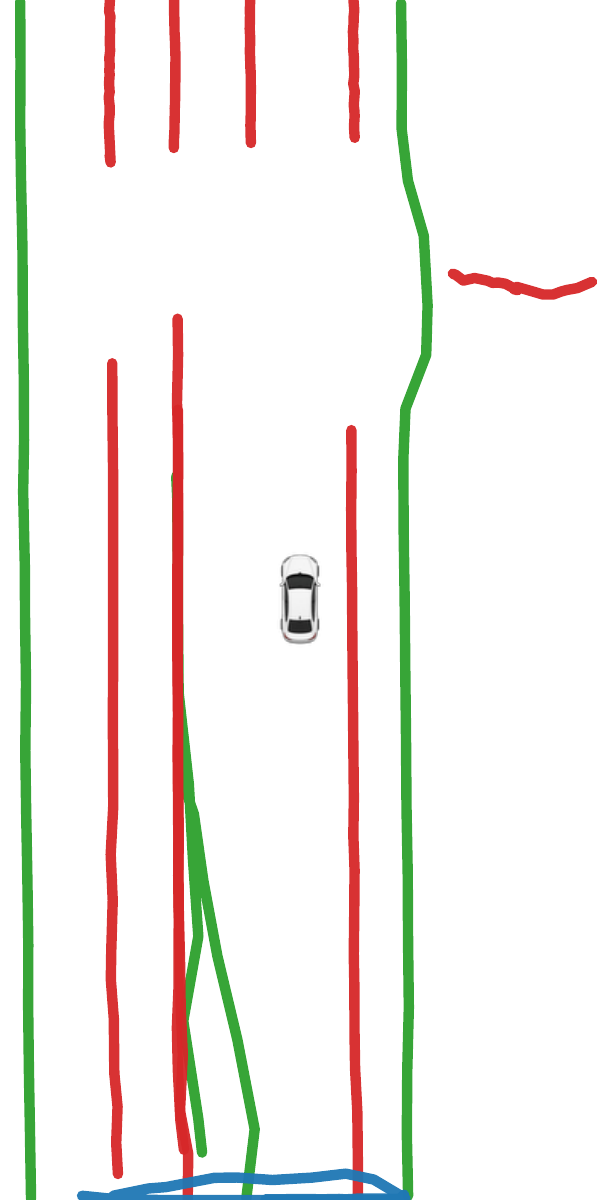} &
\bevcellsixA{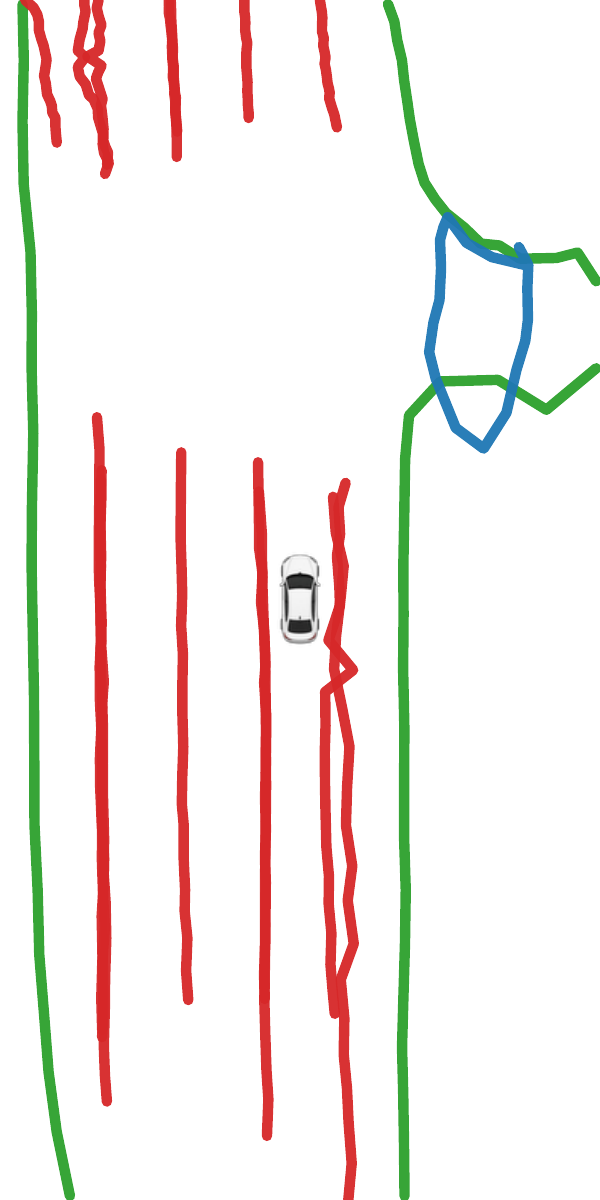} &
\bevcellsixA{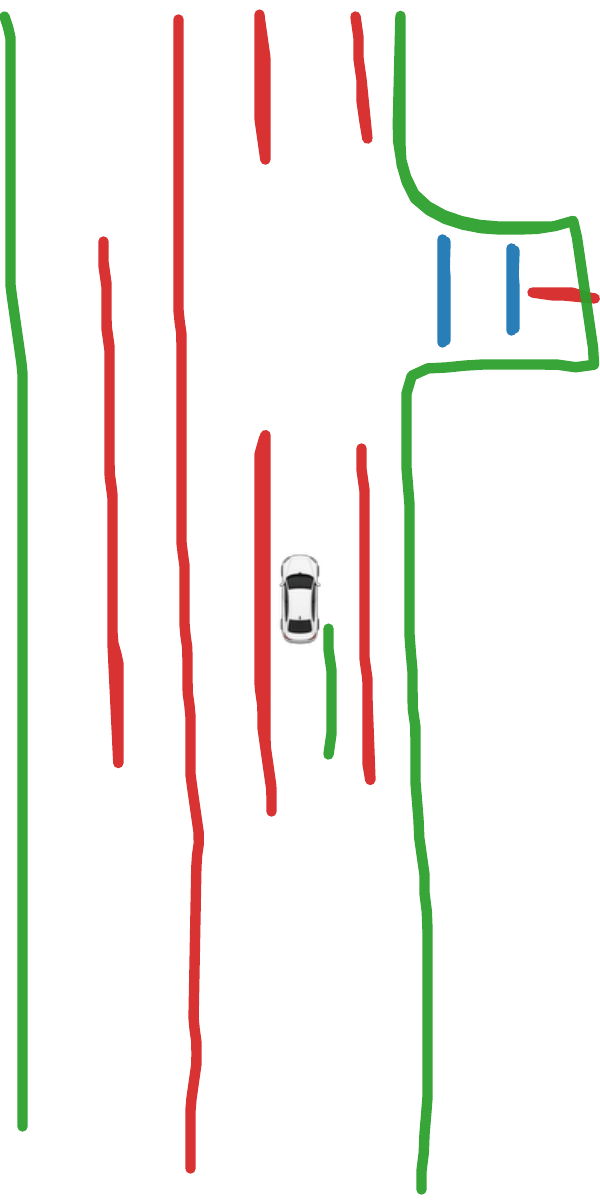} \\
\noalign{\vskip 2pt}
\rowlblsixA{Front-3 drop} &
\bevcellsixA{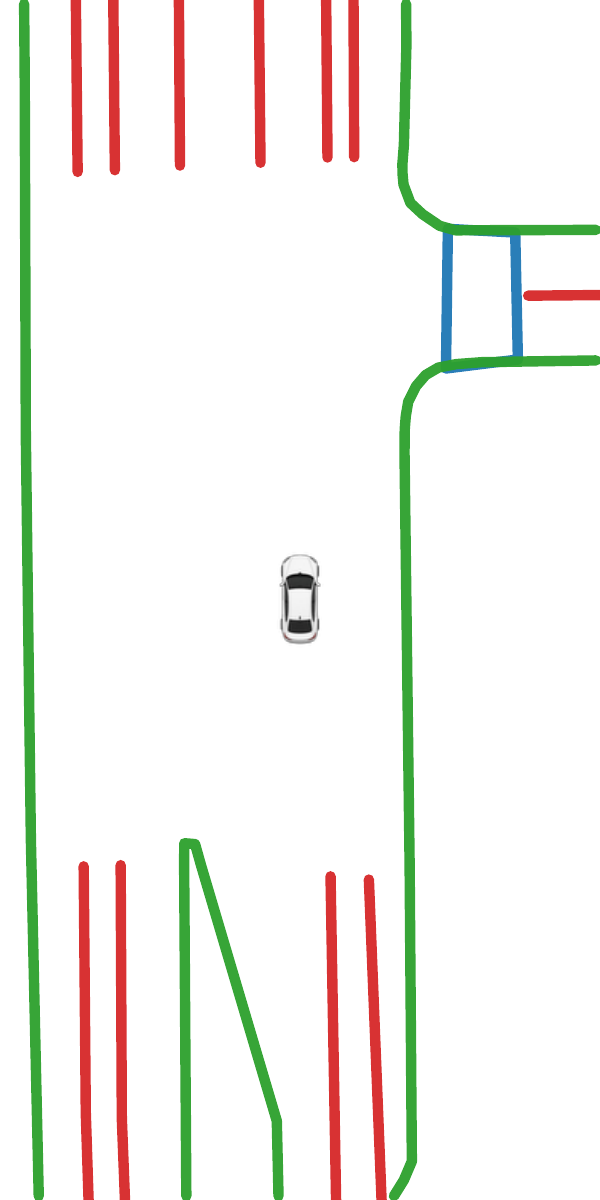} &
\bevcellsixA{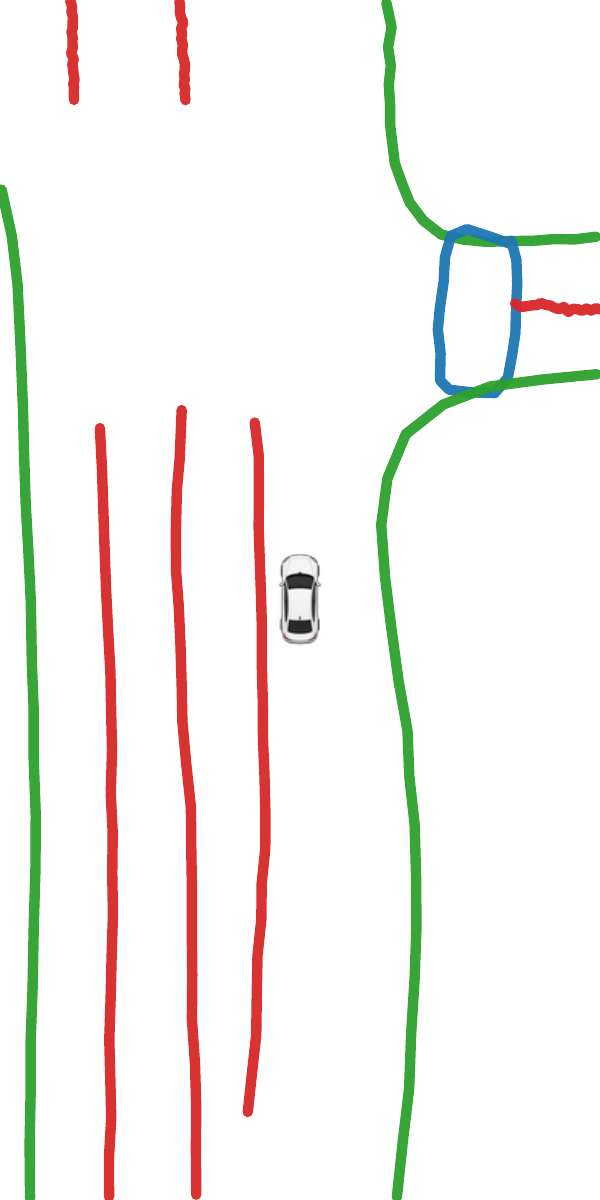} &
\bevcellsixA{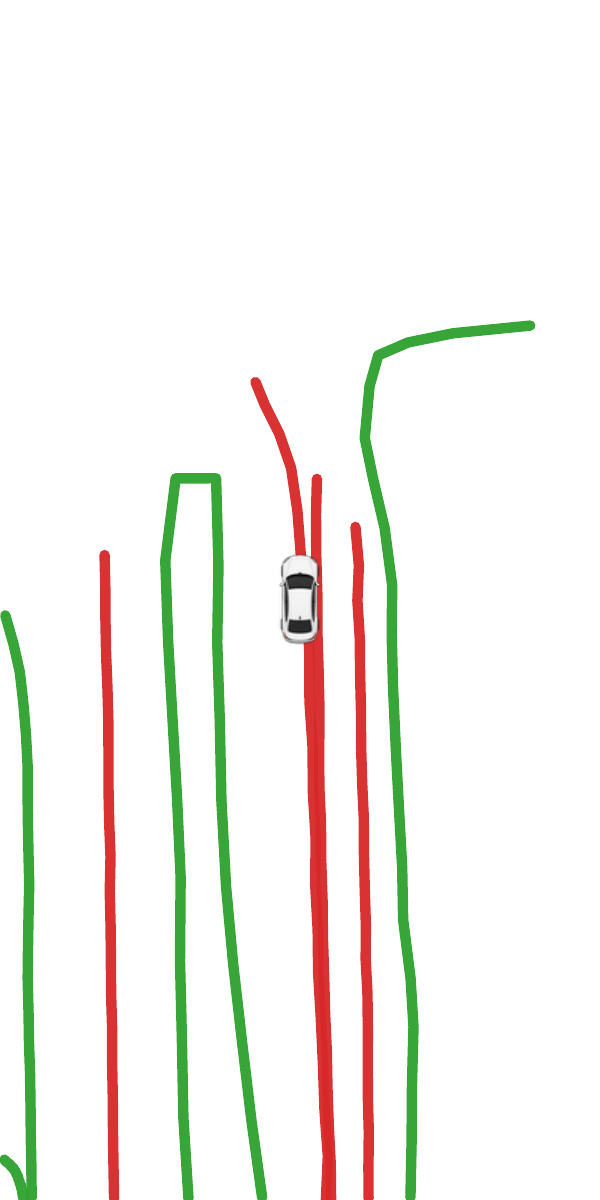} &
\bevcellsixA{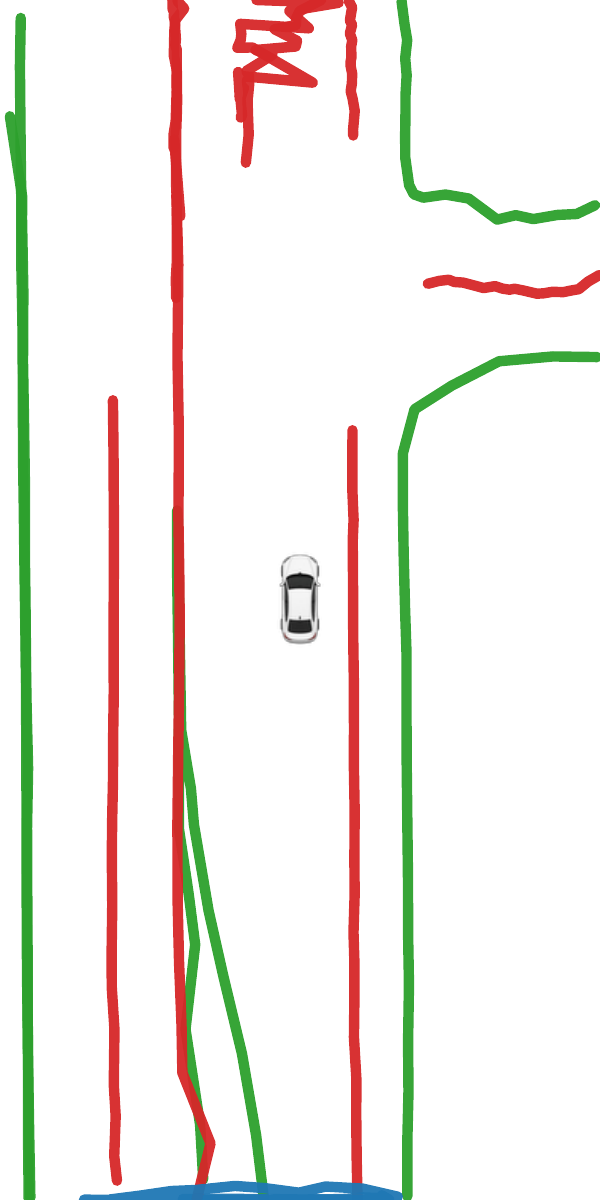} &
\bevcellsixA{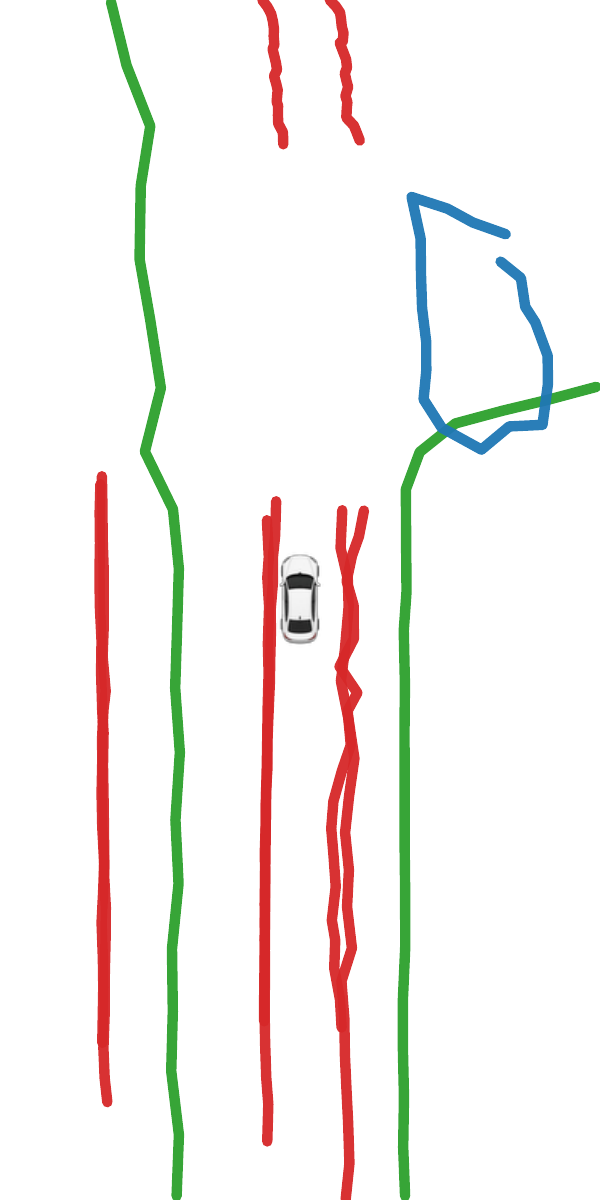} &
\bevcellsixA{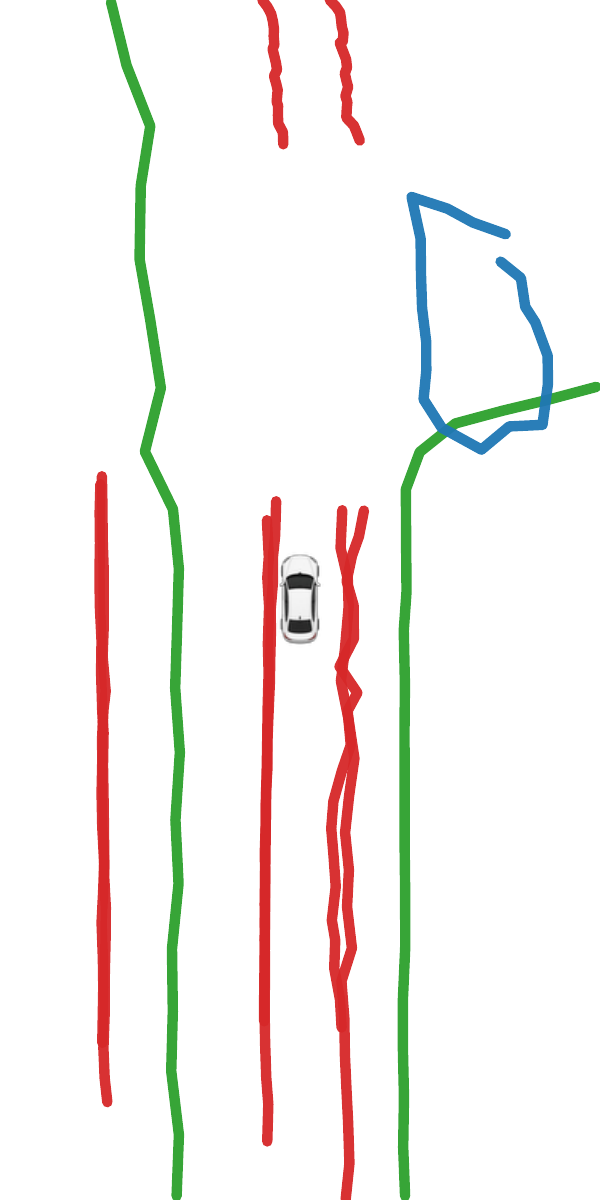} \\
\noalign{\vskip 2pt}
\rowlblsixA{100\,m$\times$50\,m} &
\bevcellsixA{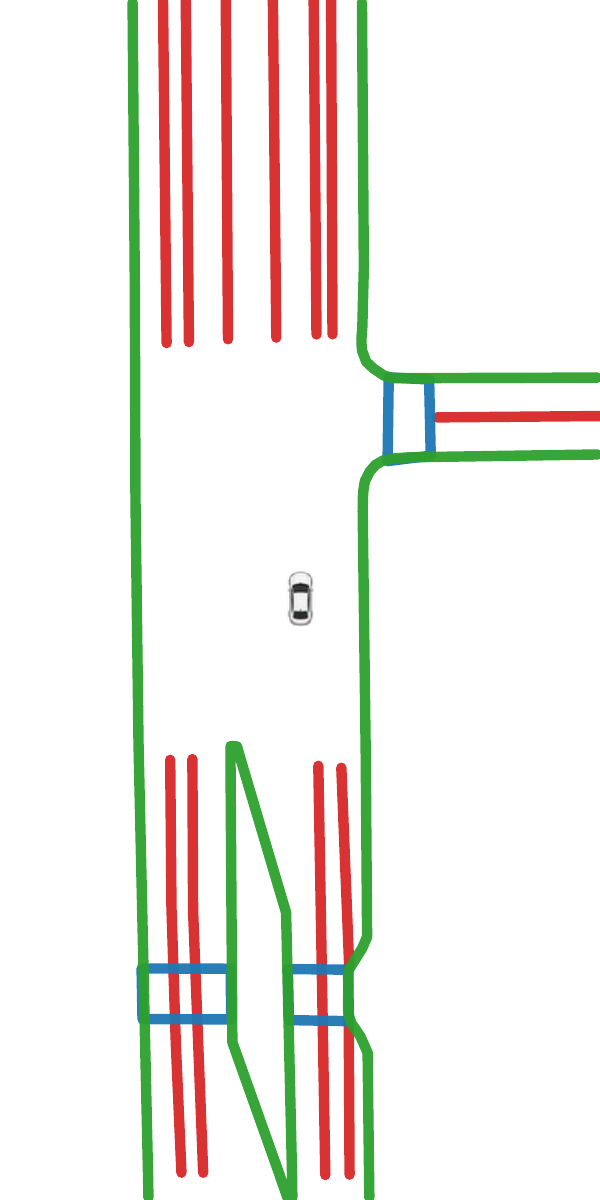} &
\bevcellsixA{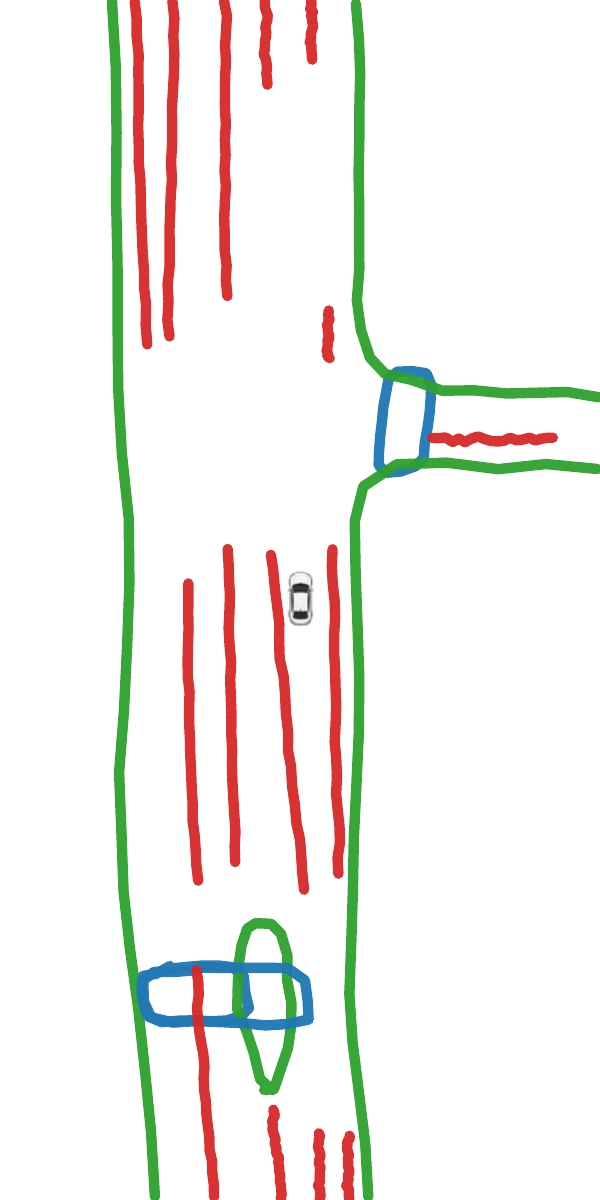} &
\bevcellsixA{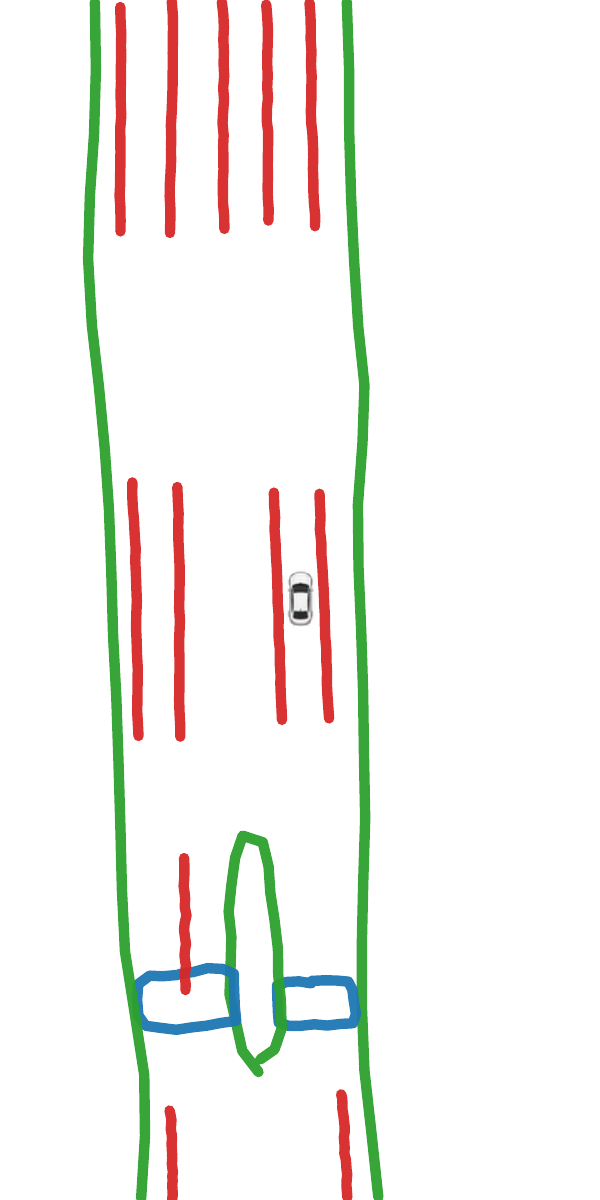} &
\bevcellsixA{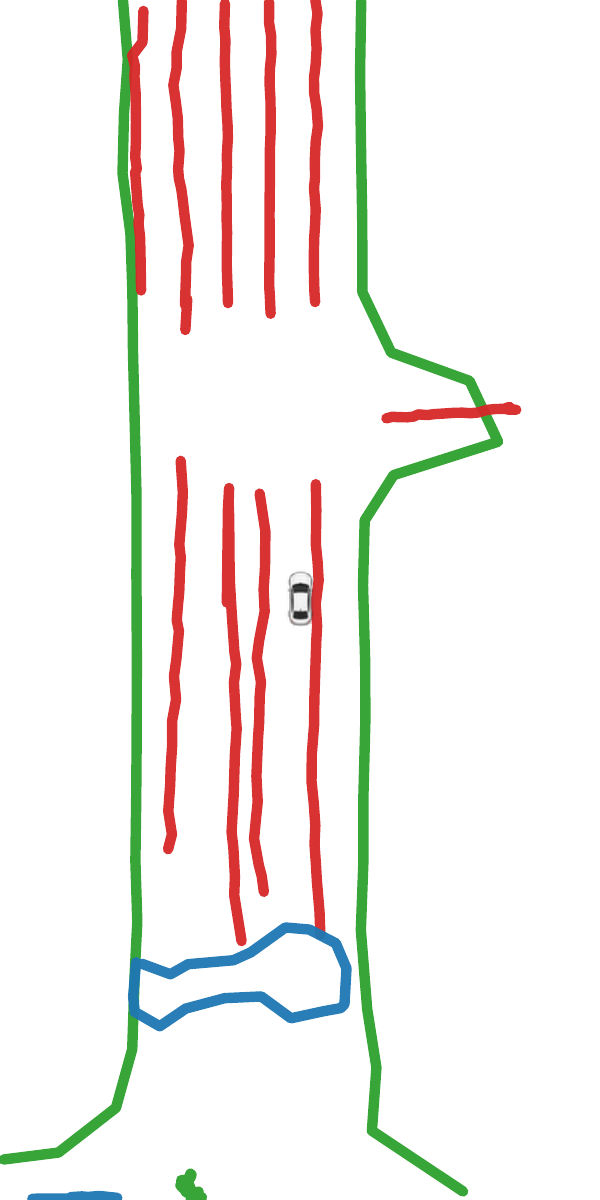} &
\bevcellsixA{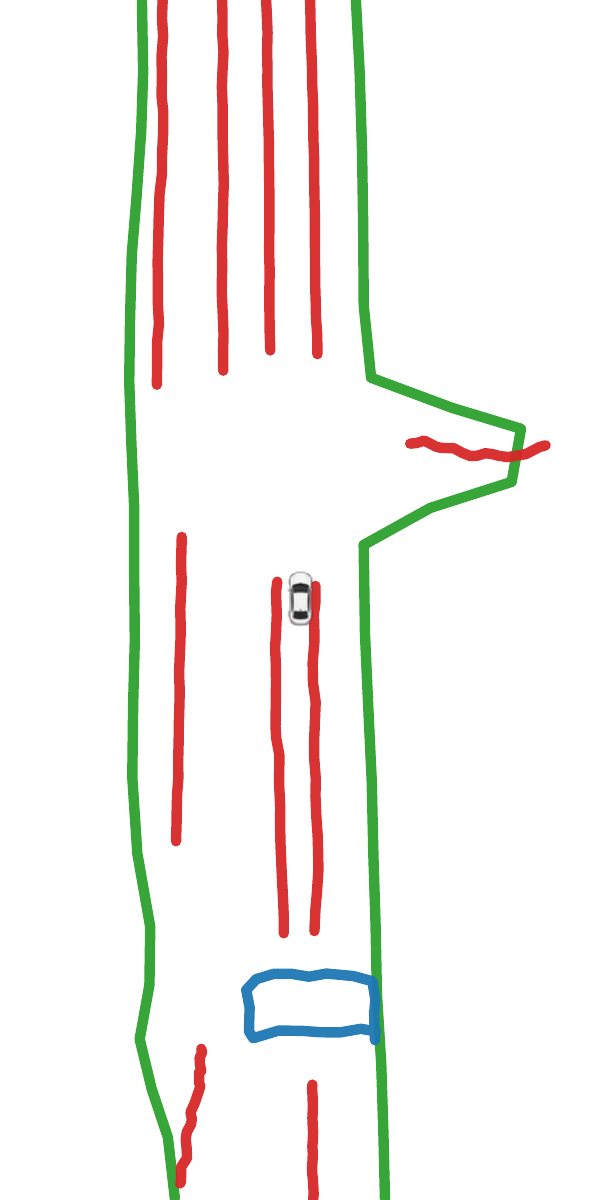} &
\bevcellsixA{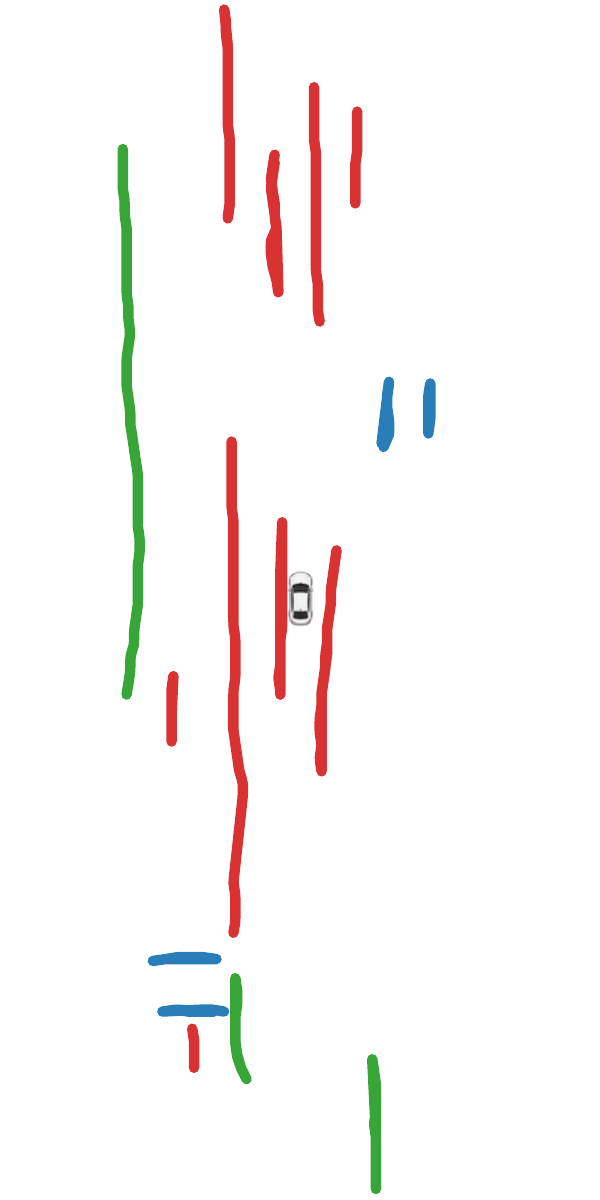} \\
\end{tabular}

\vspace{5pt}
{\footnotesize
\textcolor{clsped}{\rule[-1pt]{8pt}{8pt}}~ped.\,crossing\quad
\textcolor{clsdiv}{\rule[-1pt]{8pt}{8pt}}~divider\quad
\textcolor{clsbnd}{\rule[-1pt]{8pt}{8pt}}~boundary%
}

\caption{Additional qualitative comparison on boston-seaport.}
\label{fig:qual_appendix_scene0767_4353}
\end{figure*}

%% file: figures/qual_appendix_6frames/scene-1056_4e93423b/qual_scene-1056_4e93423b_6col.tex
\begin{figure*}[t]
\centering

\definecolor{clsped}{HTML}{1F77B4}
\definecolor{clsdiv}{HTML}{D62728}
\definecolor{clsbnd}{HTML}{2CA02C}

\newlength{\camhsixB}    \setlength{\camhsixB}{0.115\textwidth}
\newlength{\sathsixB}    \setlength{\sathsixB}{\dimexpr 2\camhsixB + 2pt\relax}
\newlength{\bevhsixB}    \setlength{\bevhsixB}{0.215\textwidth}
\newlength{\bevimgwsixB} \setlength{\bevimgwsixB}{0.105\textwidth}
\newlength{\bevcellwsixB}\setlength{\bevcellwsixB}{0.13\textwidth}

\setlength{\tabcolsep}{0pt}
\renewcommand{\arraystretch}{1.0}
\noindent
\begin{minipage}[c]{0.63\textwidth}
\centering
{\small\textsf{Multi-view Images}}\\[2pt]
\begin{tabular}{@{}c@{\hskip 2pt}c@{\hskip 2pt}c@{}}
\includegraphics[height=\camhsixB]{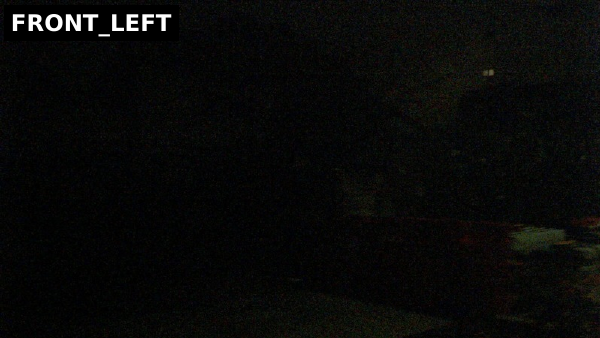} &
\includegraphics[height=\camhsixB]{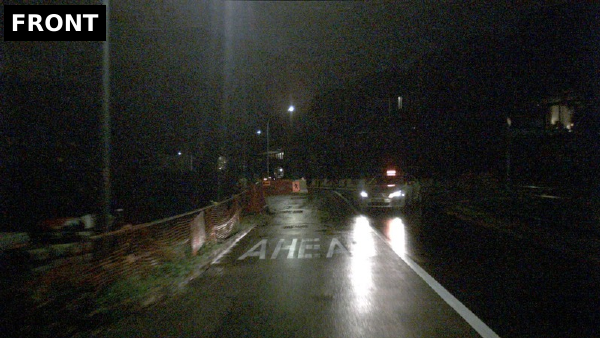} &
\includegraphics[height=\camhsixB]{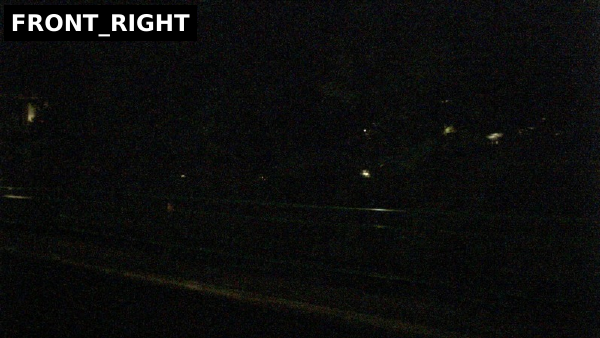} \\
\includegraphics[height=\camhsixB]{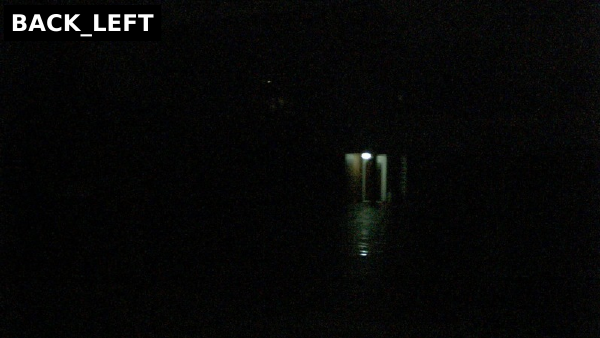} &
\includegraphics[height=\camhsixB]{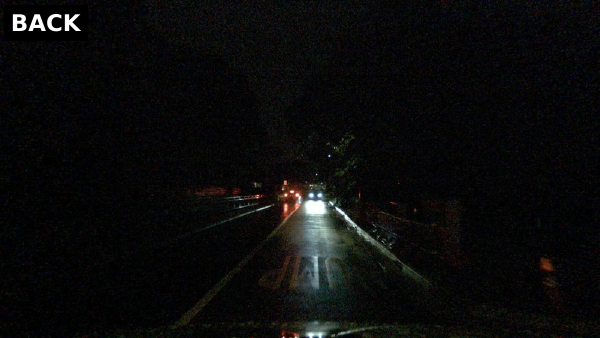} &
\includegraphics[height=\camhsixB]{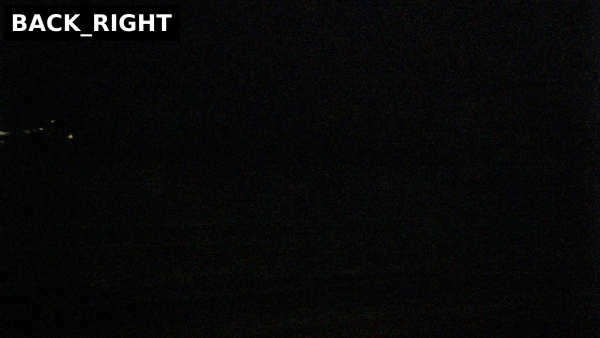} \\
\end{tabular}
\end{minipage}%
\hspace{4pt}%
\begin{minipage}[c]{0.25\textwidth}
\centering
{\small\textsf{Satellite Image}}\\[2pt]
\raisebox{4pt}{%
\mbox{%
\includegraphics[height=\sathsixB]{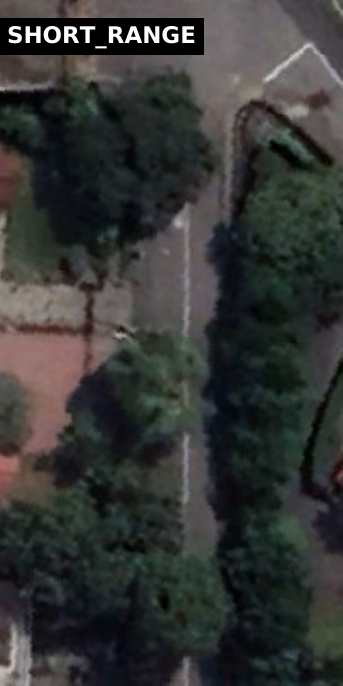}\hspace{3pt}%
\includegraphics[height=\sathsixB]{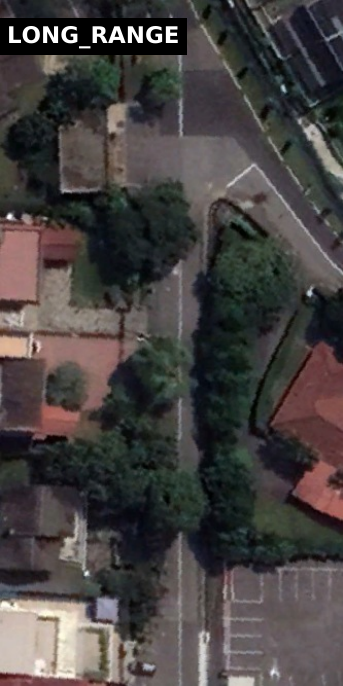}%
}
}
\end{minipage}

\vspace{8pt}

\newcommand{\rowlblsixB}[1]{\parbox[c][\bevhsixB][c]{4.5em}{\centering\footnotesize\mbox{#1}}}
\newcommand{\bevcellsixB}[1]{\parbox[c][\bevhsixB][c]{\bevcellwsixB}{\centering\fbox{\includegraphics[width=\bevimgwsixB]{#1}}}}

\setlength{\fboxsep}{0pt}
\setlength{\fboxrule}{0.4pt}
\setlength{\tabcolsep}{1pt}
\renewcommand{\arraystretch}{1.0}
\begin{tabular}{c@{\hskip 2pt}cccccc}
 & {\footnotesize\textsf{Ground Truth}} & {\footnotesize\textbf{\textsf{ReSMap(Ours)}}} & {\footnotesize\textsf{MapTracker}} & {\footnotesize\textsf{DAMap}} & {\footnotesize\textsf{SDTagNet}} & {\footnotesize\textsf{SatforHDMap}} \\
\rowlblsixB{60\,m$\times$30\,m} &
\bevcellsixB{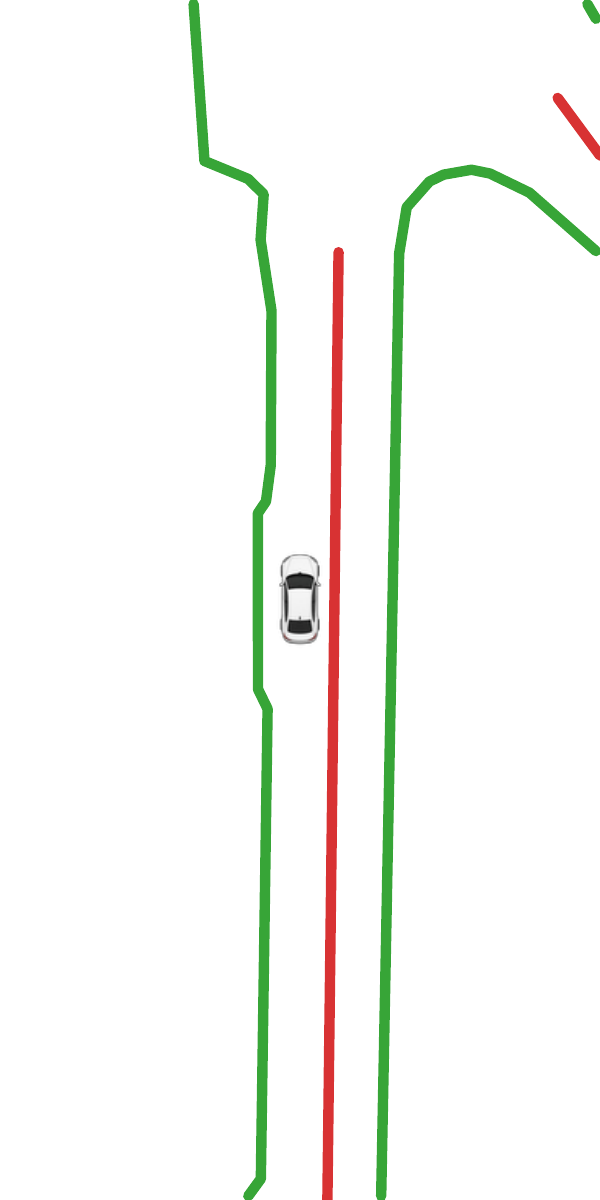} &
\bevcellsixB{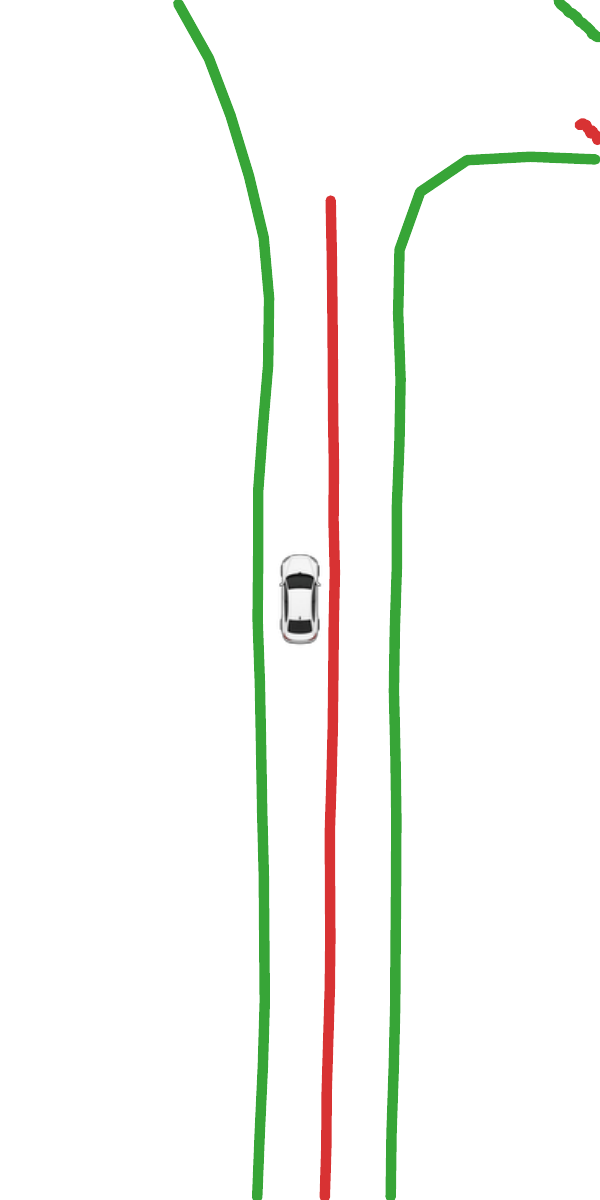} &
\bevcellsixB{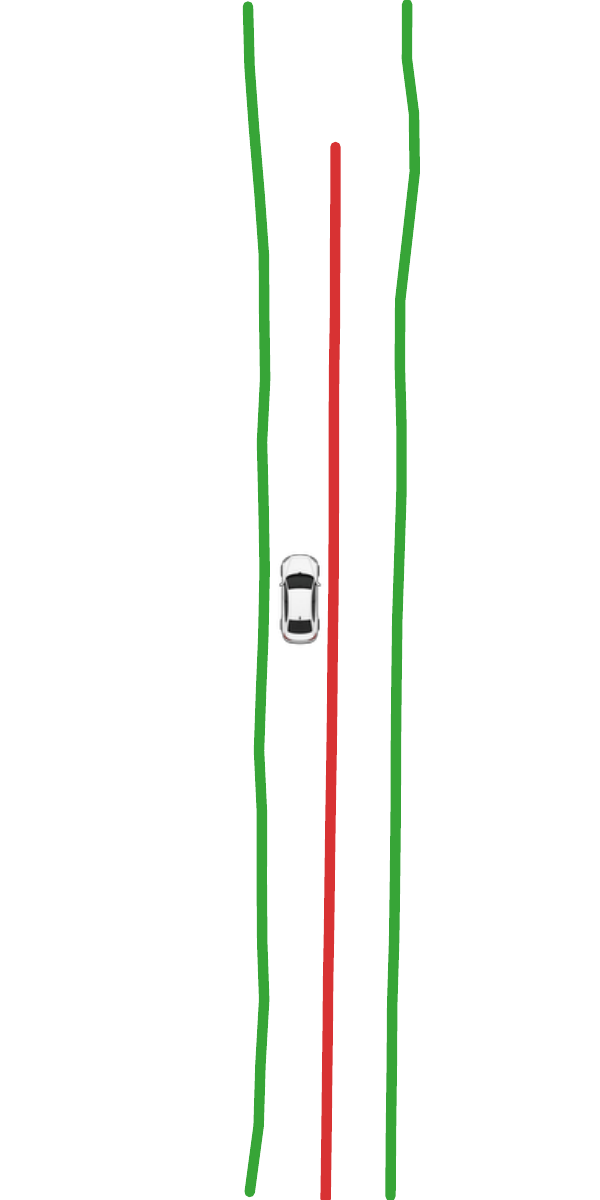} &
\bevcellsixB{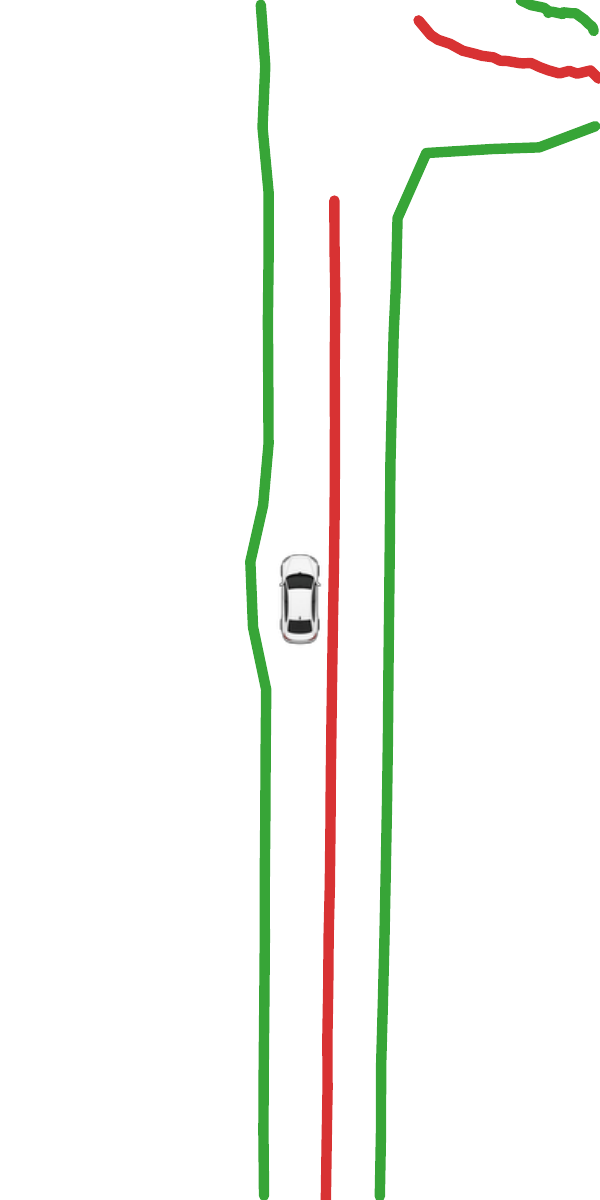} &
\bevcellsixB{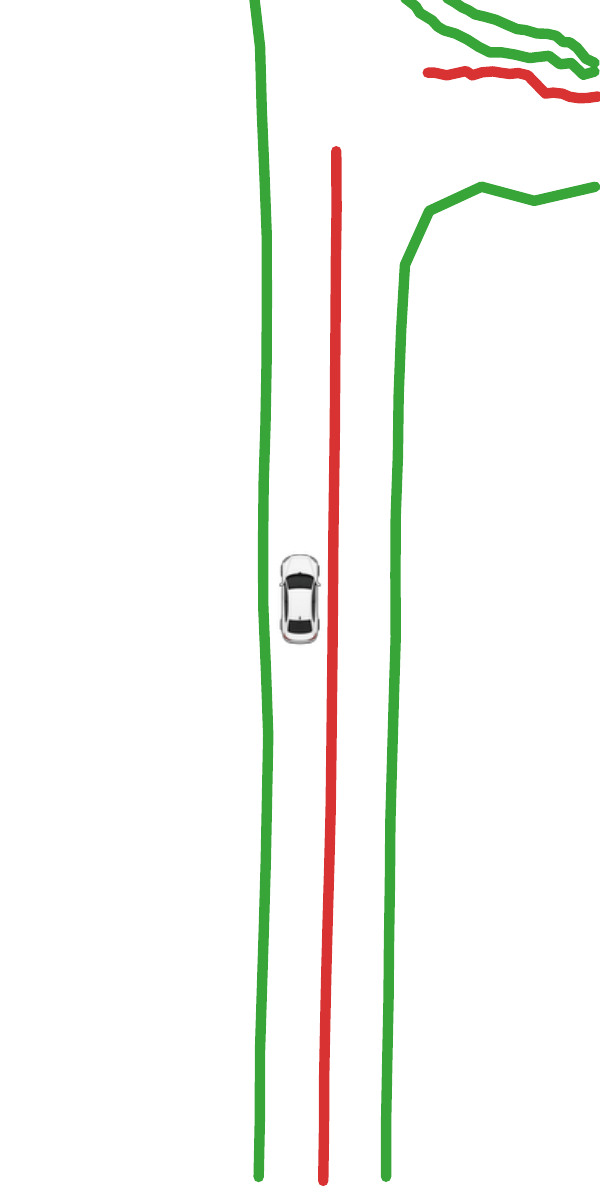} &
\bevcellsixB{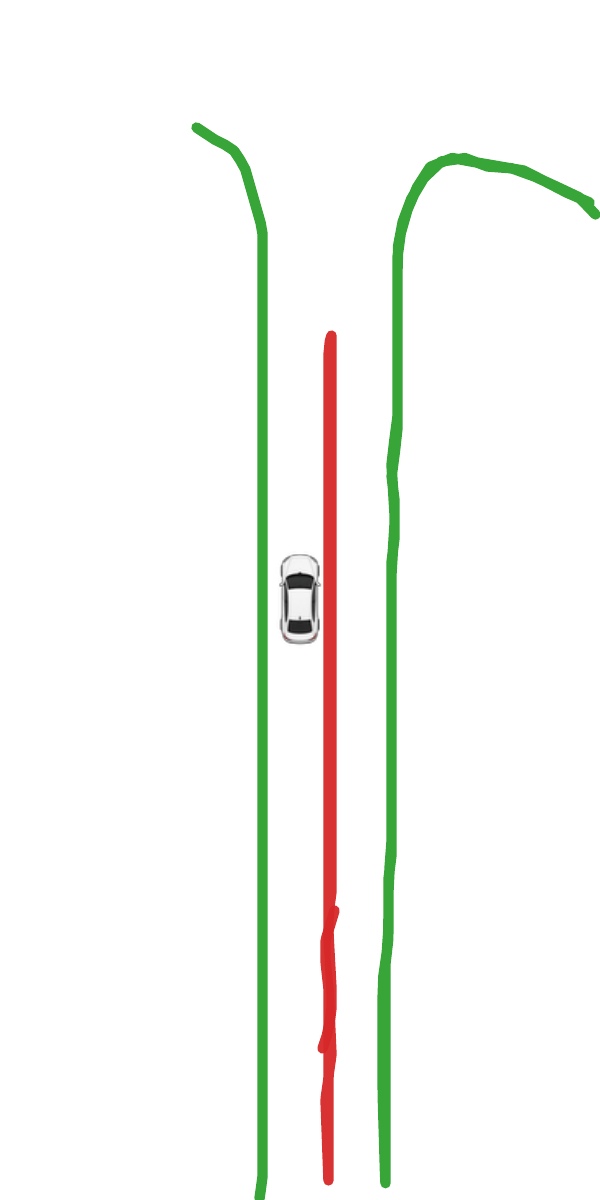} \\
\noalign{\vskip 2pt}
\rowlblsixB{Front-3 drop} &
\bevcellsixB{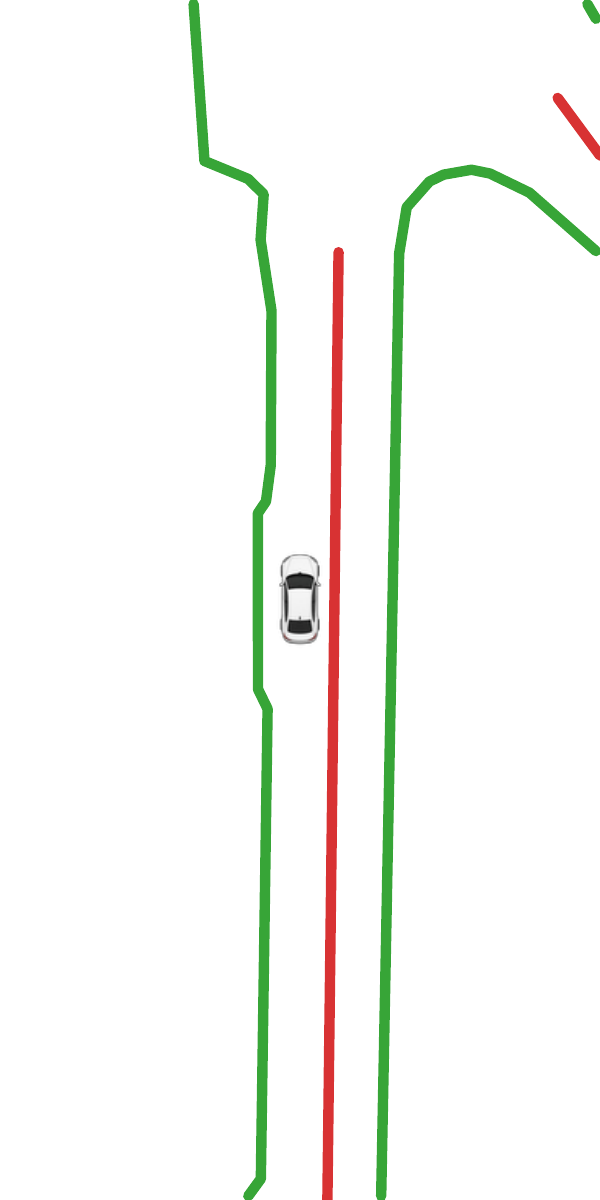} &
\bevcellsixB{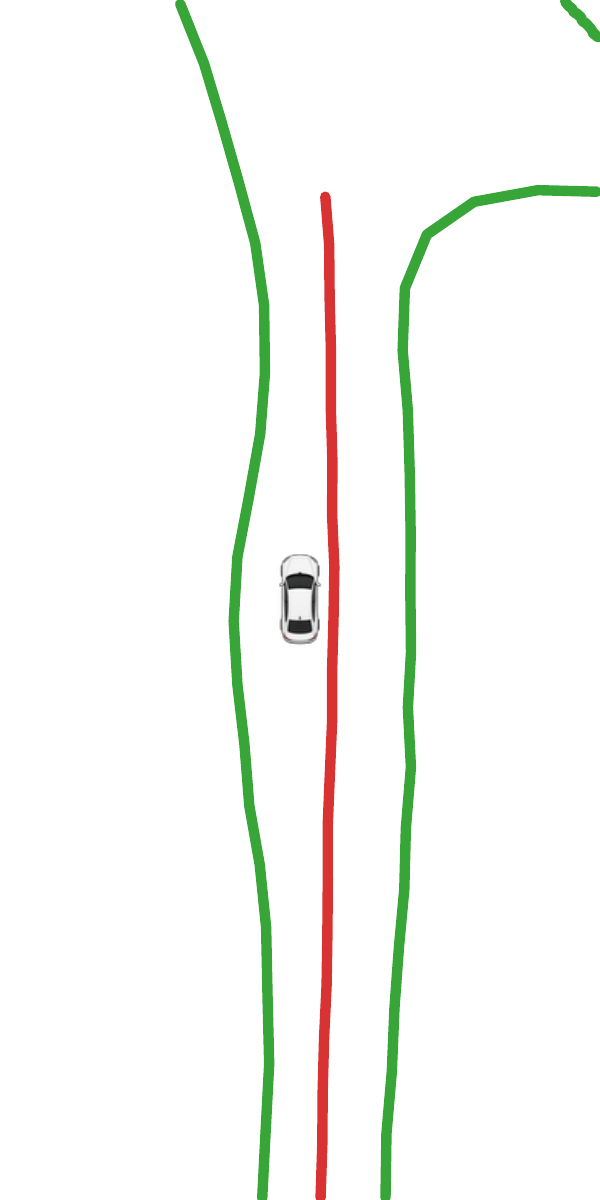} &
\bevcellsixB{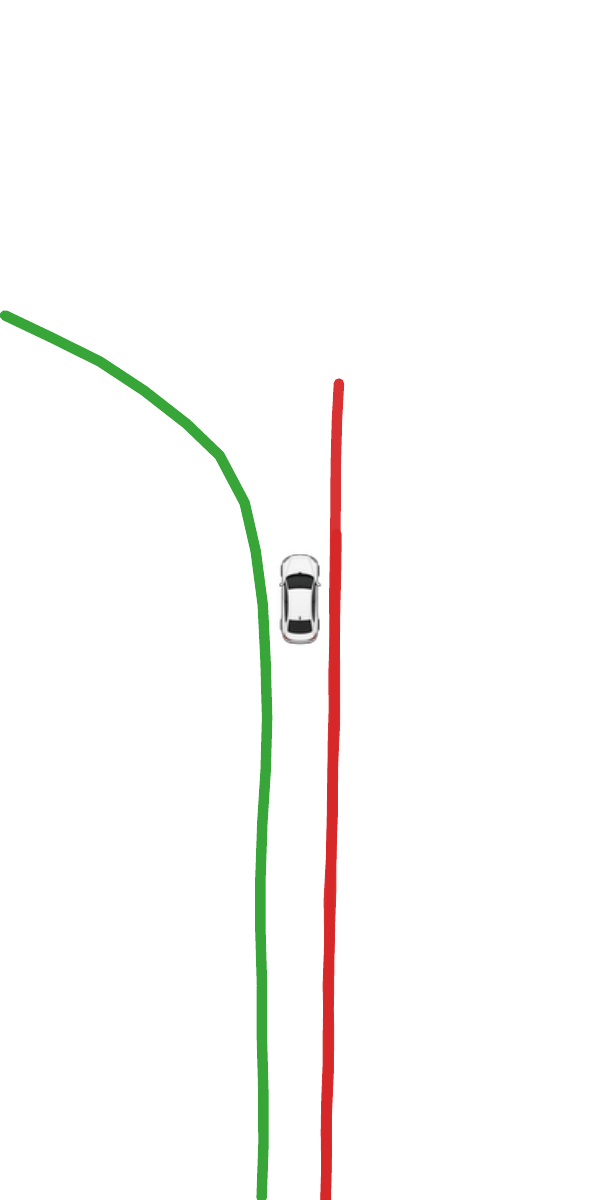} &
\bevcellsixB{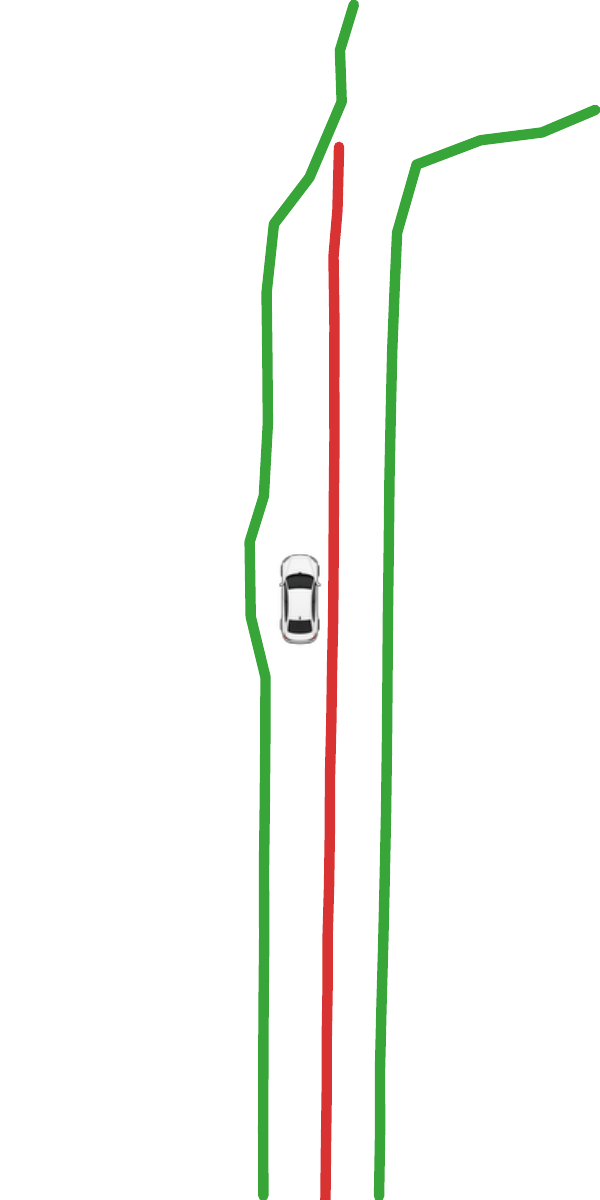} &
\bevcellsixB{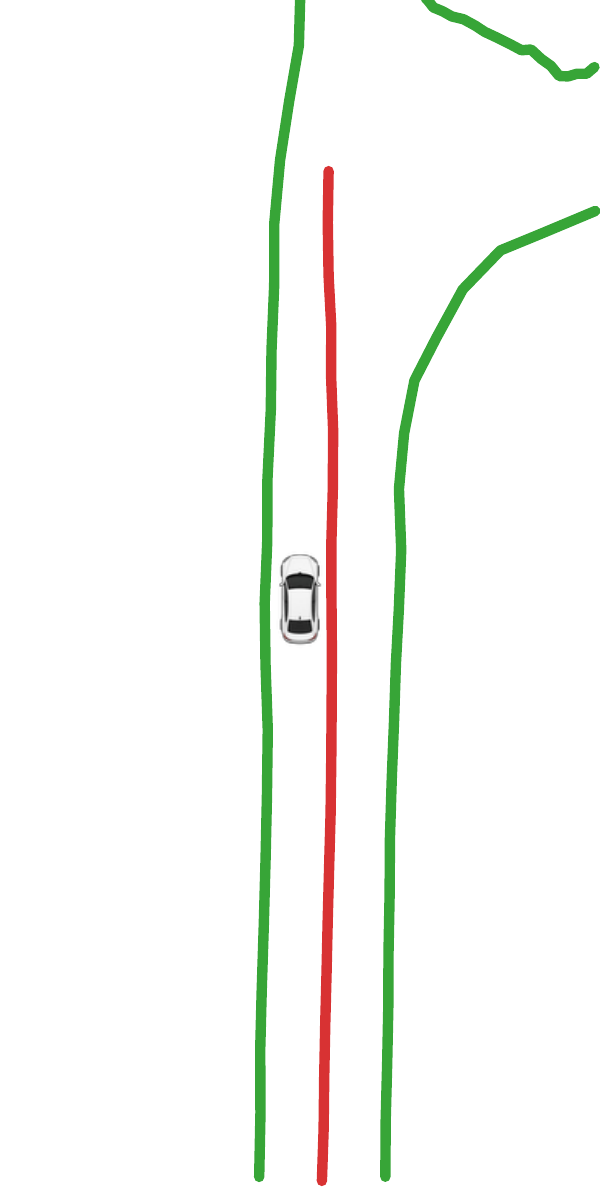} &
\bevcellsixB{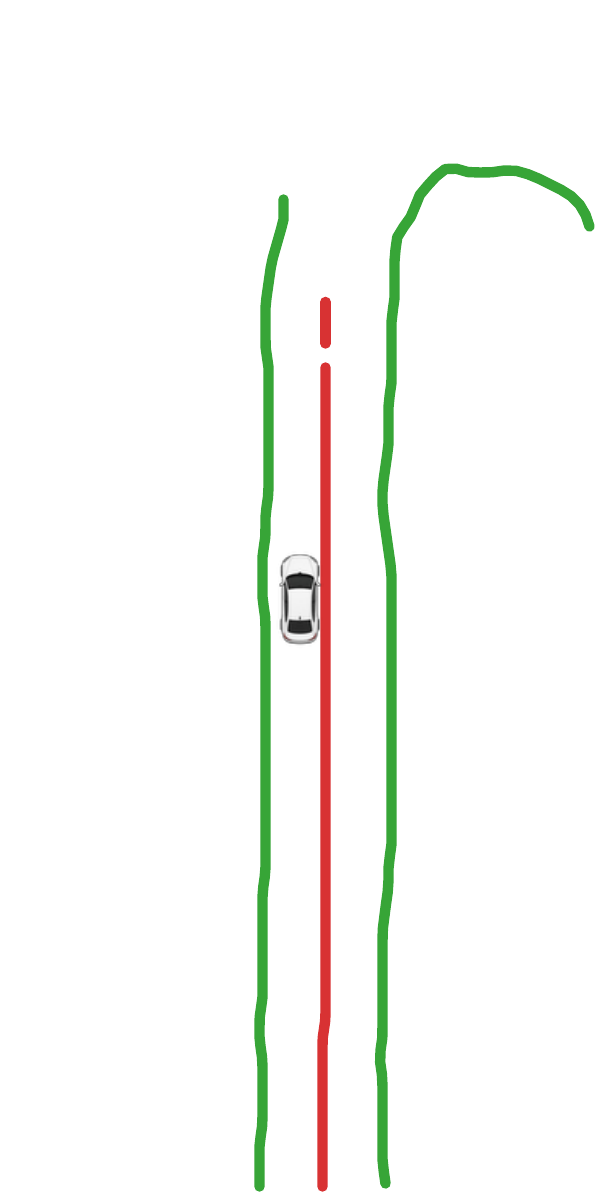} \\
\noalign{\vskip 2pt}
\rowlblsixB{100\,m$\times$50\,m} &
\bevcellsixB{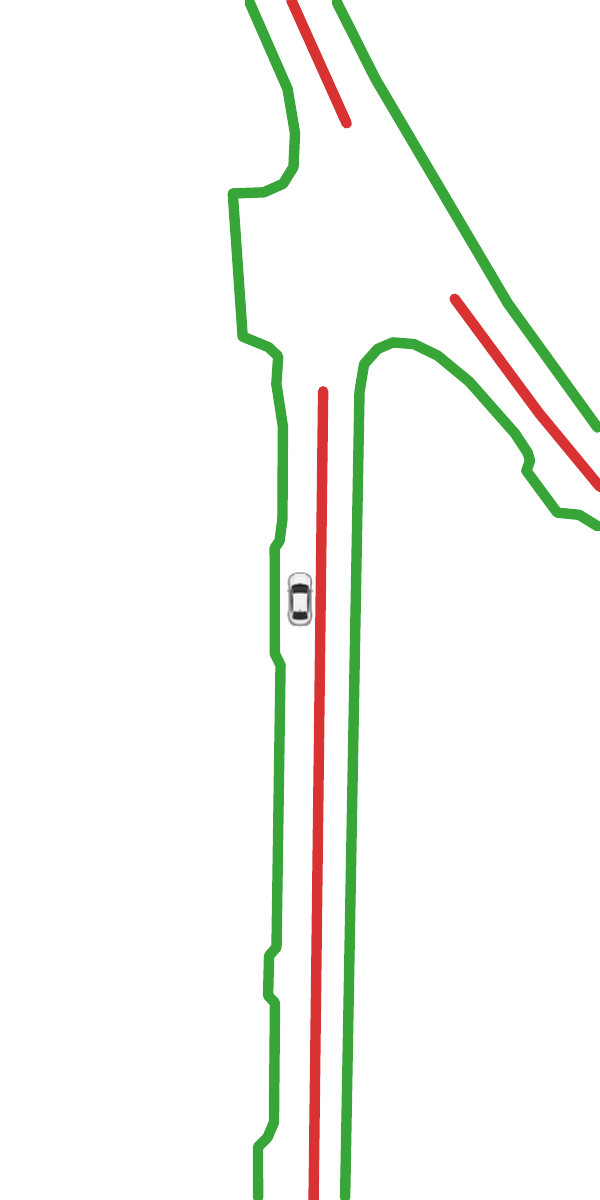} &
\bevcellsixB{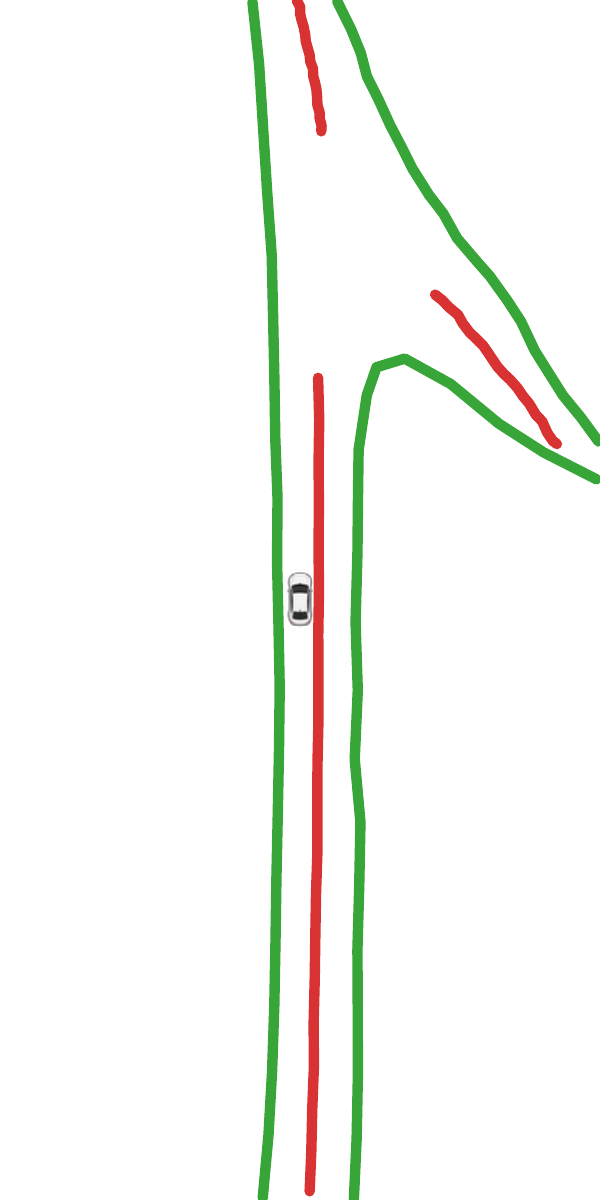} &
\bevcellsixB{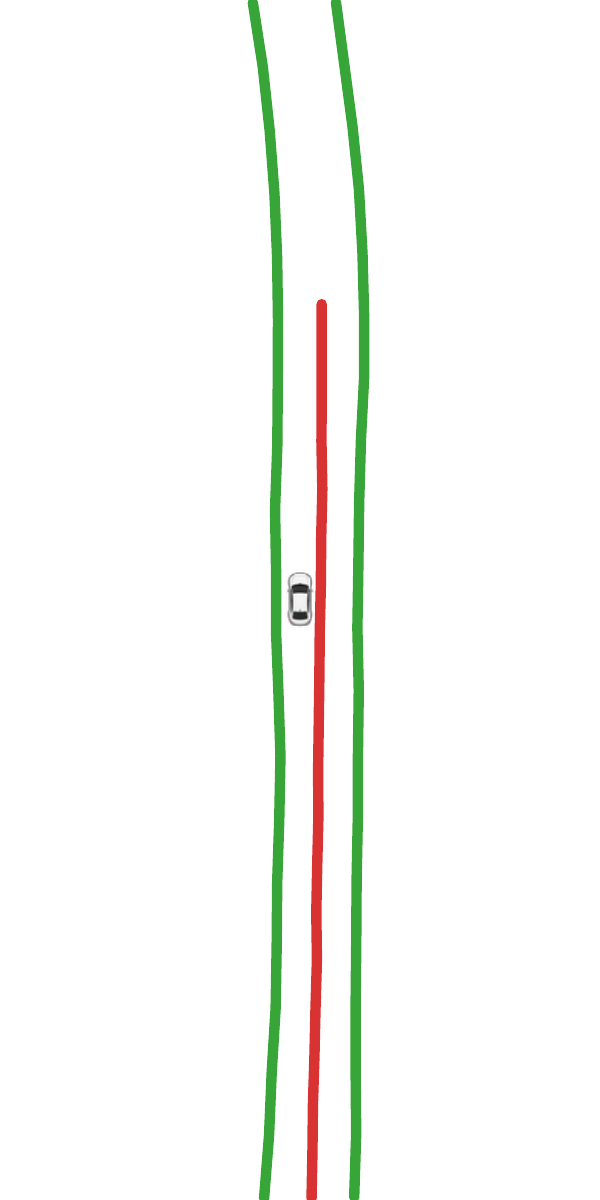} &
\bevcellsixB{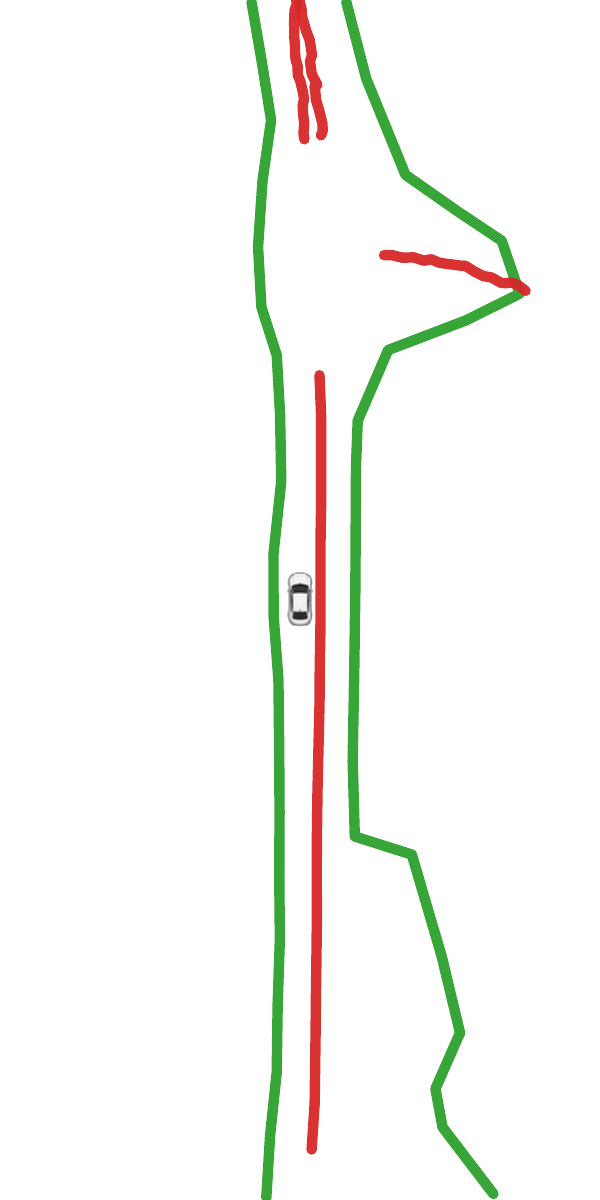} &
\bevcellsixB{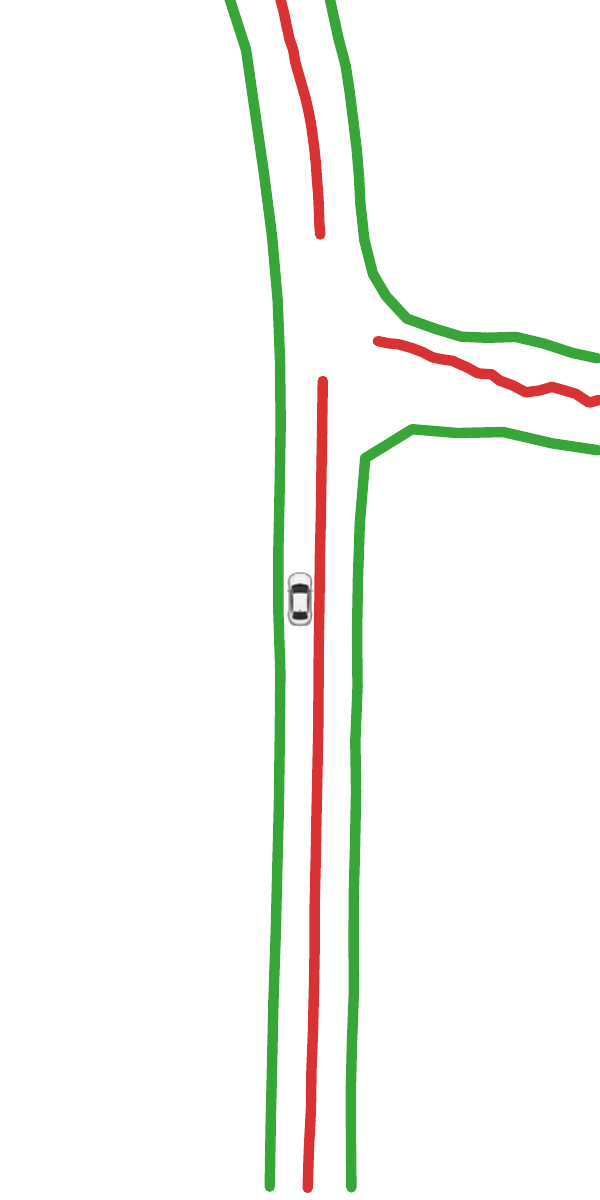} &
\bevcellsixB{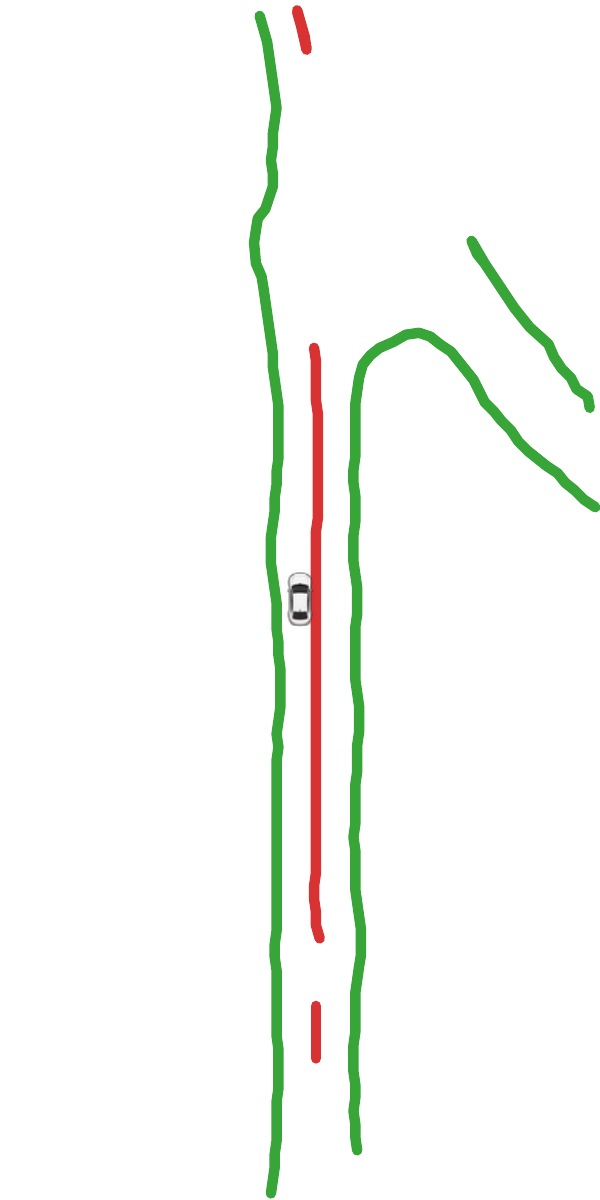} \\
\end{tabular}

\vspace{5pt}
{\footnotesize
\textcolor{clsped}{\rule[-1pt]{8pt}{8pt}}~ped.\,crossing\quad
\textcolor{clsdiv}{\rule[-1pt]{8pt}{8pt}}~divider\quad
\textcolor{clsbnd}{\rule[-1pt]{8pt}{8pt}}~boundary%
}

\caption{Additional qualitative comparison on singapore-hollandvillage.}
\label{fig:qual_appendix_scene1056}
\end{figure*}

%% file: figures/qual_appendix_6frames/scene-0572_2771/qual_scene-0572_2771_6col.tex
\begin{figure*}[t]
\centering

\definecolor{clsped}{HTML}{1F77B4}
\definecolor{clsdiv}{HTML}{D62728}
\definecolor{clsbnd}{HTML}{2CA02C}

\newlength{\camhsixE}    \setlength{\camhsixE}{0.115\textwidth}
\newlength{\sathsixE}    \setlength{\sathsixE}{\dimexpr 2\camhsixE + 2pt\relax}
\newlength{\bevhsixE}    \setlength{\bevhsixE}{0.215\textwidth}
\newlength{\bevimgwsixE} \setlength{\bevimgwsixE}{0.105\textwidth}
\newlength{\bevcellwsixE}\setlength{\bevcellwsixE}{0.13\textwidth}

\setlength{\tabcolsep}{0pt}
\renewcommand{\arraystretch}{1.0}
\noindent
\begin{minipage}[c]{0.63\textwidth}
\centering
{\small\textsf{Multi-view Images}}\\[2pt]
\begin{tabular}{@{}c@{\hskip 2pt}c@{\hskip 2pt}c@{}}
\includegraphics[height=\camhsixE]{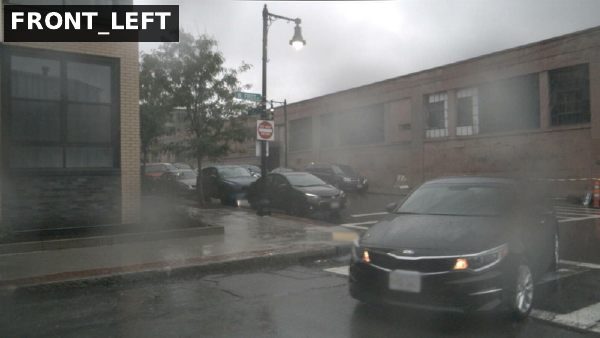} &
\includegraphics[height=\camhsixE]{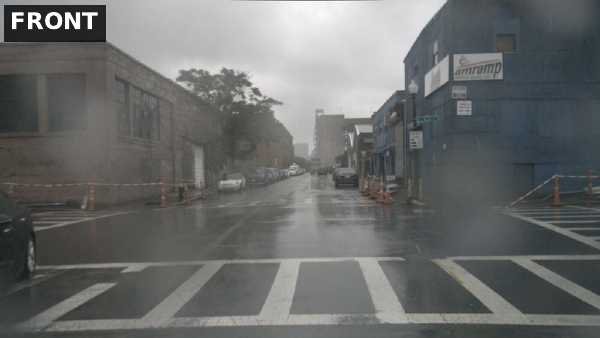} &
\includegraphics[height=\camhsixE]{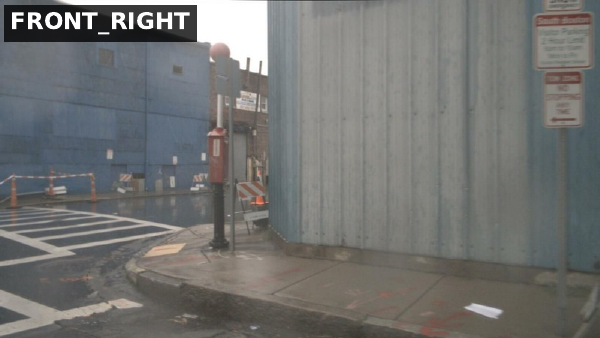} \\
\includegraphics[height=\camhsixE]{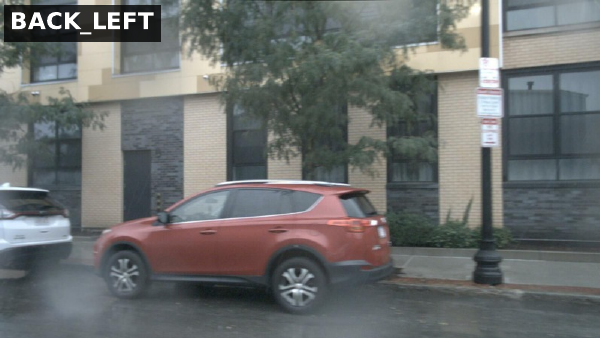} &
\includegraphics[height=\camhsixE]{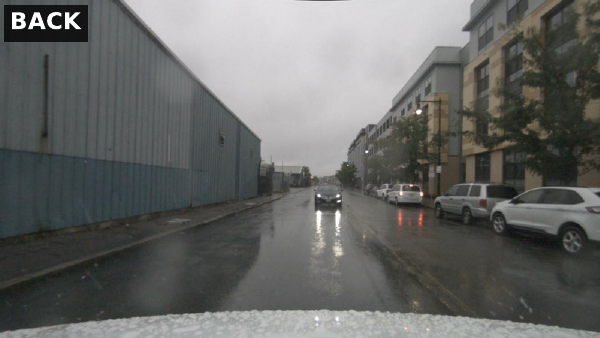} &
\includegraphics[height=\camhsixE]{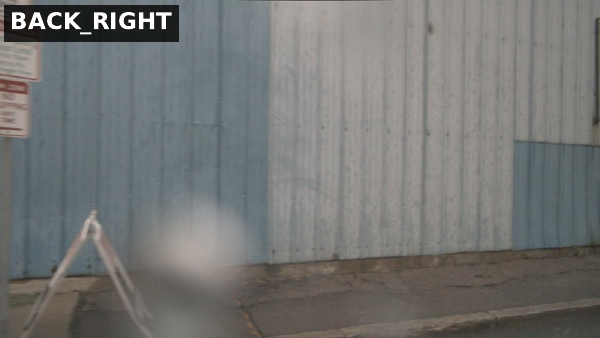} \\
\end{tabular}
\end{minipage}%
\hspace{4pt}%
\begin{minipage}[c]{0.25\textwidth}
\centering
{\small\textsf{Satellite Image}}\\[2pt]
\raisebox{4pt}{%
\mbox{%
\includegraphics[height=\sathsixE]{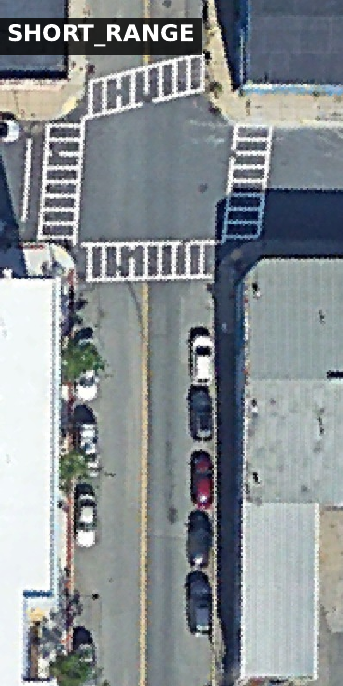}\hspace{3pt}%
\includegraphics[height=\sathsixE]{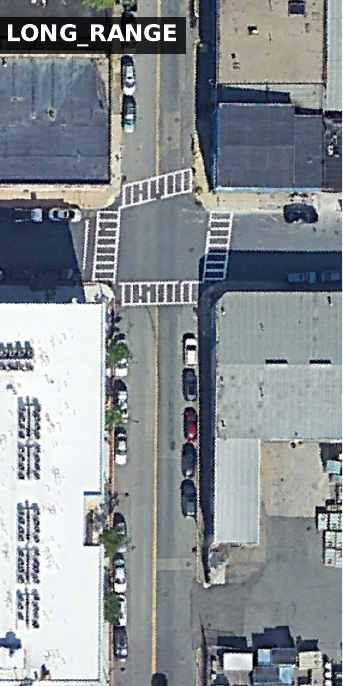}%
}
}
\end{minipage}

\vspace{8pt}

\newcommand{\rowlblsixE}[1]{\parbox[c][\bevhsixE][c]{4.5em}{\centering\footnotesize\mbox{#1}}}
\newcommand{\bevcellsixE}[1]{\parbox[c][\bevhsixE][c]{\bevcellwsixE}{\centering\fbox{\includegraphics[width=\bevimgwsixE]{#1}}}}

\setlength{\fboxsep}{0pt}
\setlength{\fboxrule}{0.4pt}
\setlength{\tabcolsep}{1pt}
\renewcommand{\arraystretch}{1.0}
\begin{tabular}{c@{\hskip 2pt}cccccc}
 & {\footnotesize\textsf{Ground Truth}} & {\footnotesize\textbf{\textsf{ReSMap(Ours)}}} & {\footnotesize\textsf{MapTracker}} & {\footnotesize\textsf{DAMap}} & {\footnotesize\textsf{SDTagNet}} & {\footnotesize\textsf{SatforHDMap}} \\
\rowlblsixE{60\,m$\times$30\,m} &
\bevcellsixE{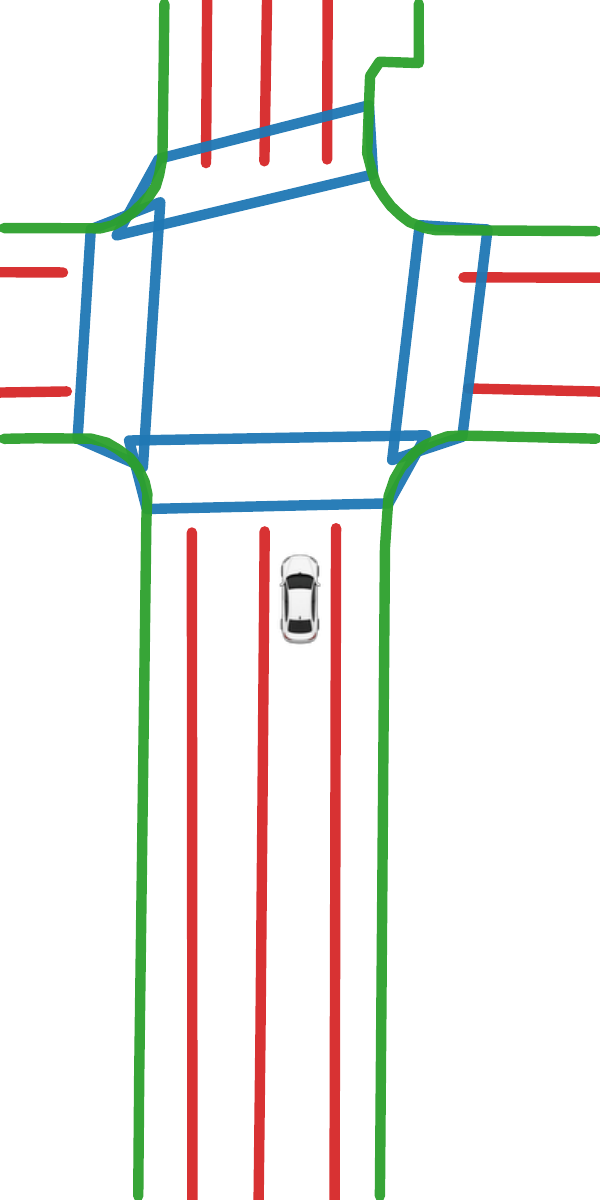} &
\bevcellsixE{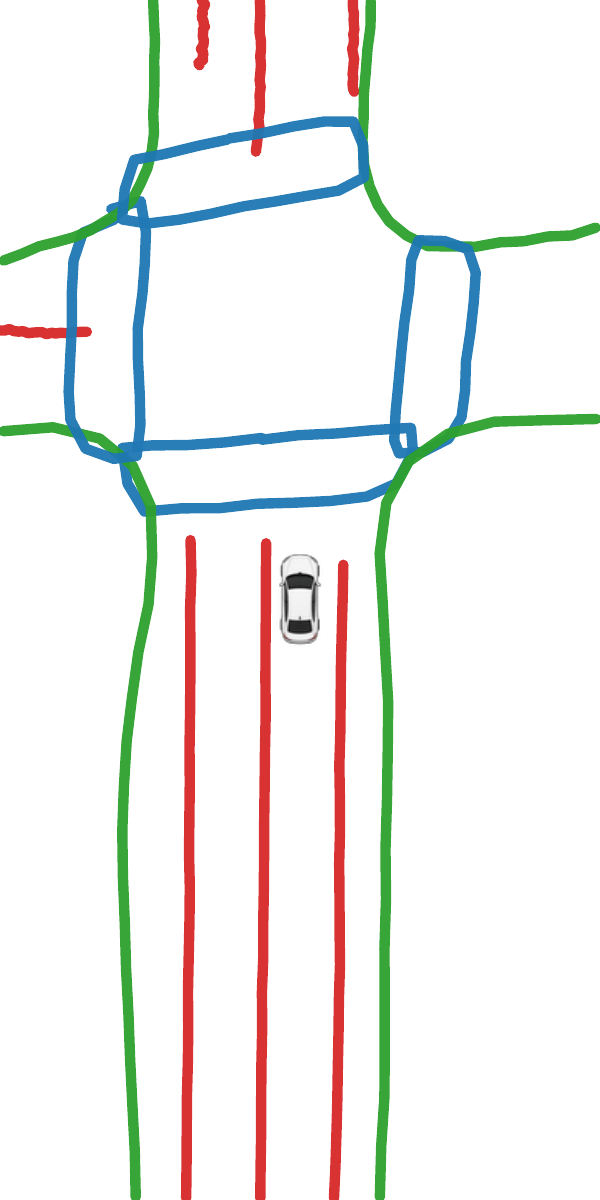} &
\bevcellsixE{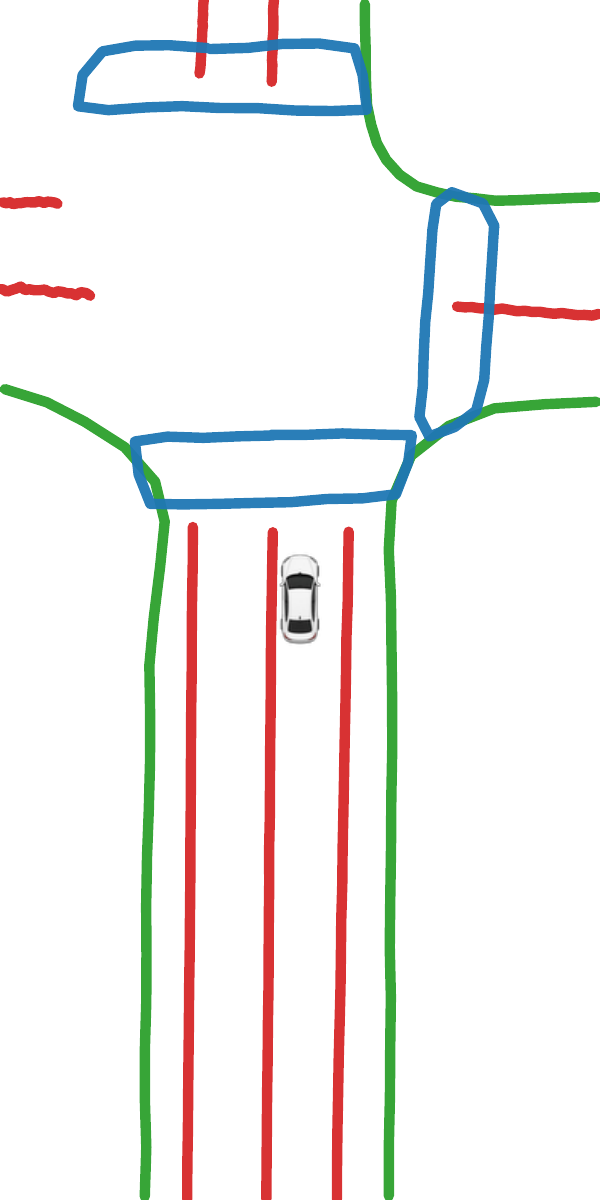} &
\bevcellsixE{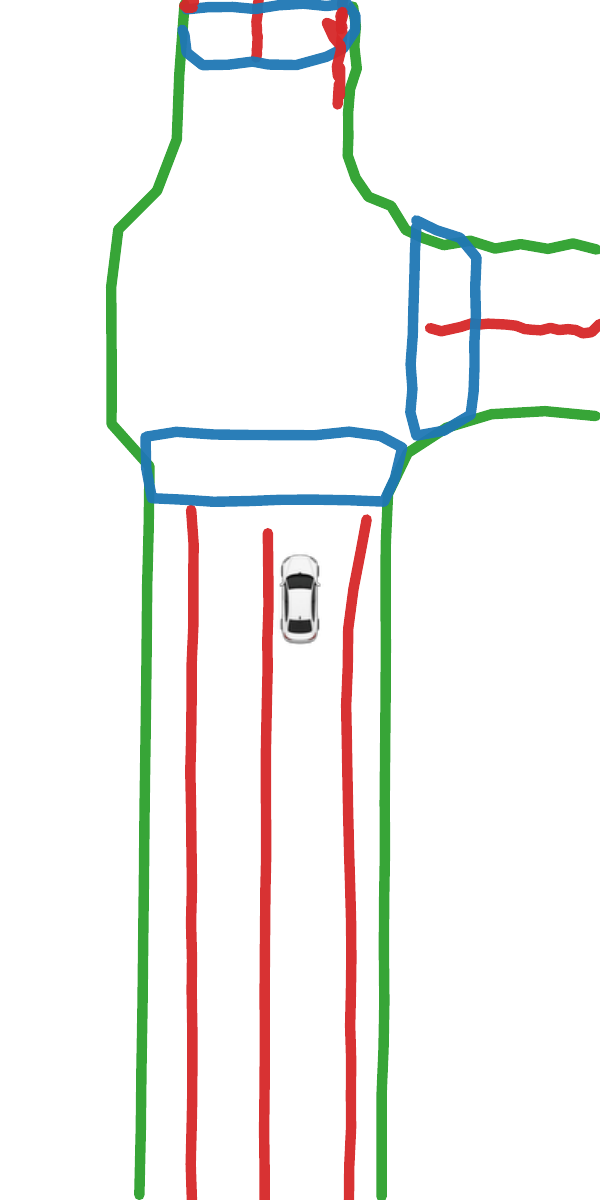} &
\bevcellsixE{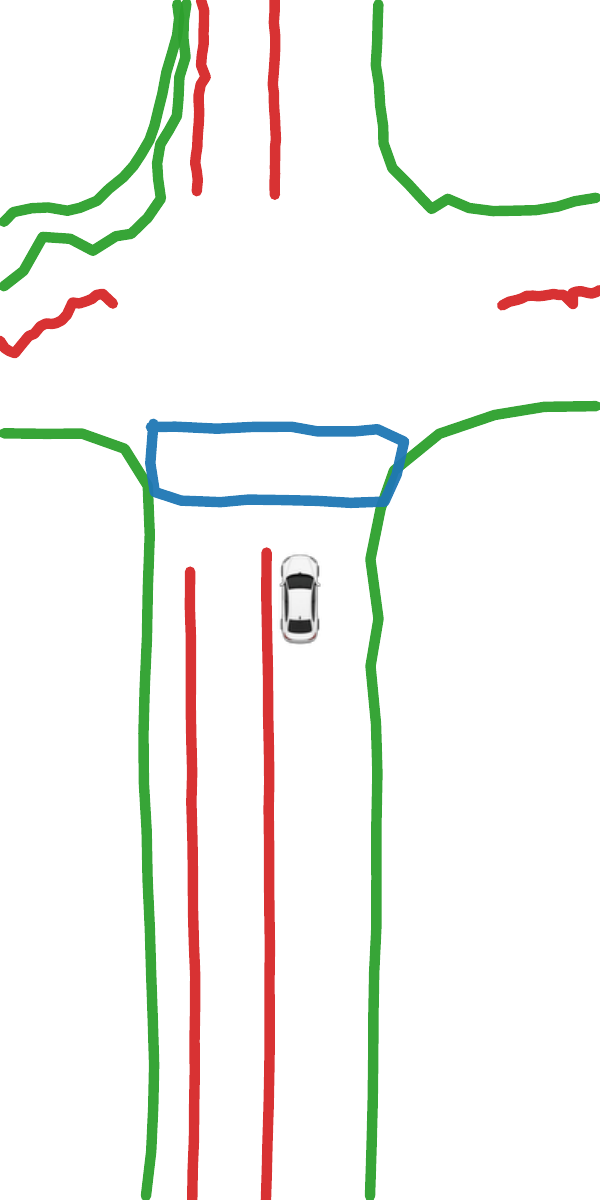} &
\bevcellsixE{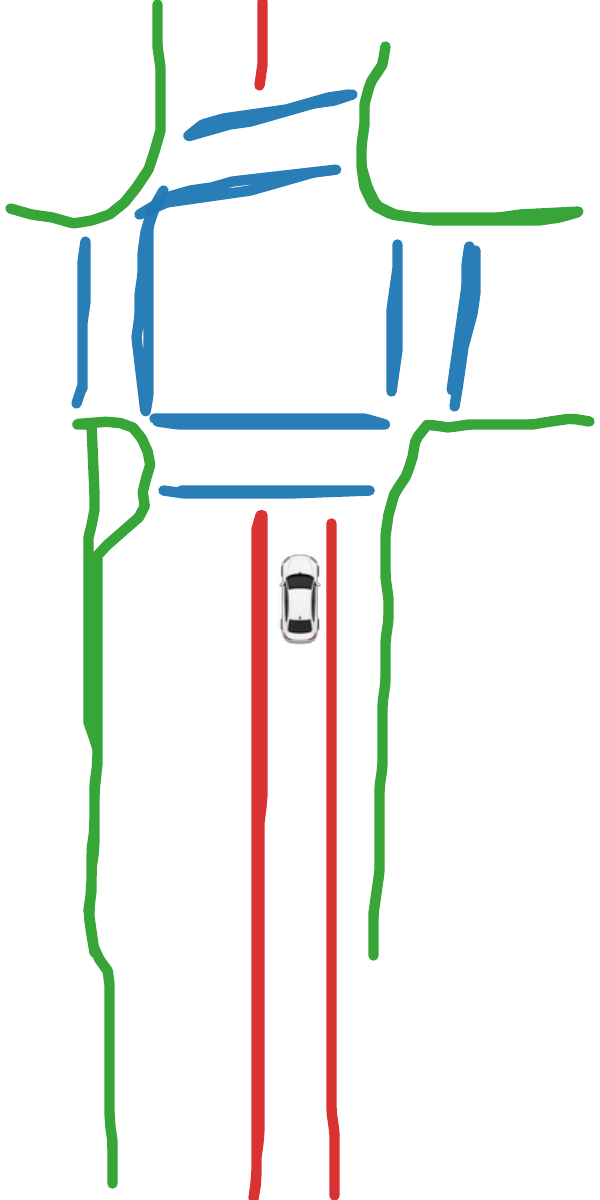} \\
\noalign{\vskip 2pt}
\rowlblsixE{Front-3 drop} &
\bevcellsixE{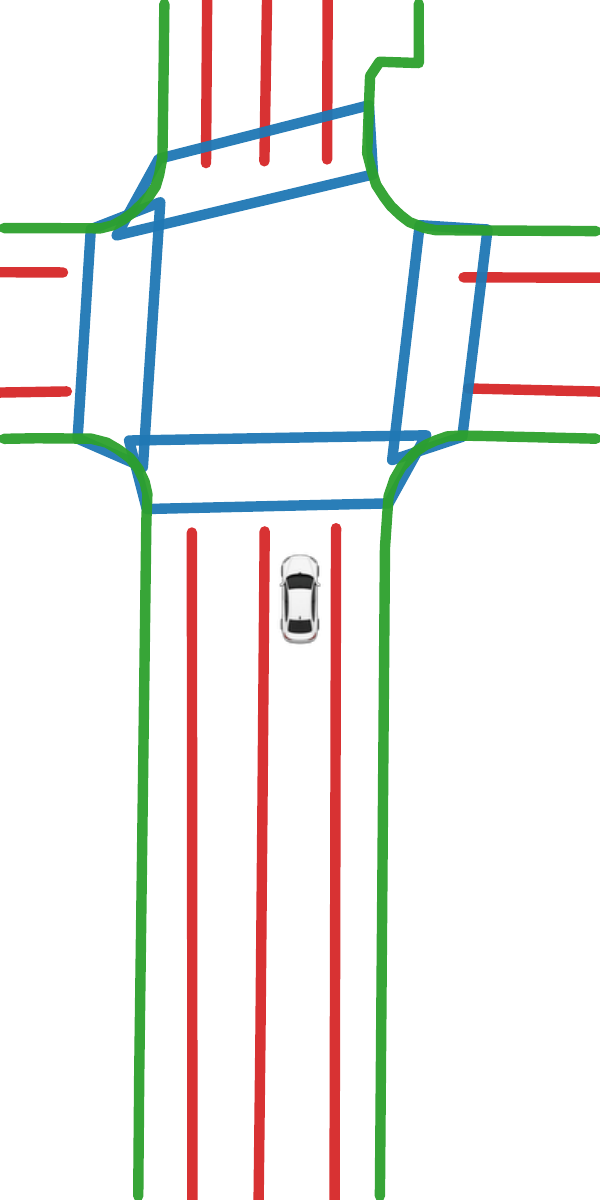} &
\bevcellsixE{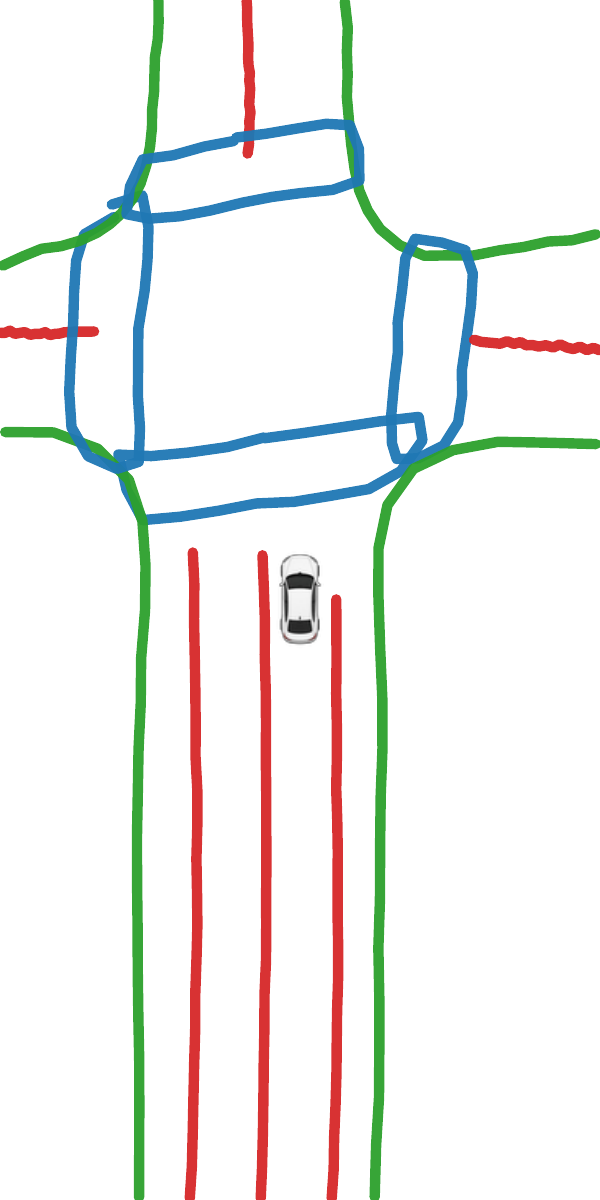} &
\bevcellsixE{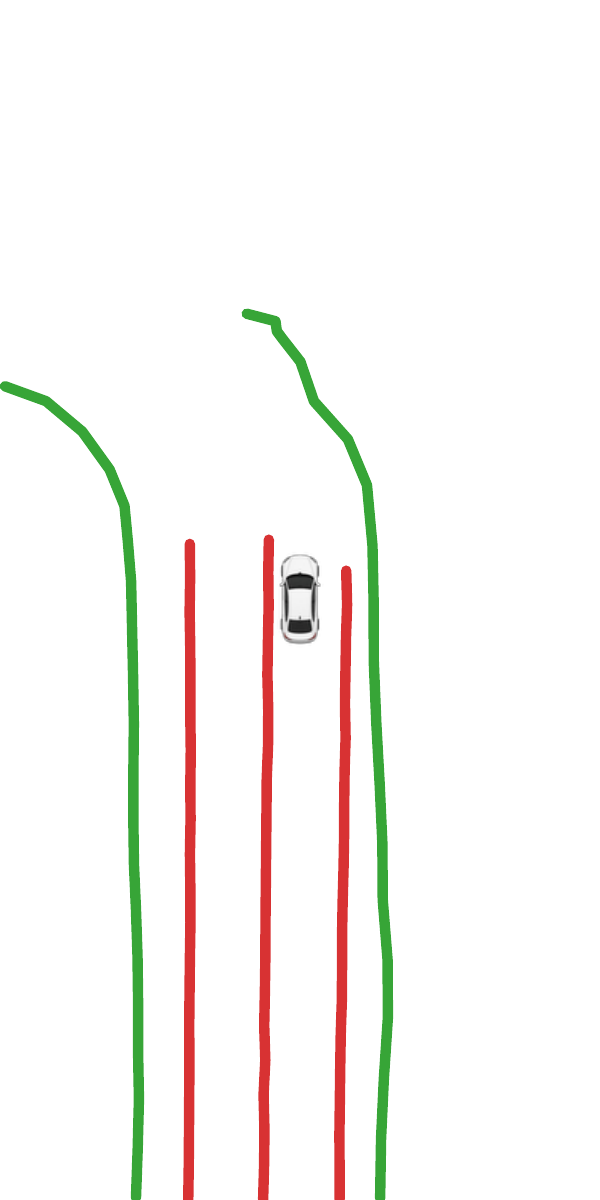} &
\bevcellsixE{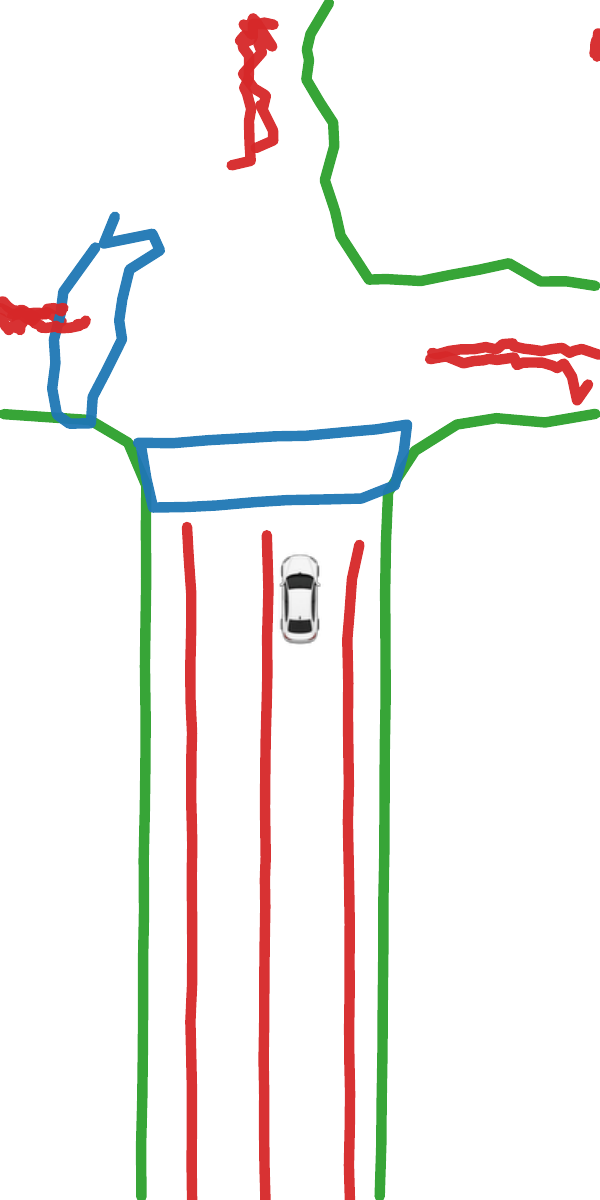} &
\bevcellsixE{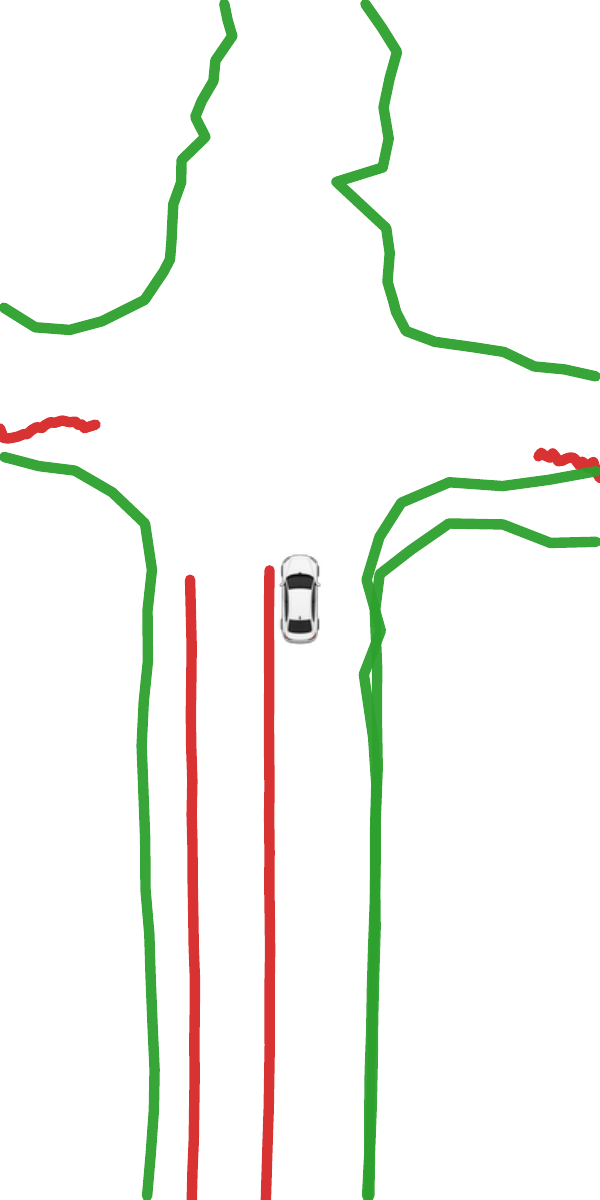} &
\bevcellsixE{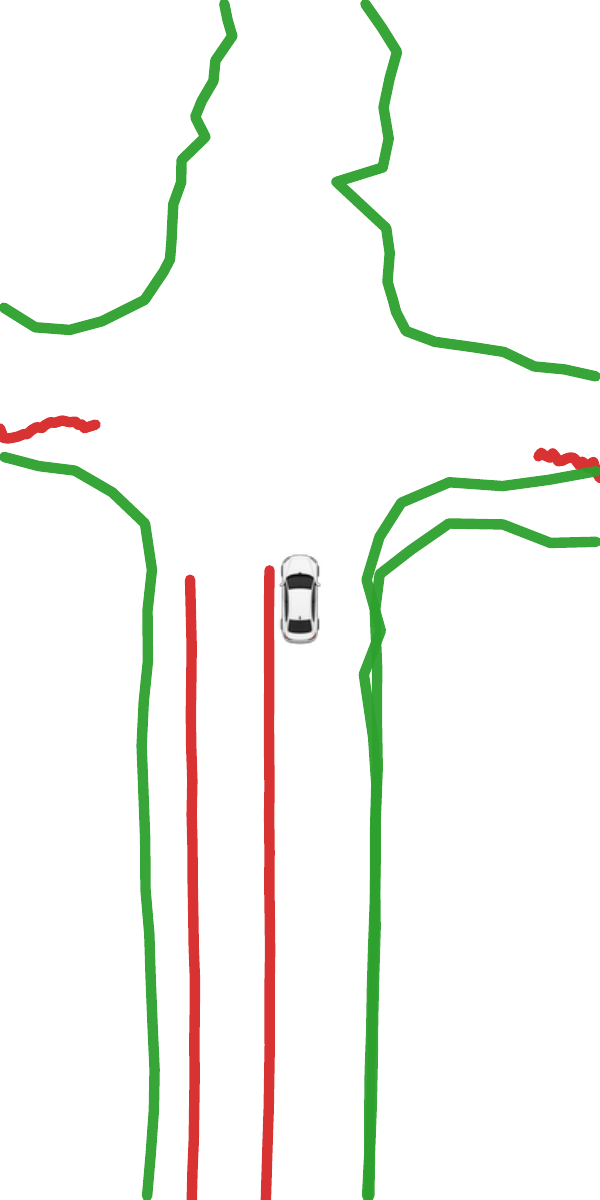} \\
\noalign{\vskip 2pt}
\rowlblsixE{100\,m$\times$50\,m} &
\bevcellsixE{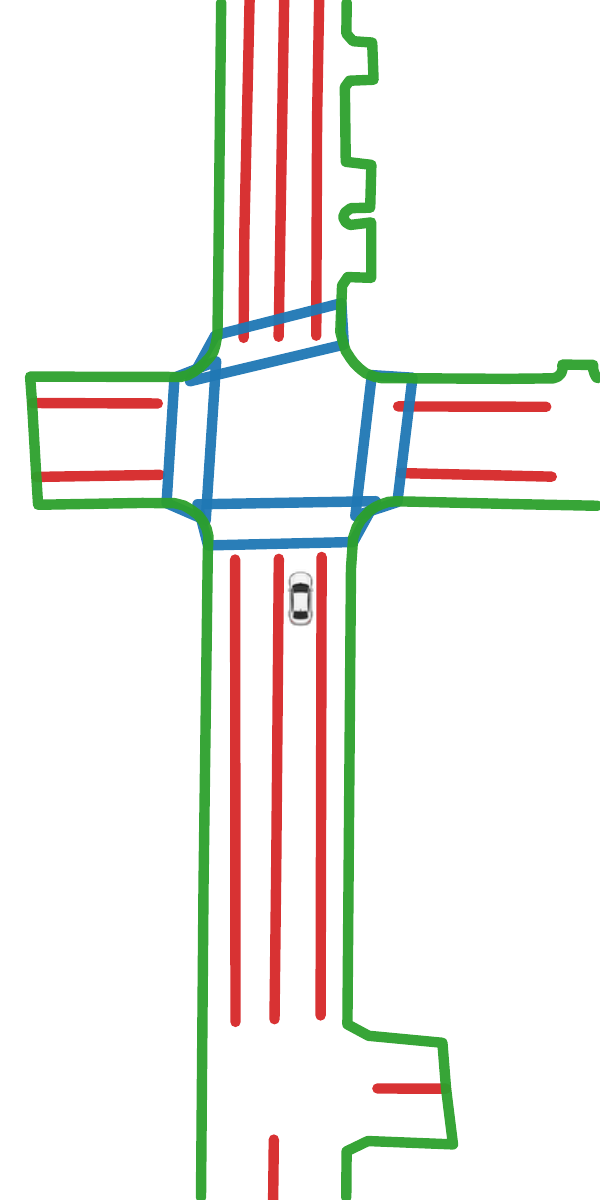} &
\bevcellsixE{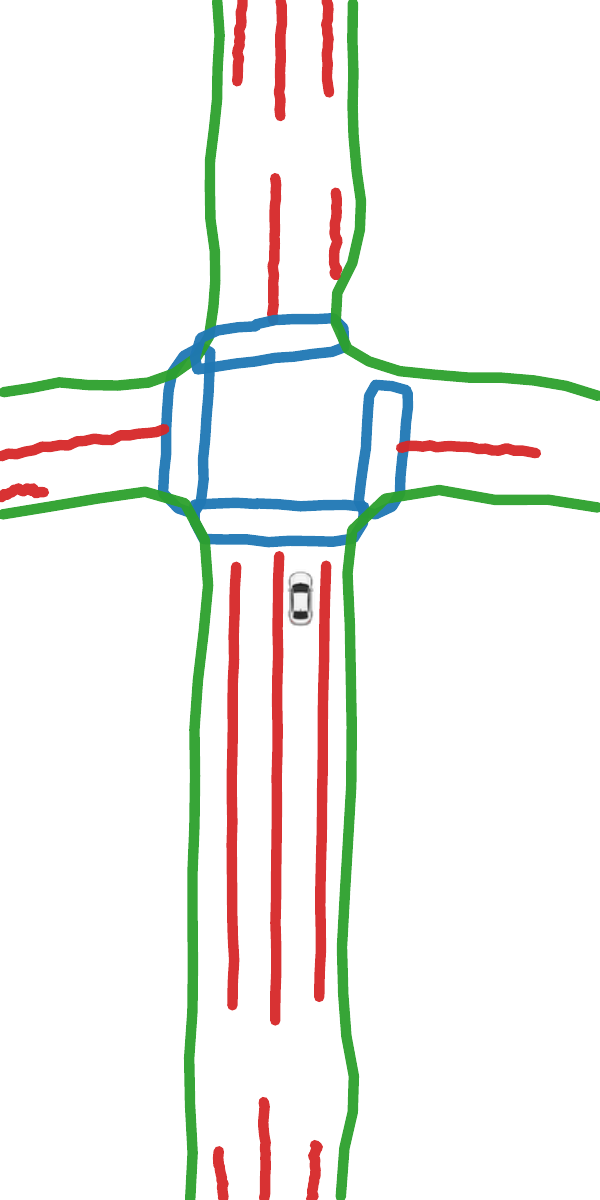} &
\bevcellsixE{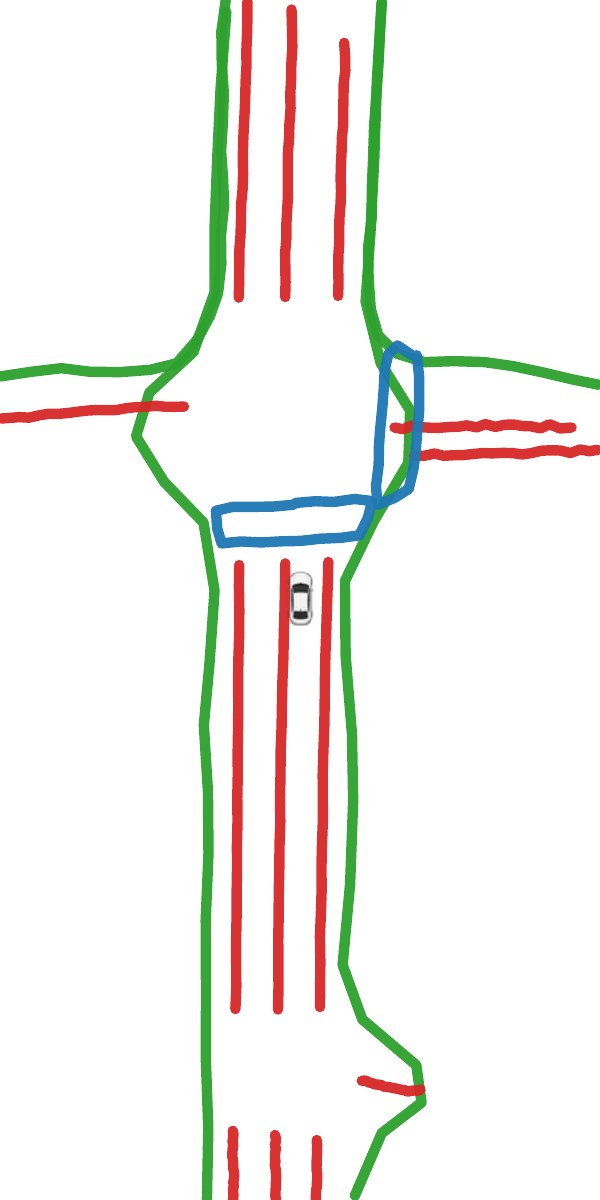} &
\bevcellsixE{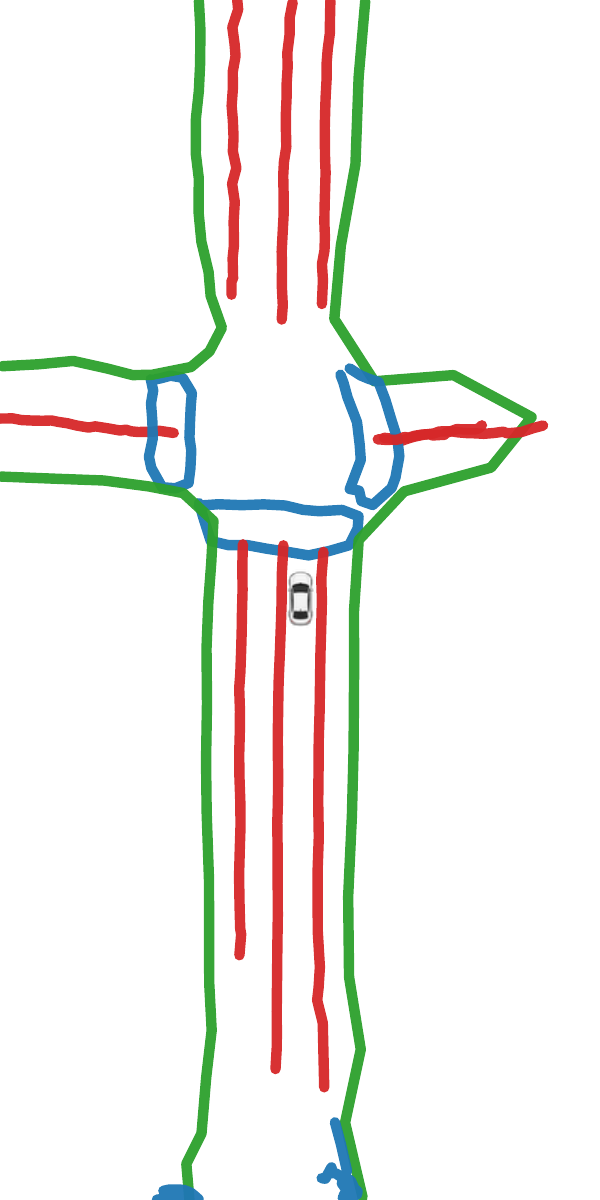} &
\bevcellsixE{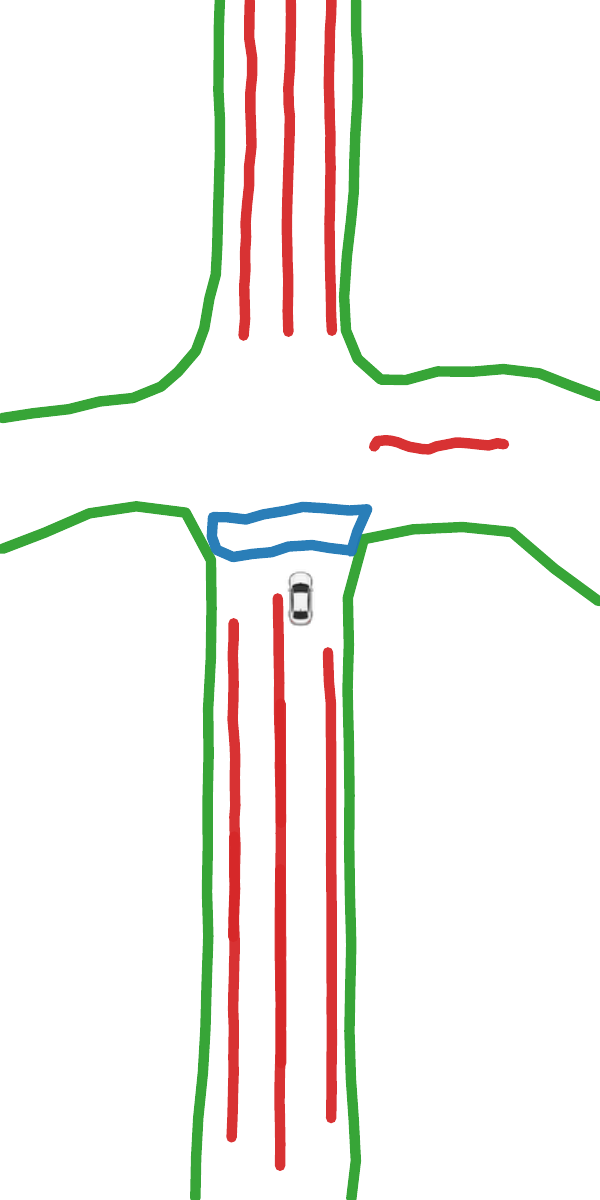} &
\bevcellsixE{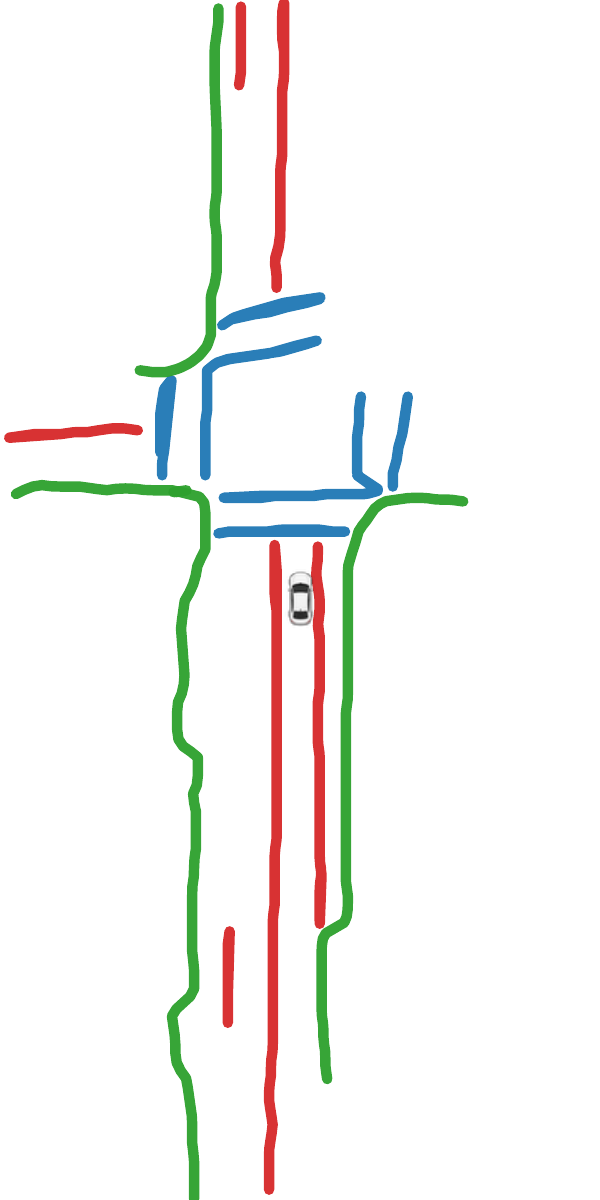} \\
\end{tabular}

\vspace{5pt}
{\footnotesize
\textcolor{clsped}{\rule[-1pt]{8pt}{8pt}}~ped.\,crossing\quad
\textcolor{clsdiv}{\rule[-1pt]{8pt}{8pt}}~divider\quad
\textcolor{clsbnd}{\rule[-1pt]{8pt}{8pt}}~boundary%
}

\caption{Additional qualitative comparison on boston-seaport.}
\label{fig:qual_appendix_scene0572_2771}
\end{figure*}

%% file: figures/qual_appendix_6frames/scene-0795_4690/qual_scene-0795_4690_6col.tex
\begin{figure*}[t]
\centering

\definecolor{clsped}{HTML}{1F77B4}
\definecolor{clsdiv}{HTML}{D62728}
\definecolor{clsbnd}{HTML}{2CA02C}

\newlength{\camhsixF}    \setlength{\camhsixF}{0.115\textwidth}
\newlength{\sathsixF}    \setlength{\sathsixF}{\dimexpr 2\camhsixF + 2pt\relax}
\newlength{\bevhsixF}    \setlength{\bevhsixF}{0.215\textwidth}
\newlength{\bevimgwsixF} \setlength{\bevimgwsixF}{0.105\textwidth}
\newlength{\bevcellwsixF}\setlength{\bevcellwsixF}{0.13\textwidth}

\setlength{\tabcolsep}{0pt}
\renewcommand{\arraystretch}{1.0}
\noindent
\begin{minipage}[c]{0.63\textwidth}
\centering
{\small\textsf{Multi-view Images}}\\[2pt]
\begin{tabular}{@{}c@{\hskip 2pt}c@{\hskip 2pt}c@{}}
\includegraphics[height=\camhsixF]{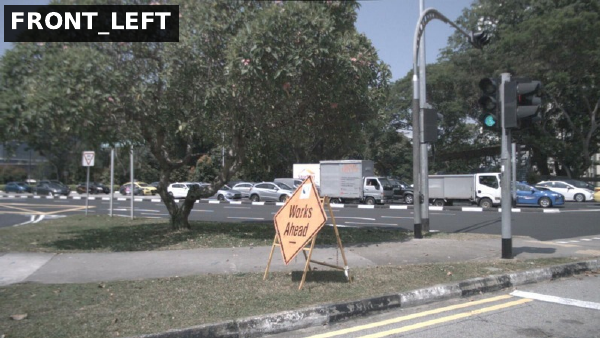} &
\includegraphics[height=\camhsixF]{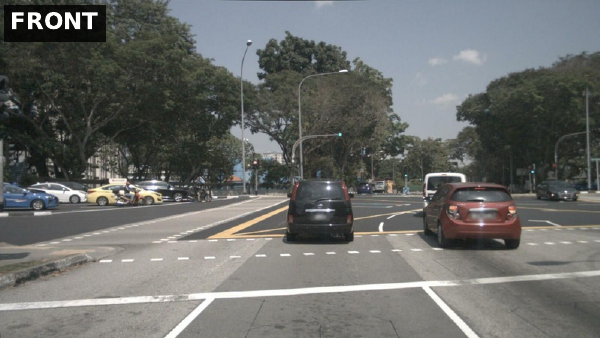} &
\includegraphics[height=\camhsixF]{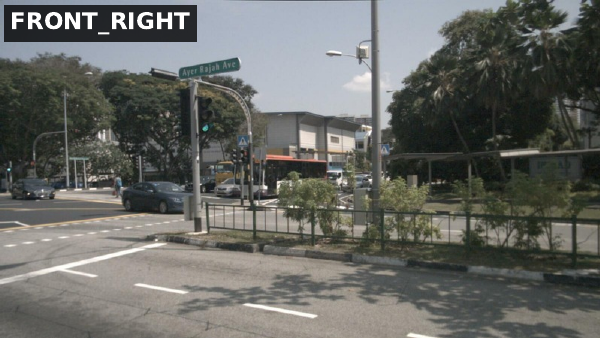} \\
\includegraphics[height=\camhsixF]{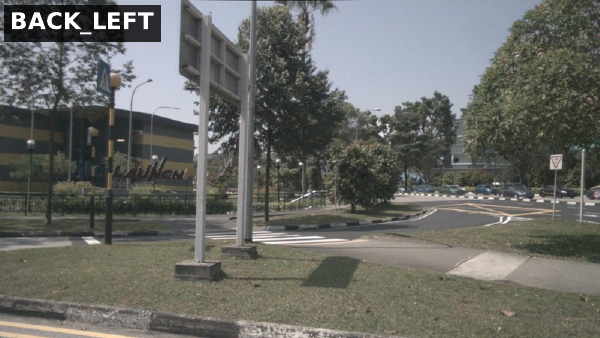} &
\includegraphics[height=\camhsixF]{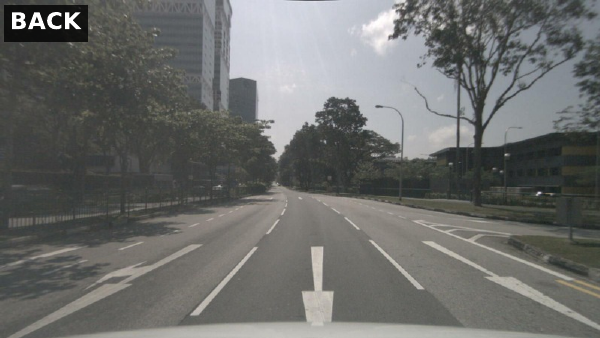} &
\includegraphics[height=\camhsixF]{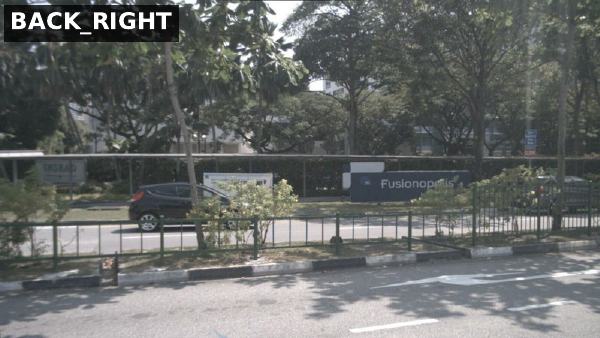} \\
\end{tabular}
\end{minipage}%
\hspace{4pt}%
\begin{minipage}[c]{0.25\textwidth}
\centering
{\small\textsf{Satellite Image}}\\[2pt]
\raisebox{4pt}{%
\mbox{%
\includegraphics[height=\sathsixF]{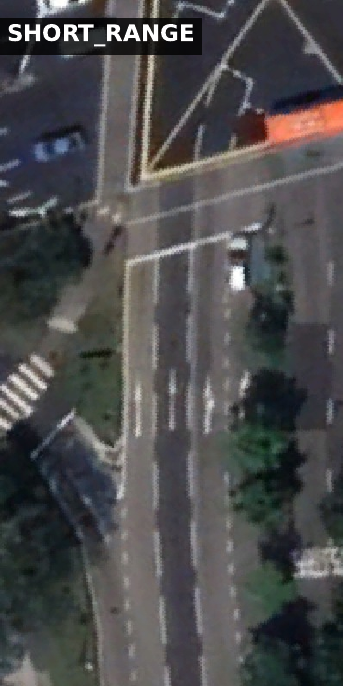}\hspace{3pt}%
\includegraphics[height=\sathsixF]{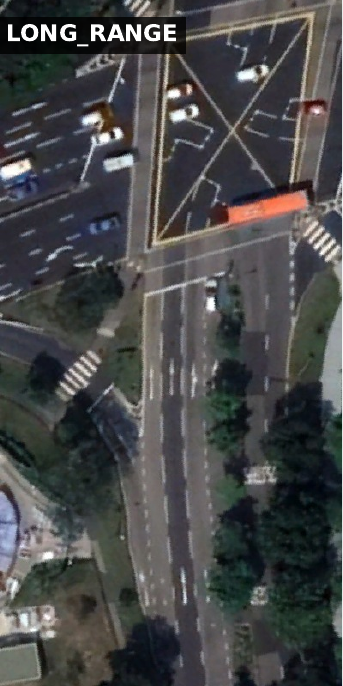}%
}
}
\end{minipage}

\vspace{8pt}

\newcommand{\rowlblsixF}[1]{\parbox[c][\bevhsixF][c]{4.5em}{\centering\footnotesize\mbox{#1}}}
\newcommand{\bevcellsixF}[1]{\parbox[c][\bevhsixF][c]{\bevcellwsixF}{\centering\fbox{\includegraphics[width=\bevimgwsixF]{#1}}}}

\setlength{\fboxsep}{0pt}
\setlength{\fboxrule}{0.4pt}
\setlength{\tabcolsep}{1pt}
\renewcommand{\arraystretch}{1.0}
\begin{tabular}{c@{\hskip 2pt}cccccc}
 & {\footnotesize\textsf{Ground Truth}} & {\footnotesize\textbf{\textsf{ReSMap(Ours)}}} & {\footnotesize\textsf{MapTracker}} & {\footnotesize\textsf{DAMap}} & {\footnotesize\textsf{SDTagNet}} & {\footnotesize\textsf{SatforHDMap}} \\
\rowlblsixF{60\,m$\times$30\,m} &
\bevcellsixF{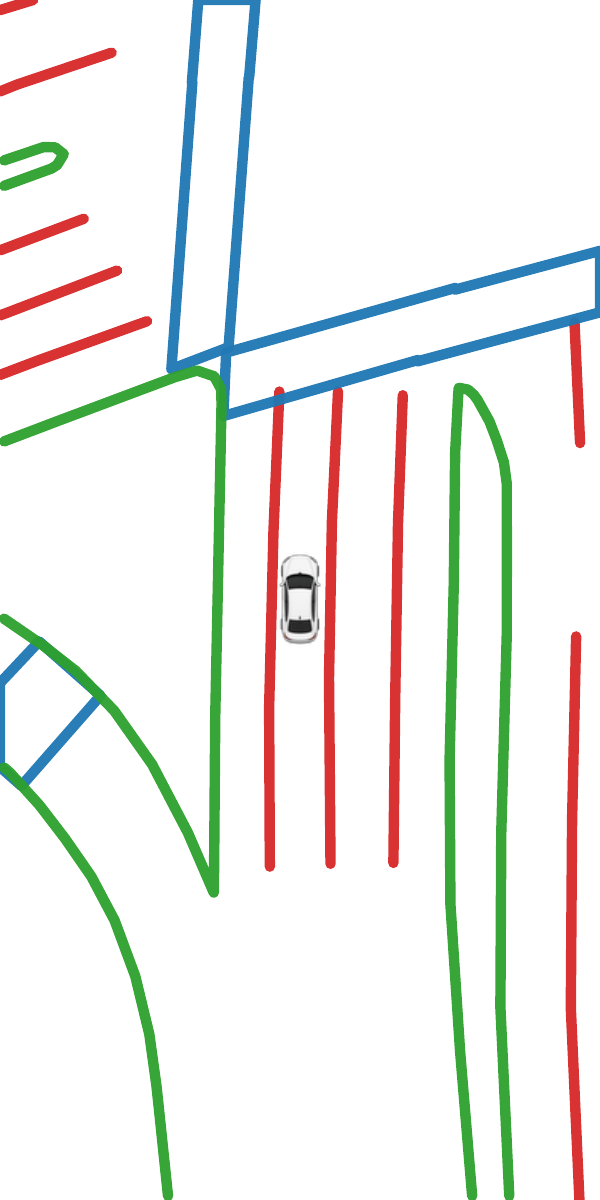} &
\bevcellsixF{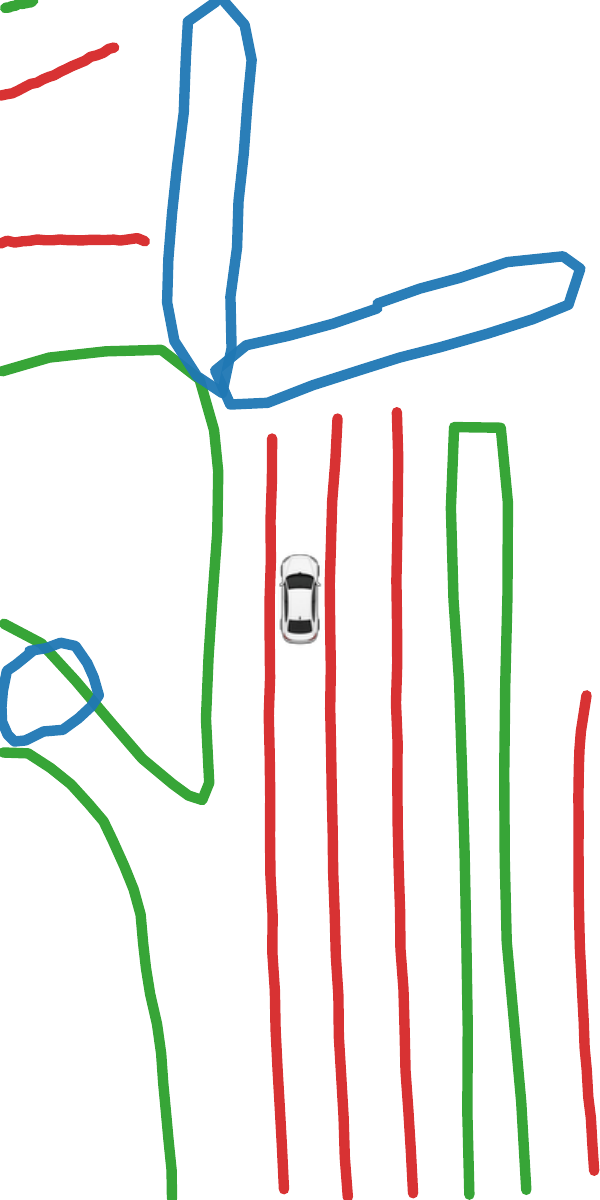} &
\bevcellsixF{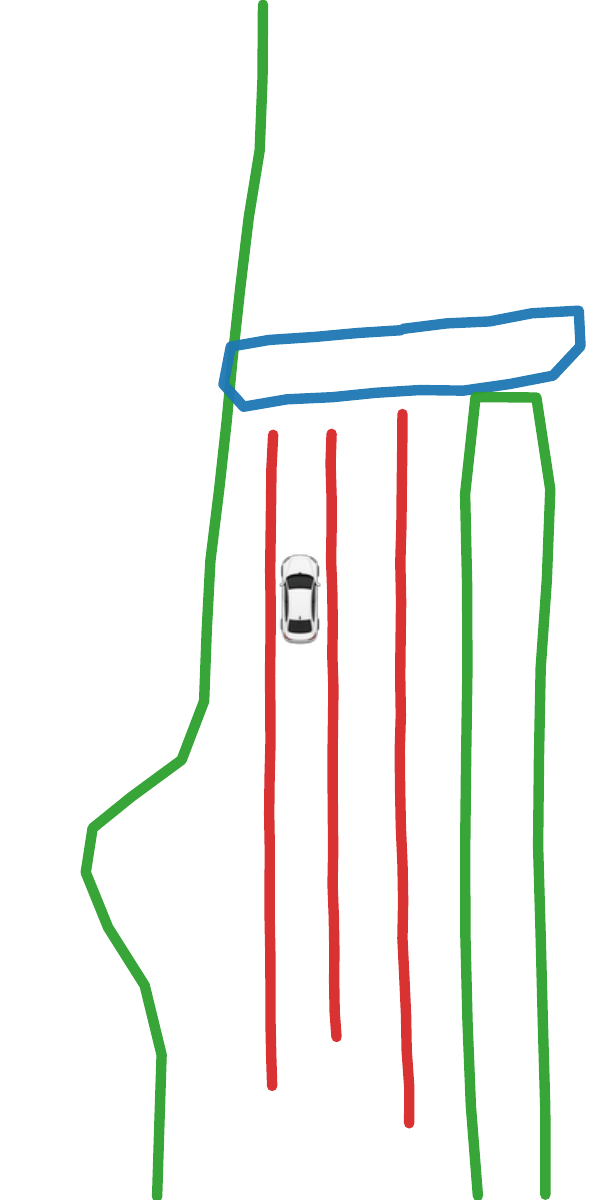} &
\bevcellsixF{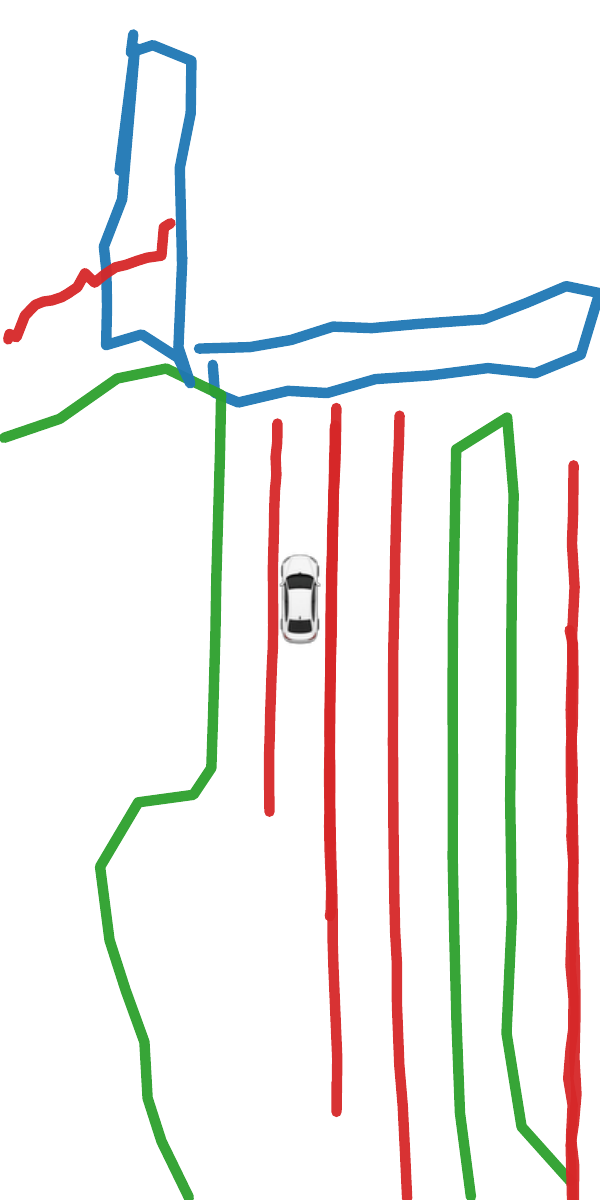} &
\bevcellsixF{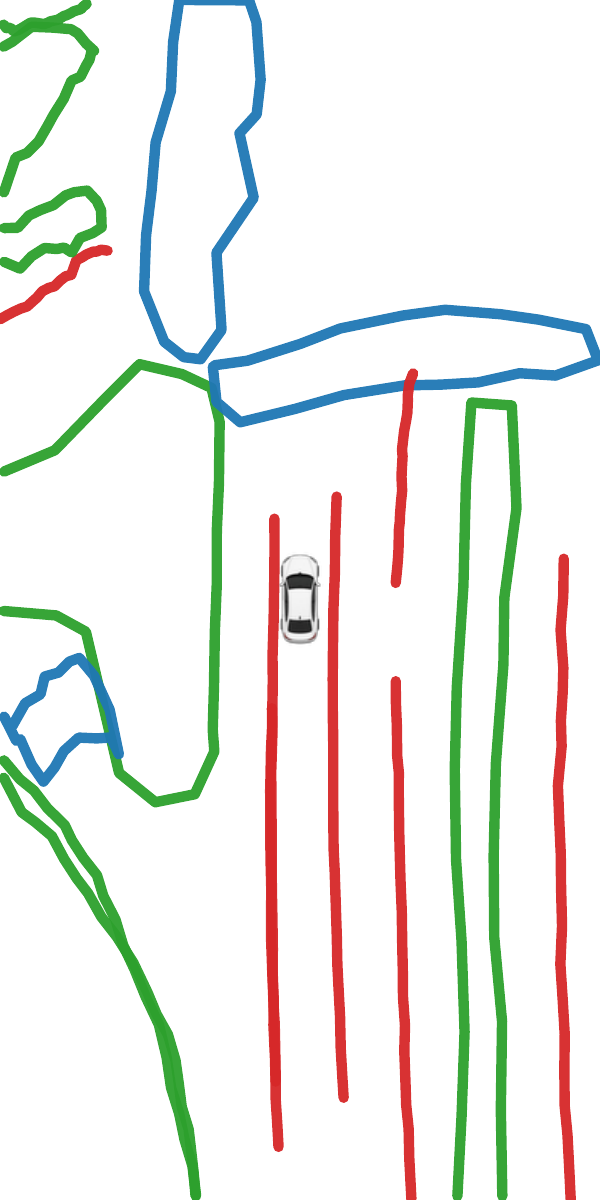} &
\bevcellsixF{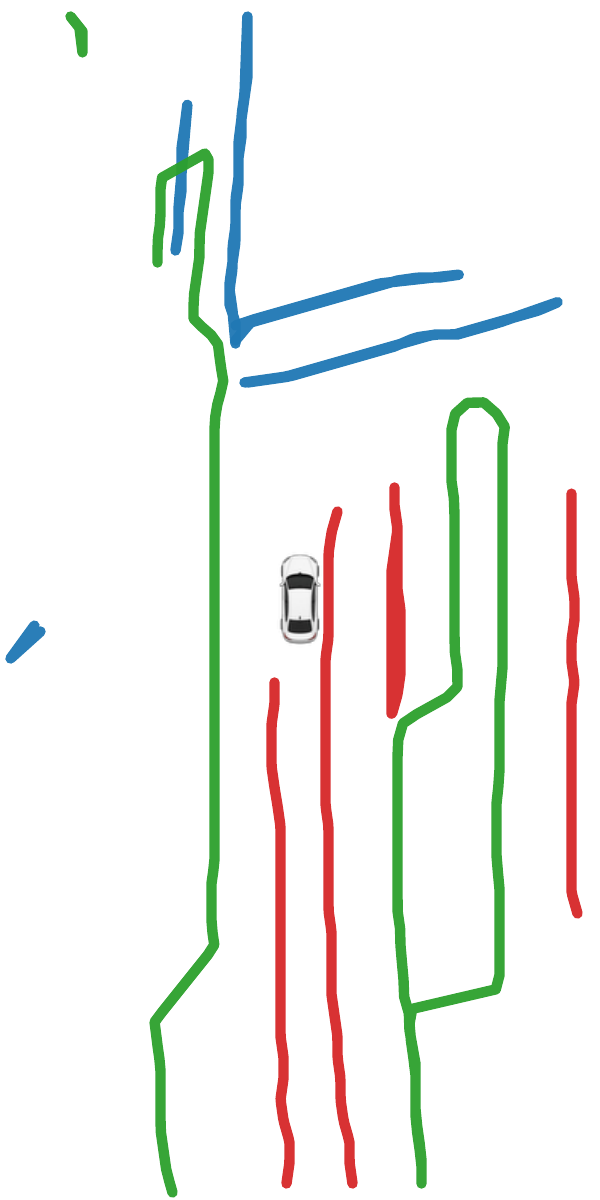} \\
\noalign{\vskip 2pt}
\rowlblsixF{Front-3 drop} &
\bevcellsixF{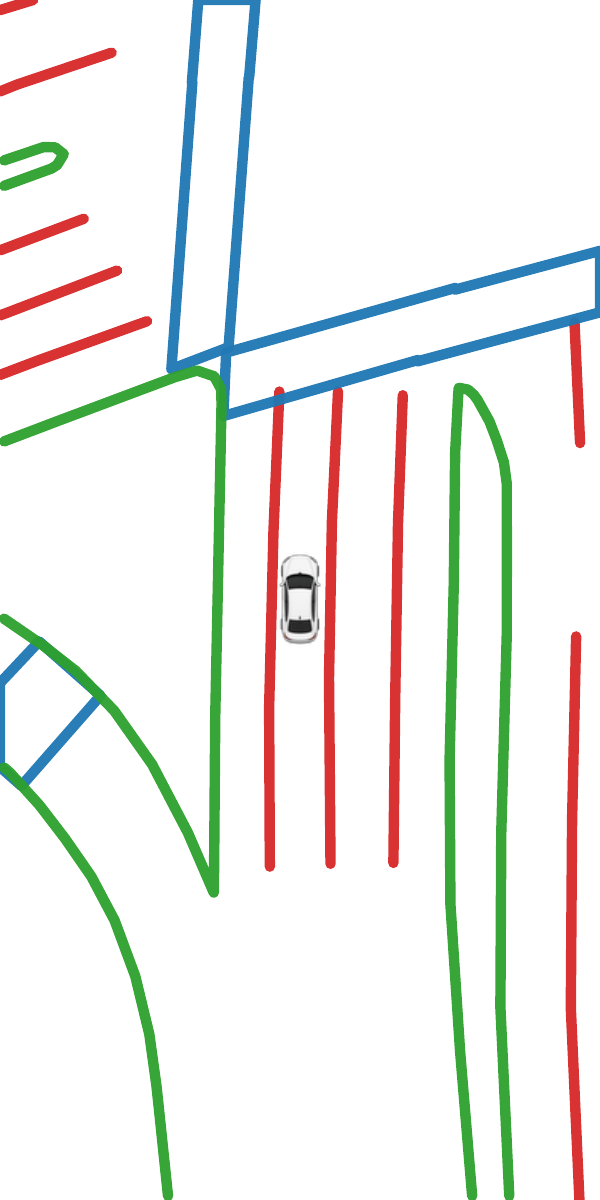} &
\bevcellsixF{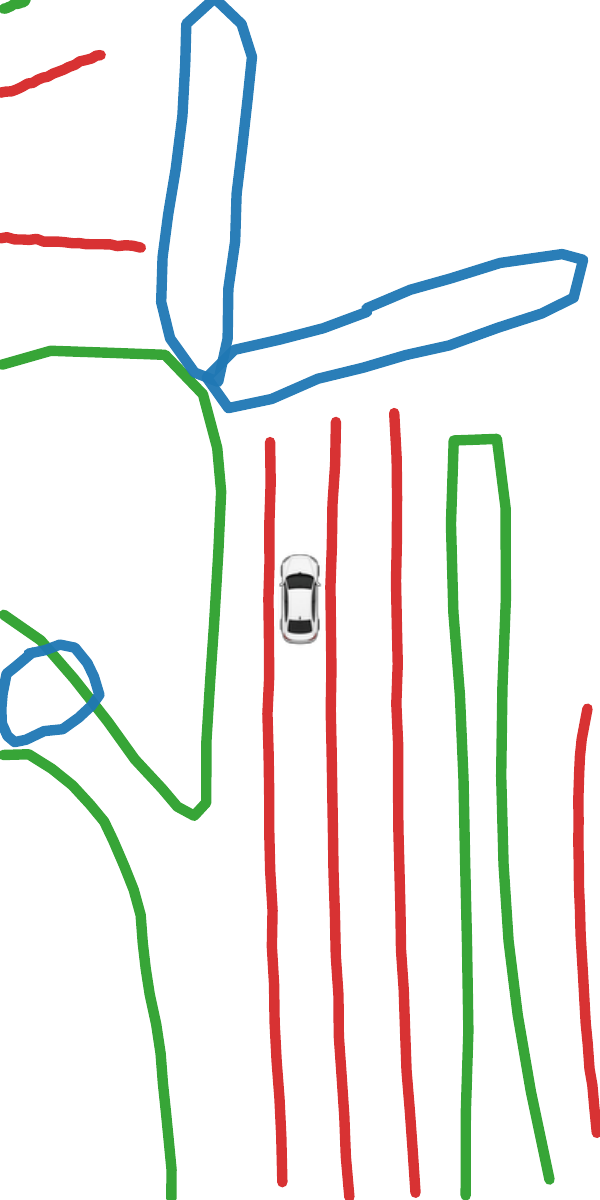} &
\bevcellsixF{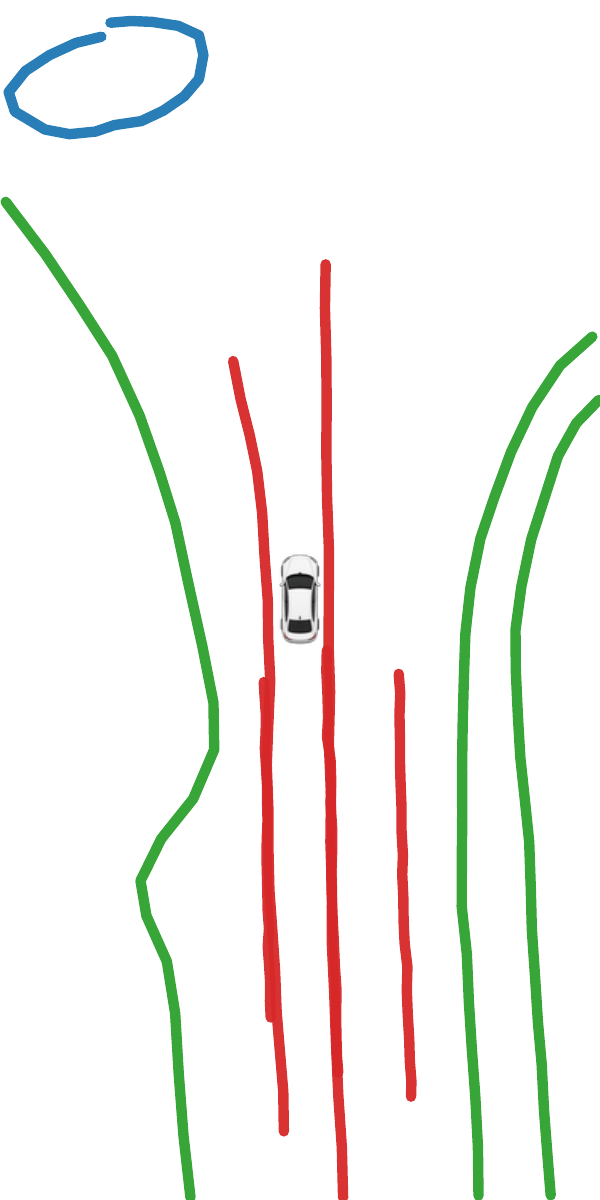} &
\bevcellsixF{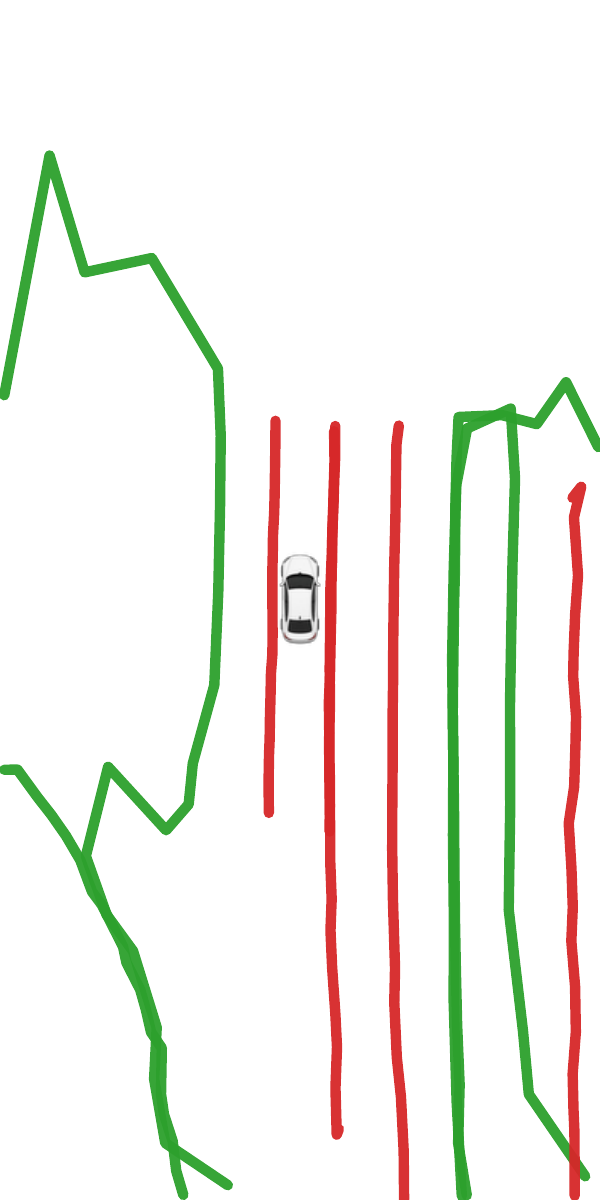} &
\bevcellsixF{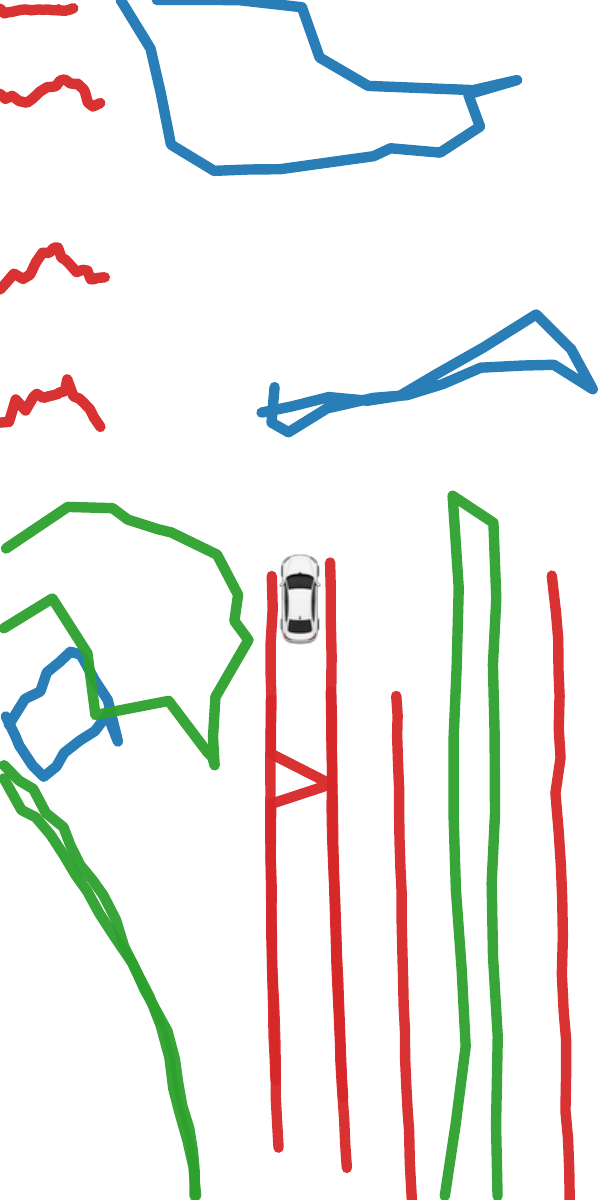} &
\bevcellsixF{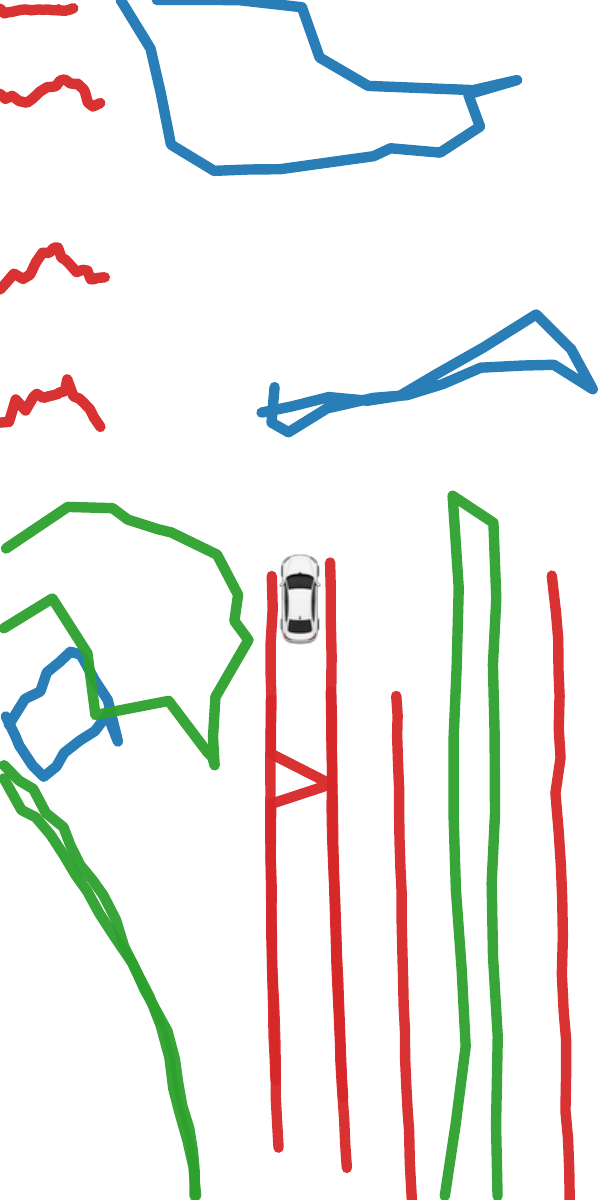} \\
\noalign{\vskip 2pt}
\rowlblsixF{100\,m$\times$50\,m} &
\bevcellsixF{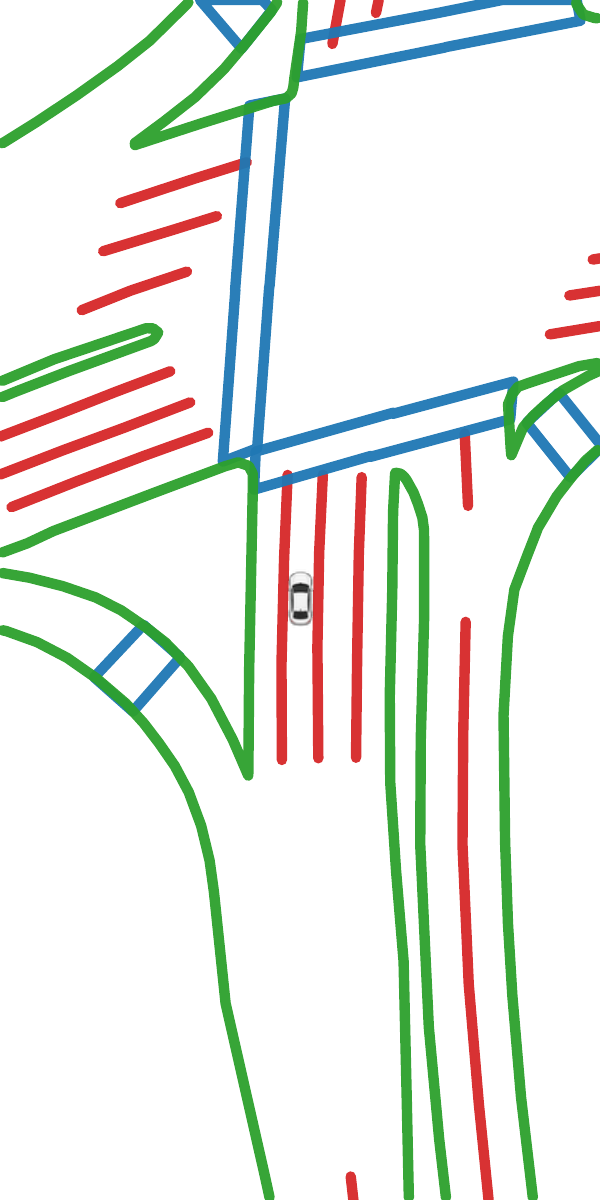} &
\bevcellsixF{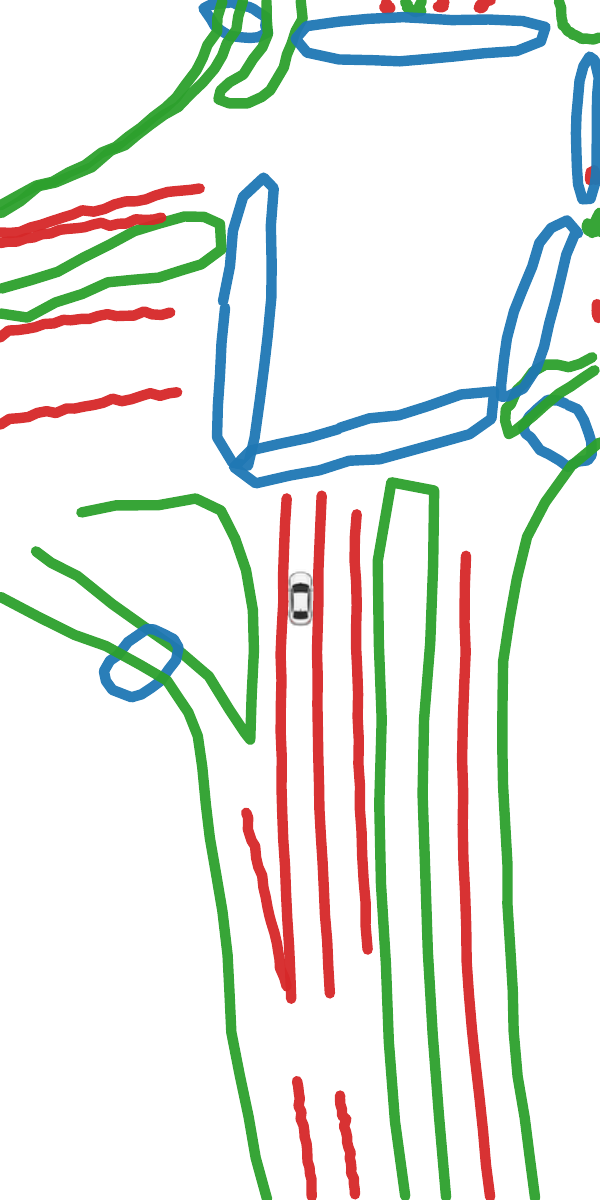} &
\bevcellsixF{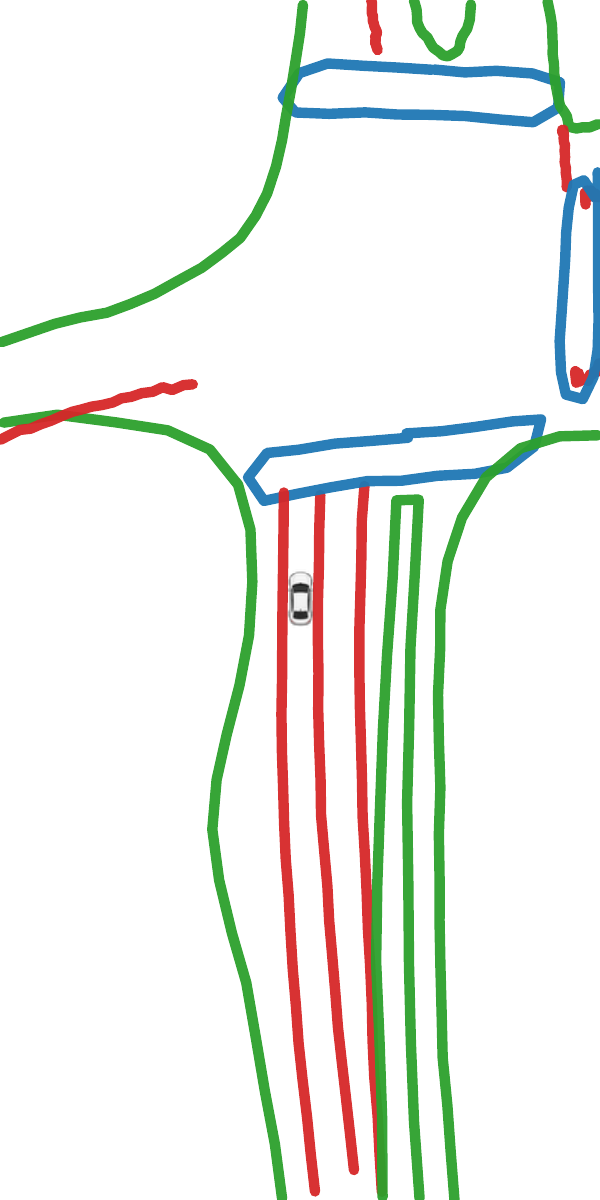} &
\bevcellsixF{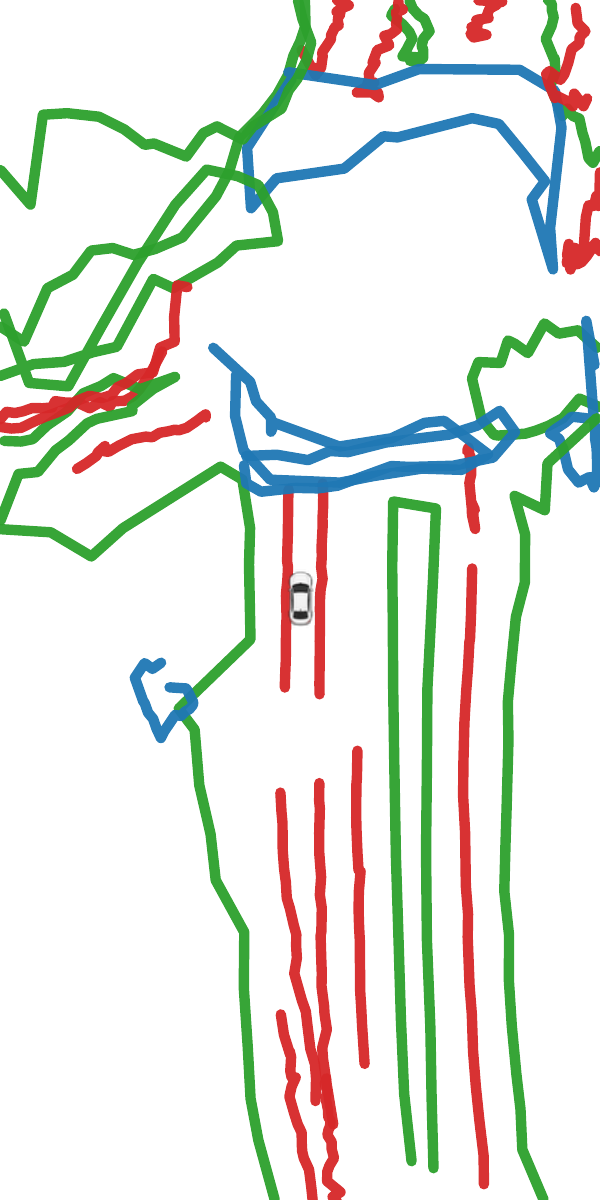} &
\bevcellsixF{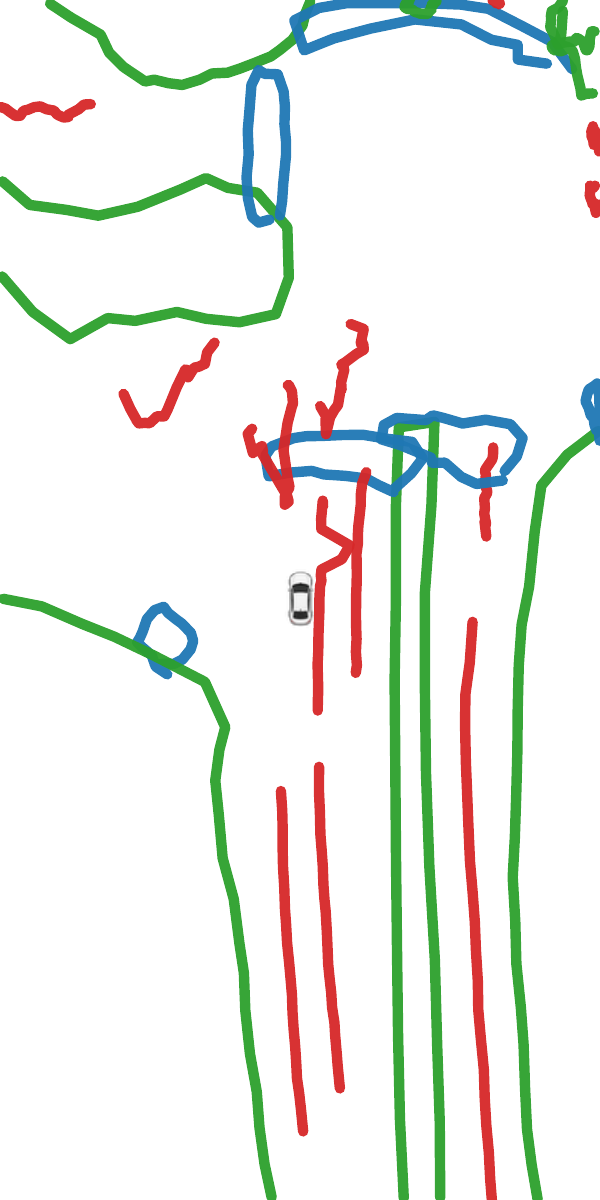} &
\bevcellsixF{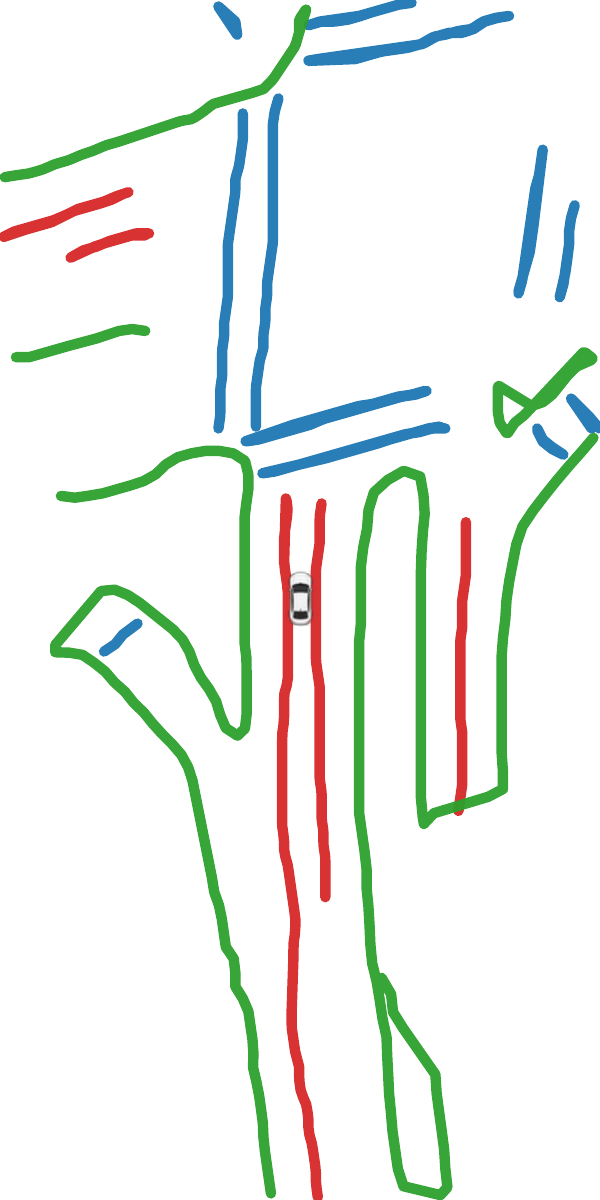} \\
\end{tabular}

\vspace{5pt}
{\footnotesize
\textcolor{clsped}{\rule[-1pt]{8pt}{8pt}}~ped.\,crossing\quad
\textcolor{clsdiv}{\rule[-1pt]{8pt}{8pt}}~divider\quad
\textcolor{clsbnd}{\rule[-1pt]{8pt}{8pt}}~boundary%
}

\caption{Additional qualitative comparison on singapore-queenstown.}
\label{fig:qual_appendix_scene0795_4690}
\end{figure*}

%% file: figures/qual_appendix_6frames/scene-0962_5338/qual_scene-0962_5338_6col.tex
\begin{figure*}[t]
\centering

\definecolor{clsped}{HTML}{1F77B4}
\definecolor{clsdiv}{HTML}{D62728}
\definecolor{clsbnd}{HTML}{2CA02C}

\newlength{\camhsixD}    \setlength{\camhsixD}{0.115\textwidth}
\newlength{\sathsixD}    \setlength{\sathsixD}{\dimexpr 2\camhsixD + 2pt\relax}
\newlength{\bevhsixD}    \setlength{\bevhsixD}{0.215\textwidth}
\newlength{\bevimgwsixD} \setlength{\bevimgwsixD}{0.105\textwidth}
\newlength{\bevcellwsixD}\setlength{\bevcellwsixD}{0.13\textwidth}

\setlength{\tabcolsep}{0pt}
\renewcommand{\arraystretch}{1.0}
\noindent
\begin{minipage}[c]{0.63\textwidth}
\centering
{\small\textsf{Multi-view Images}}\\[2pt]
\begin{tabular}{@{}c@{\hskip 2pt}c@{\hskip 2pt}c@{}}
\includegraphics[height=\camhsixD]{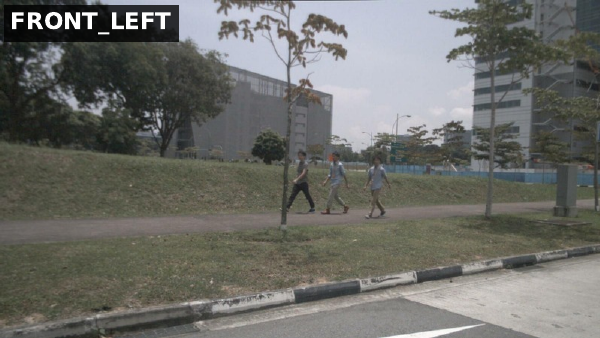} &
\includegraphics[height=\camhsixD]{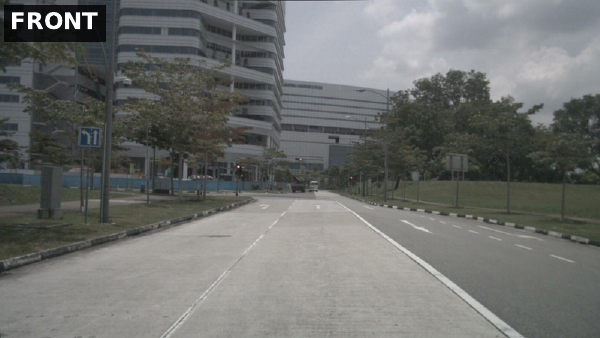} &
\includegraphics[height=\camhsixD]{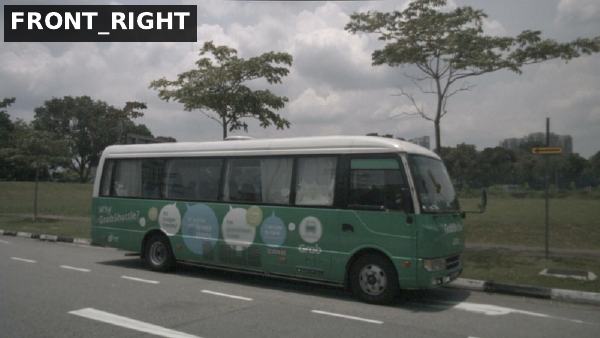} \\
\includegraphics[height=\camhsixD]{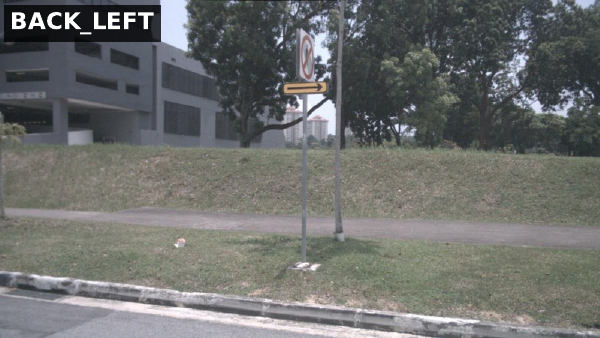} &
\includegraphics[height=\camhsixD]{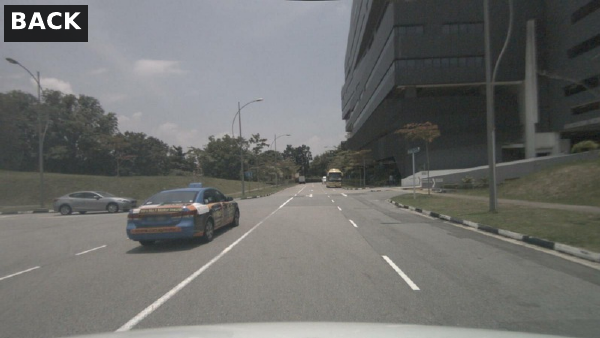} &
\includegraphics[height=\camhsixD]{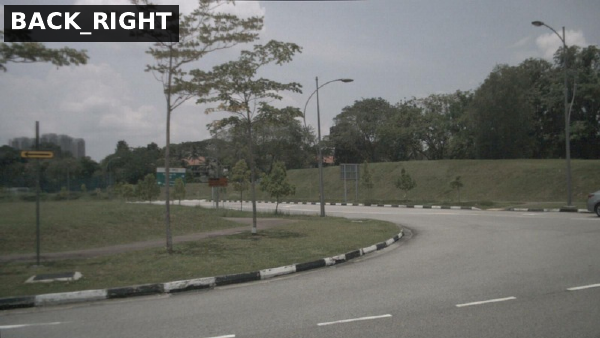} \\
\end{tabular}
\end{minipage}%
\hspace{4pt}%
\begin{minipage}[c]{0.25\textwidth}
\centering
{\small\textsf{Satellite Image}}\\[2pt]
\raisebox{4pt}{%
\mbox{%
\includegraphics[height=\sathsixD]{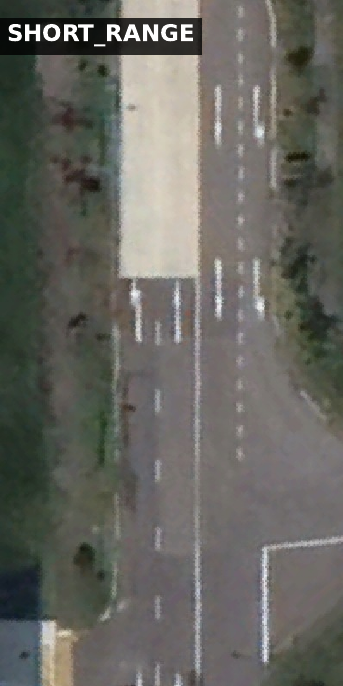}\hspace{3pt}%
\includegraphics[height=\sathsixD]{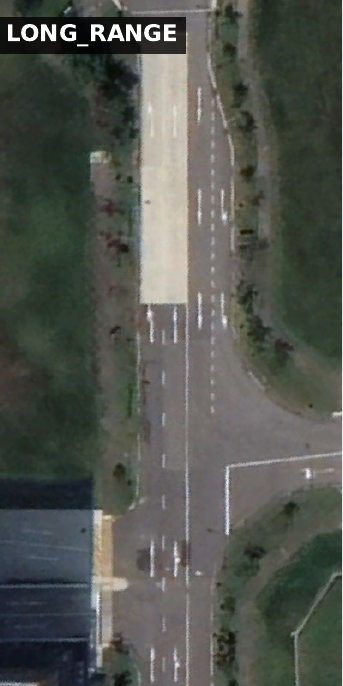}%
}
}
\end{minipage}

\vspace{8pt}

\newcommand{\rowlblsixD}[1]{\parbox[c][\bevhsixD][c]{4.5em}{\centering\footnotesize\mbox{#1}}}
\newcommand{\bevcellsixD}[1]{\parbox[c][\bevhsixD][c]{\bevcellwsixD}{\centering\fbox{\includegraphics[width=\bevimgwsixD]{#1}}}}

\setlength{\fboxsep}{0pt}
\setlength{\fboxrule}{0.4pt}
\setlength{\tabcolsep}{1pt}
\renewcommand{\arraystretch}{1.0}
\begin{tabular}{c@{\hskip 2pt}cccccc}
 & {\footnotesize\textsf{Ground Truth}} & {\footnotesize\textbf{\textsf{ReSMap(Ours)}}} & {\footnotesize\textsf{MapTracker}} & {\footnotesize\textsf{DAMap}} & {\footnotesize\textsf{SDTagNet}} & {\footnotesize\textsf{SatforHDMap}} \\
\rowlblsixD{60\,m$\times$30\,m} &
\bevcellsixD{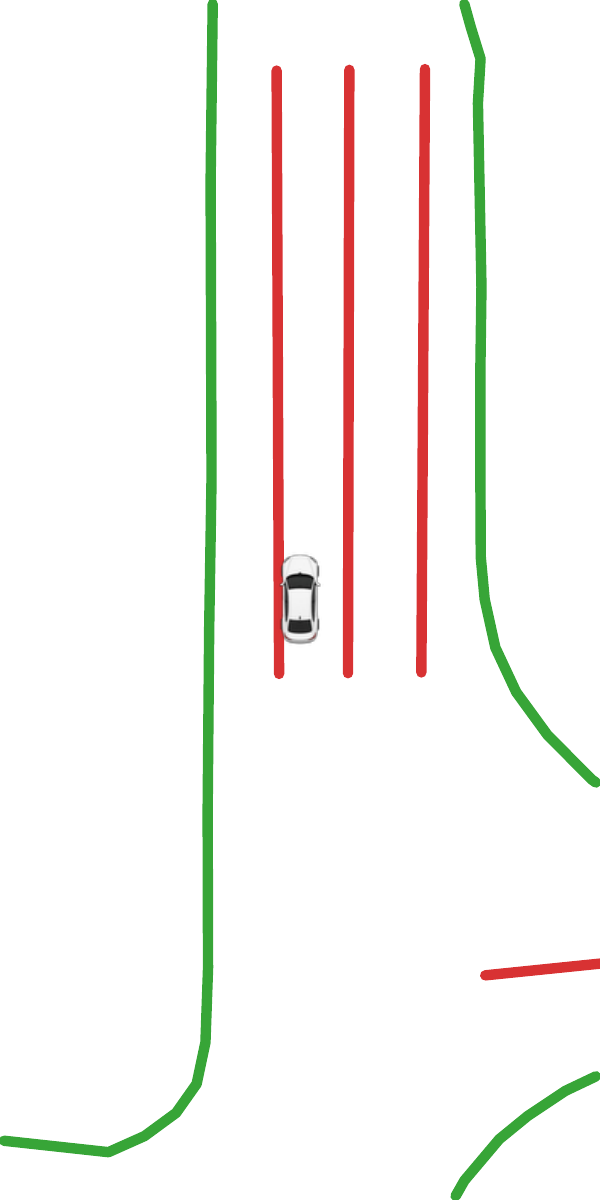} &
\bevcellsixD{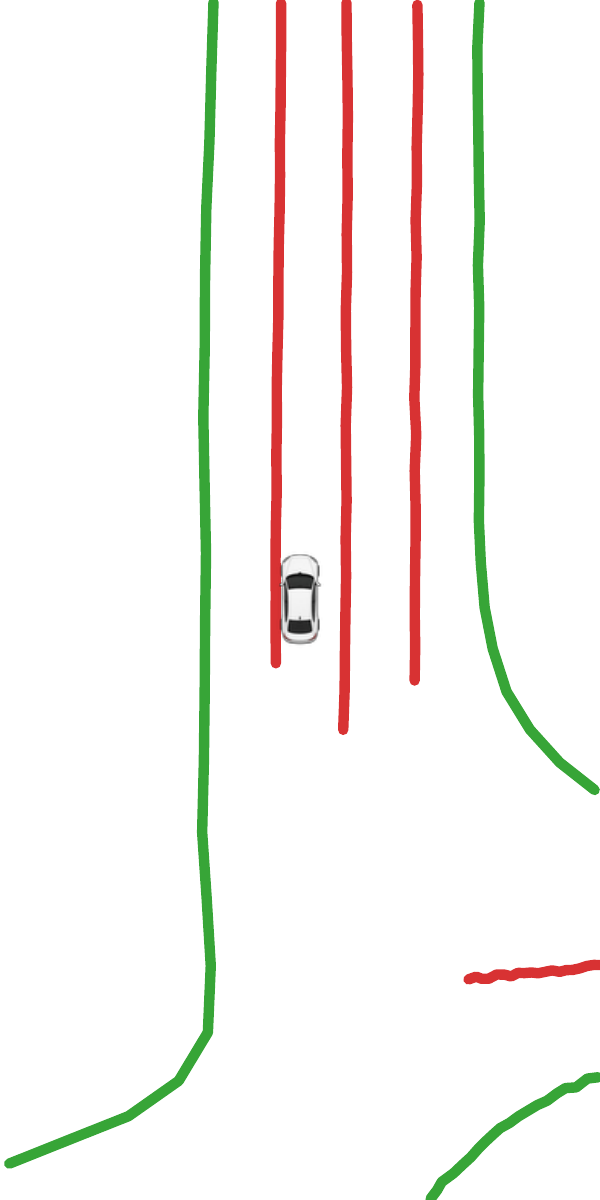} &
\bevcellsixD{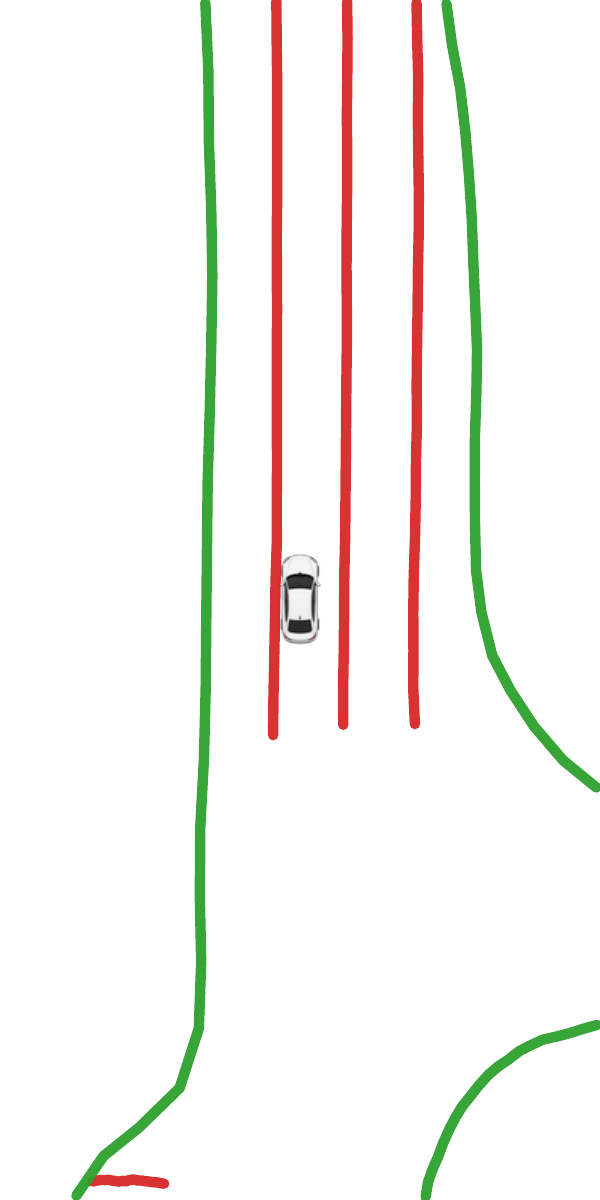} &
\bevcellsixD{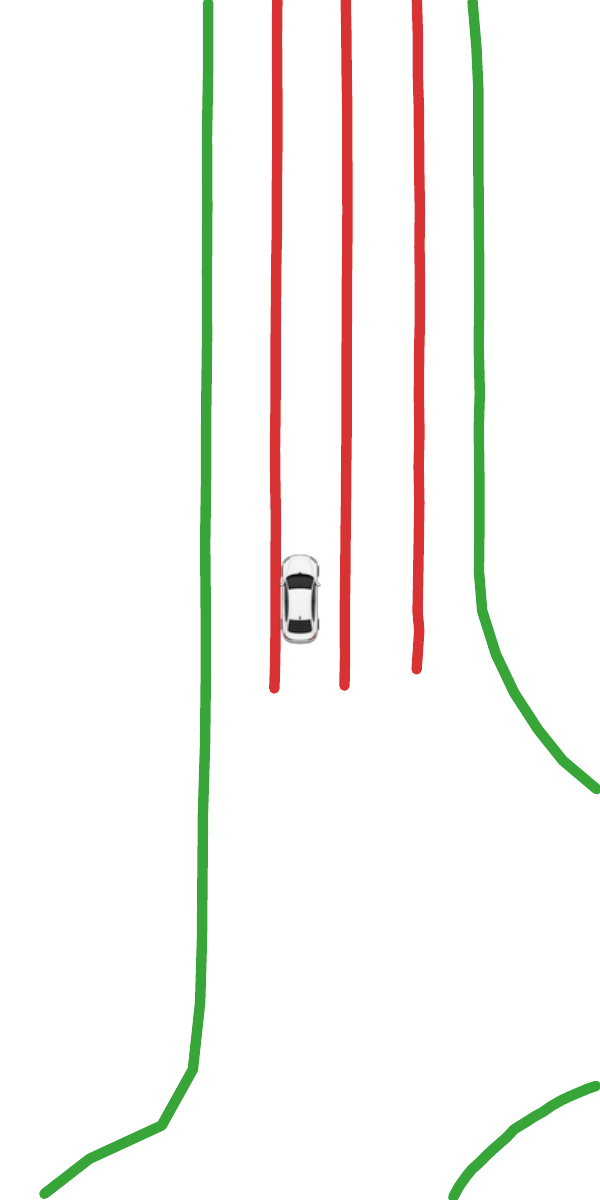} &
\bevcellsixD{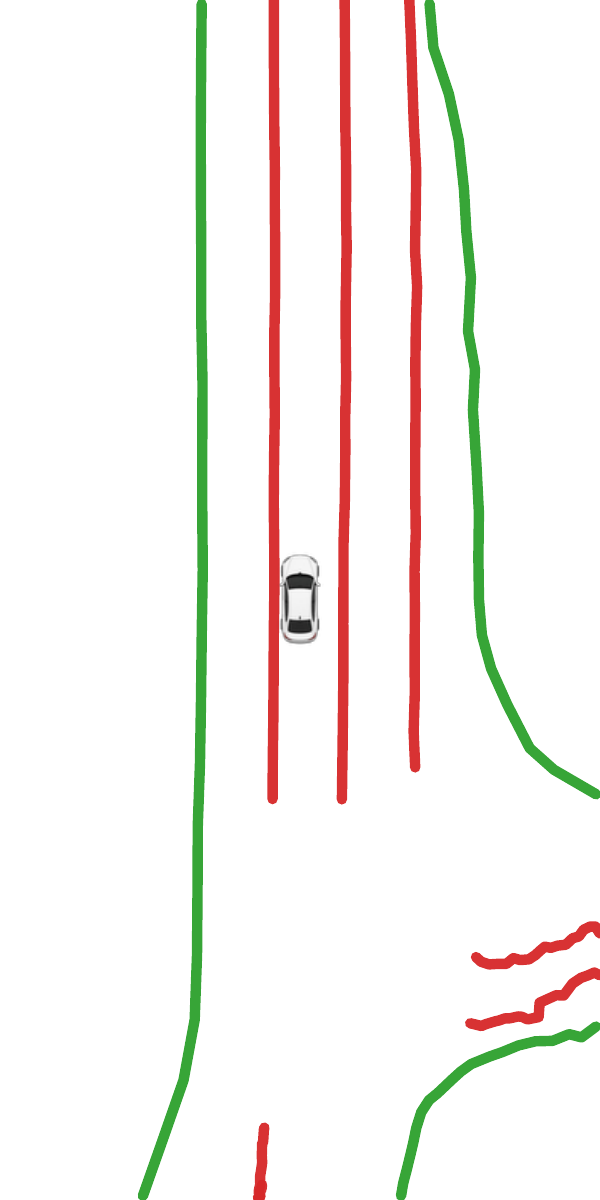} &
\bevcellsixD{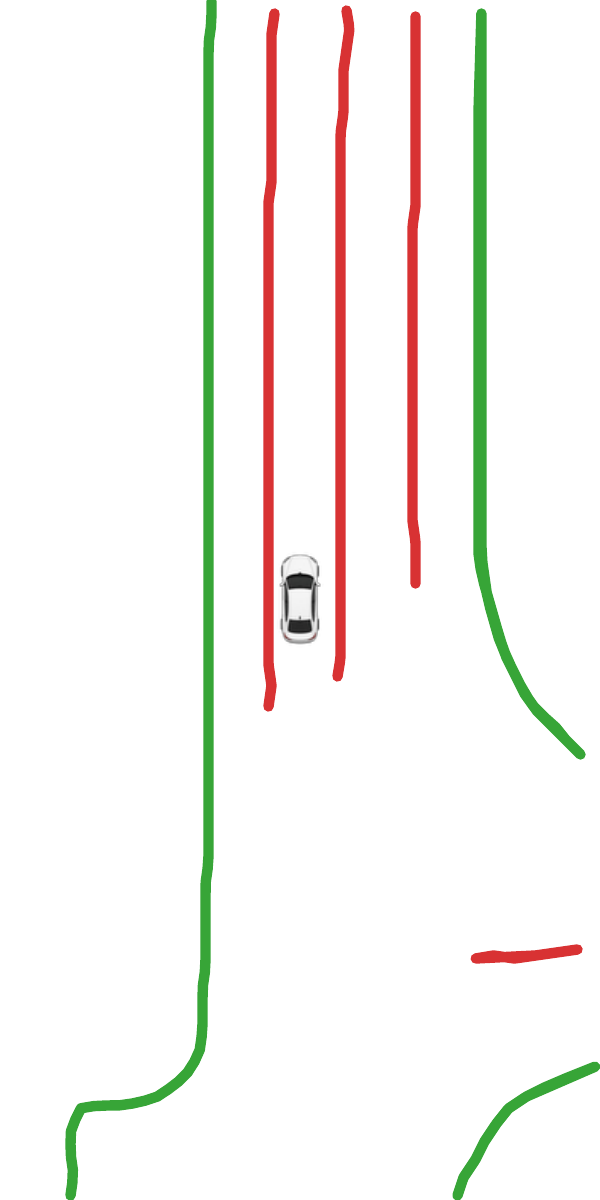} \\
\noalign{\vskip 2pt}
\rowlblsixD{Front-3 drop} &
\bevcellsixD{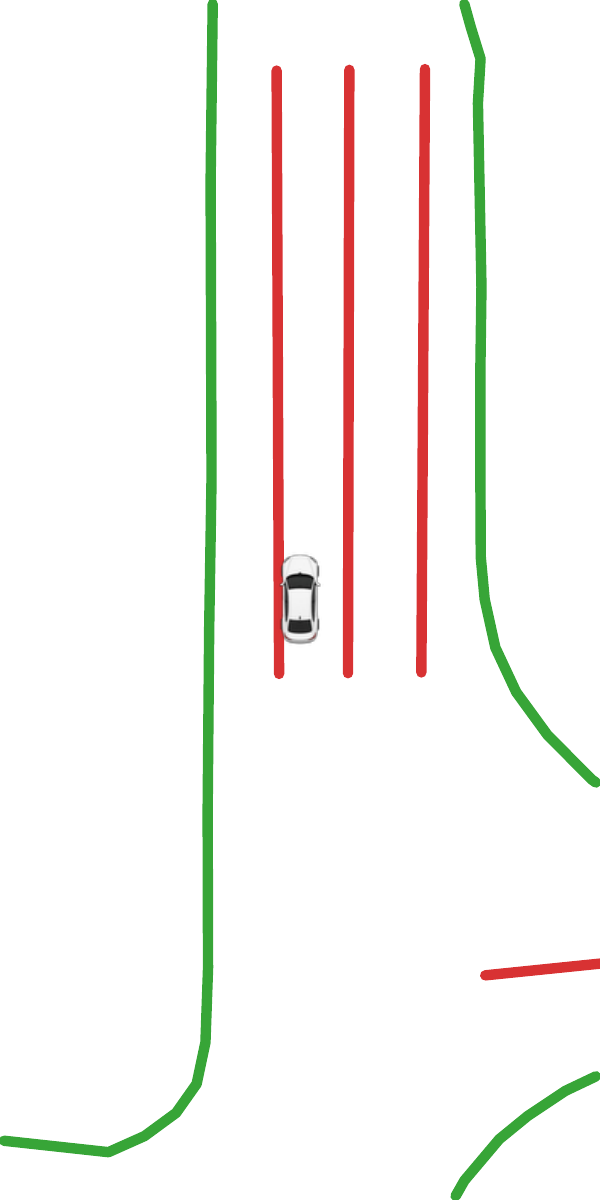} &
\bevcellsixD{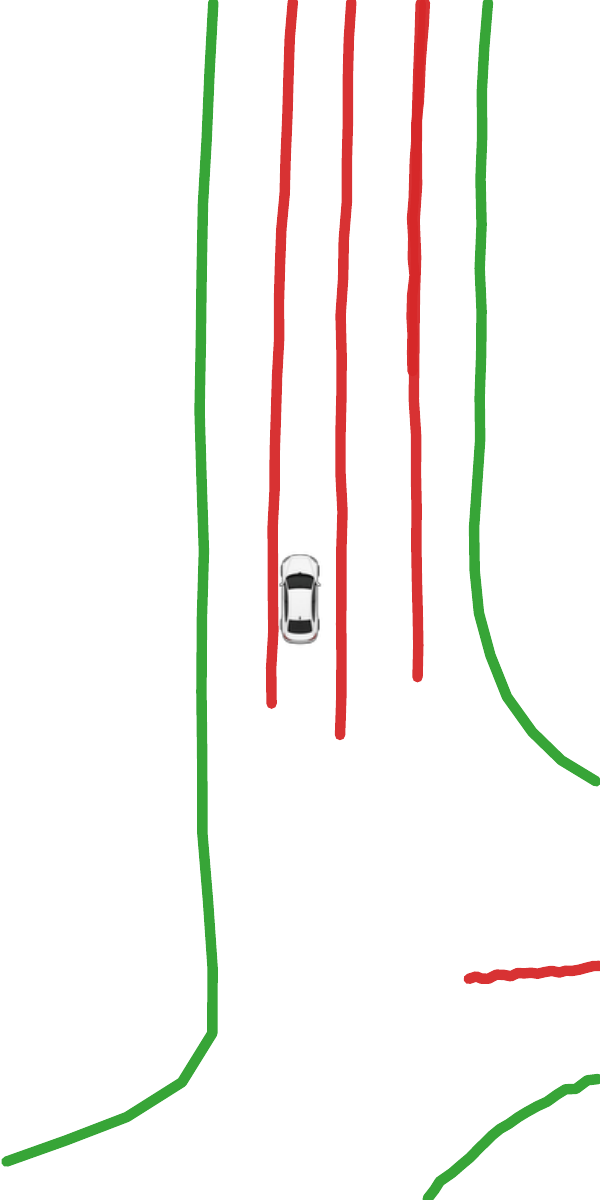} &
\bevcellsixD{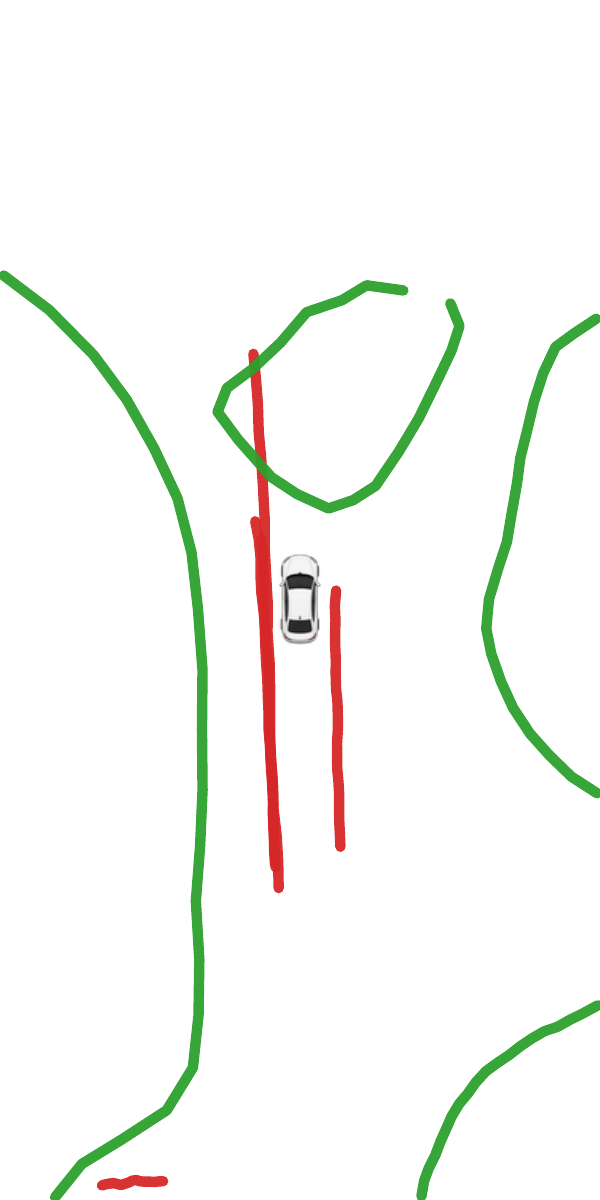} &
\bevcellsixD{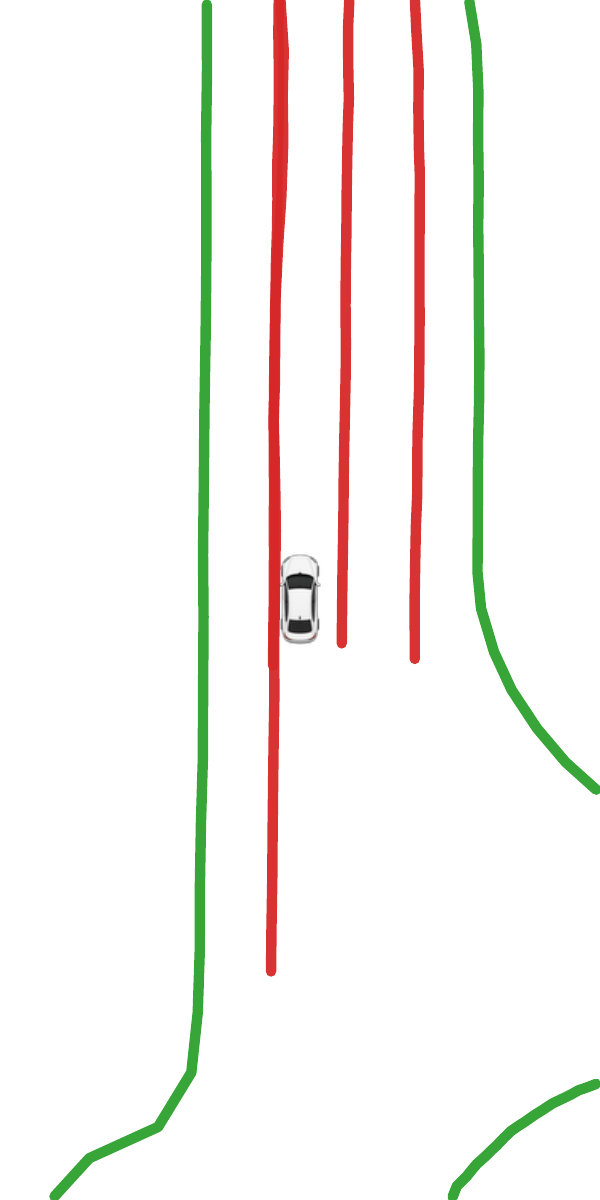} &
\bevcellsixD{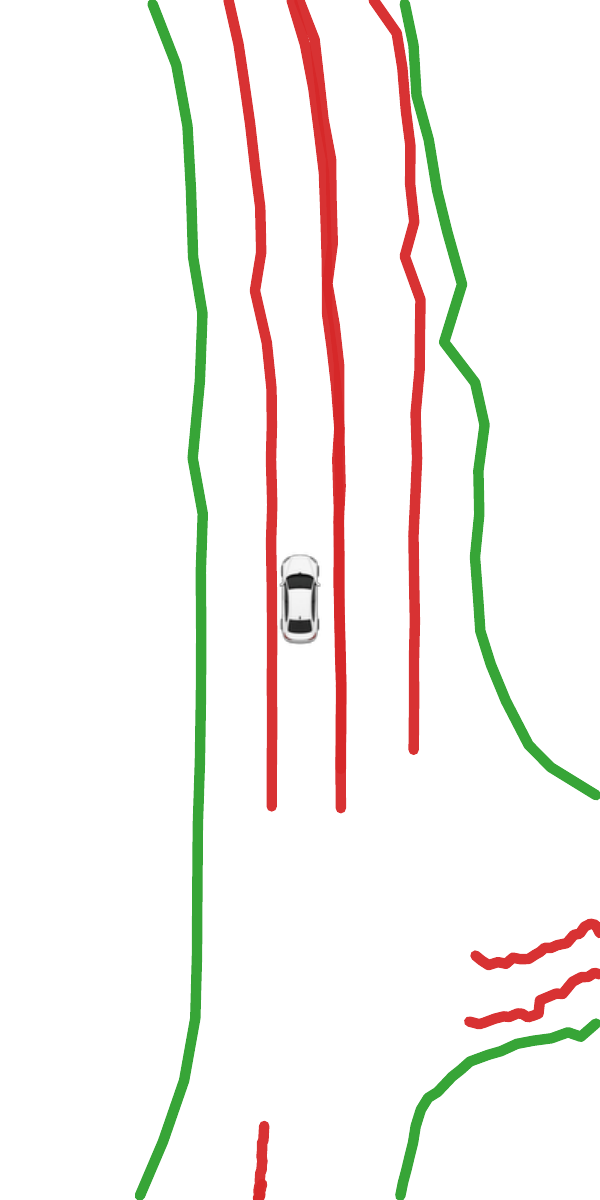} &
\bevcellsixD{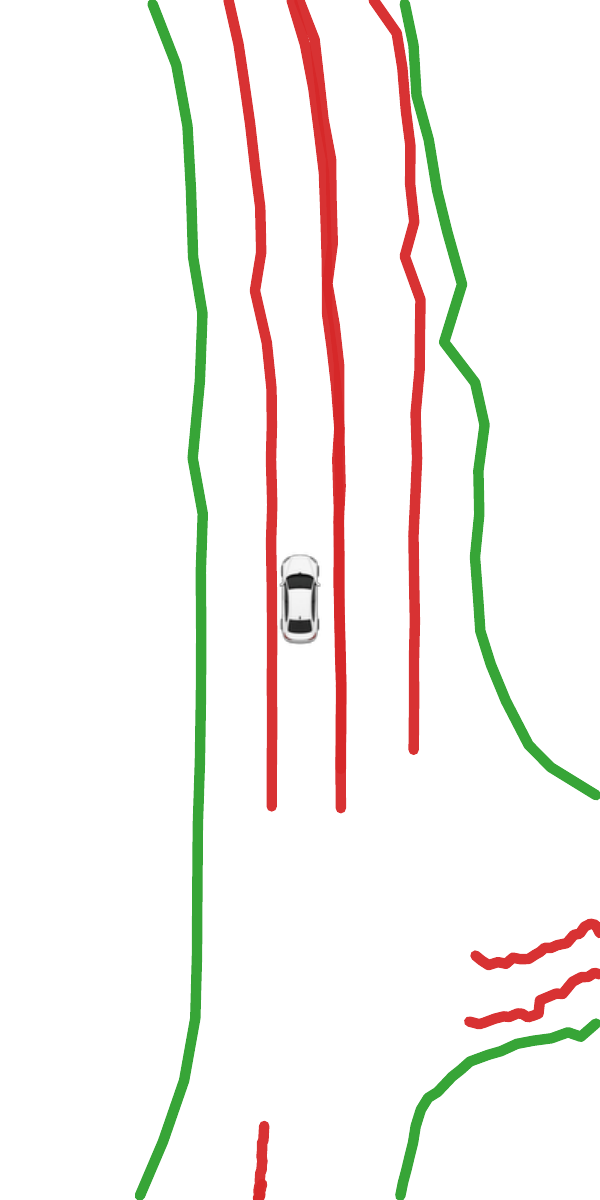} \\
\noalign{\vskip 2pt}
\rowlblsixD{100\,m$\times$50\,m} &
\bevcellsixD{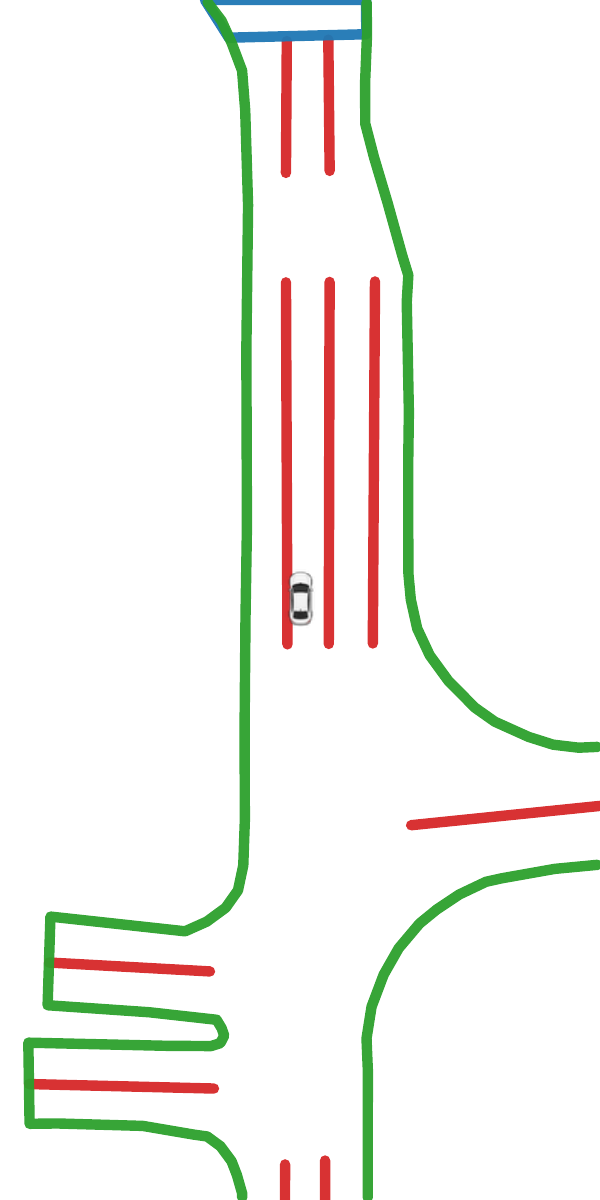} &
\bevcellsixD{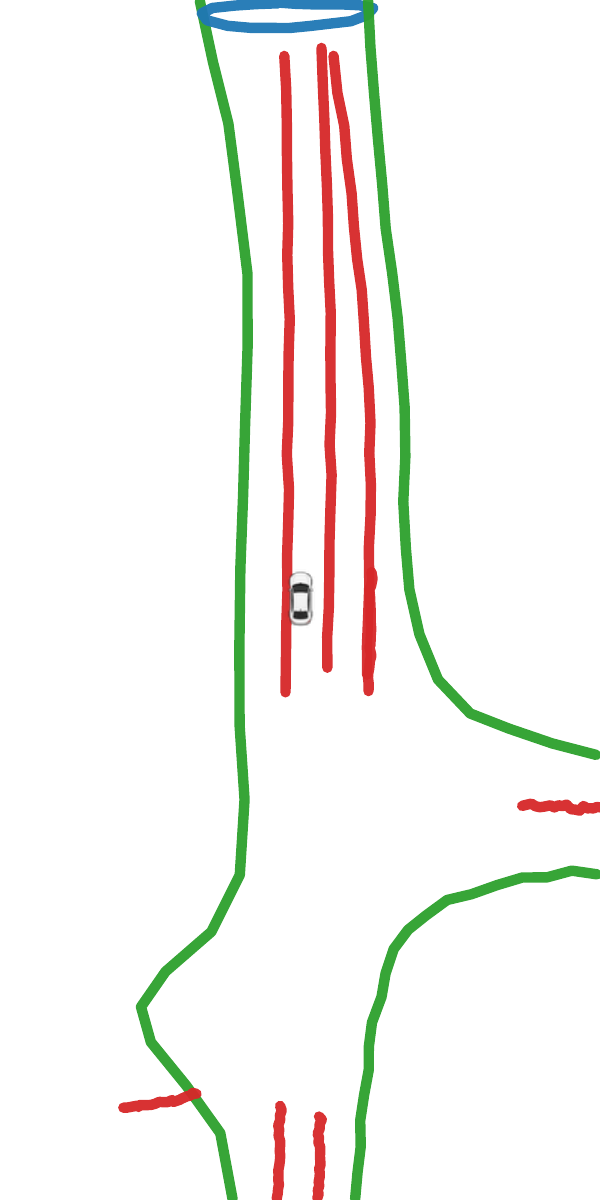} &
\bevcellsixD{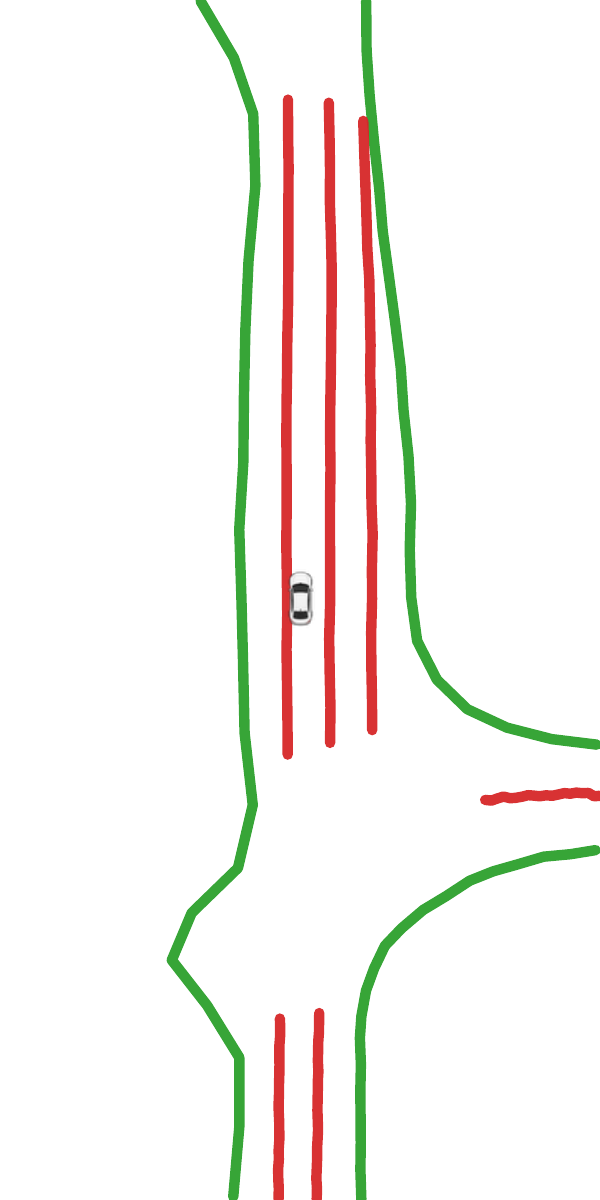} &
\bevcellsixD{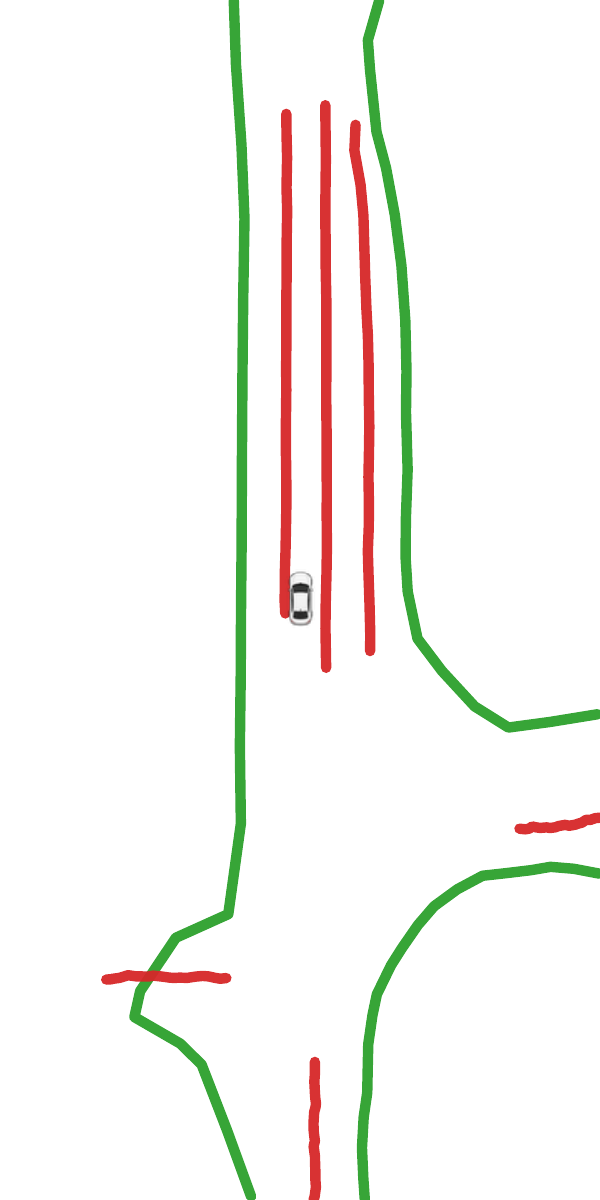} &
\bevcellsixD{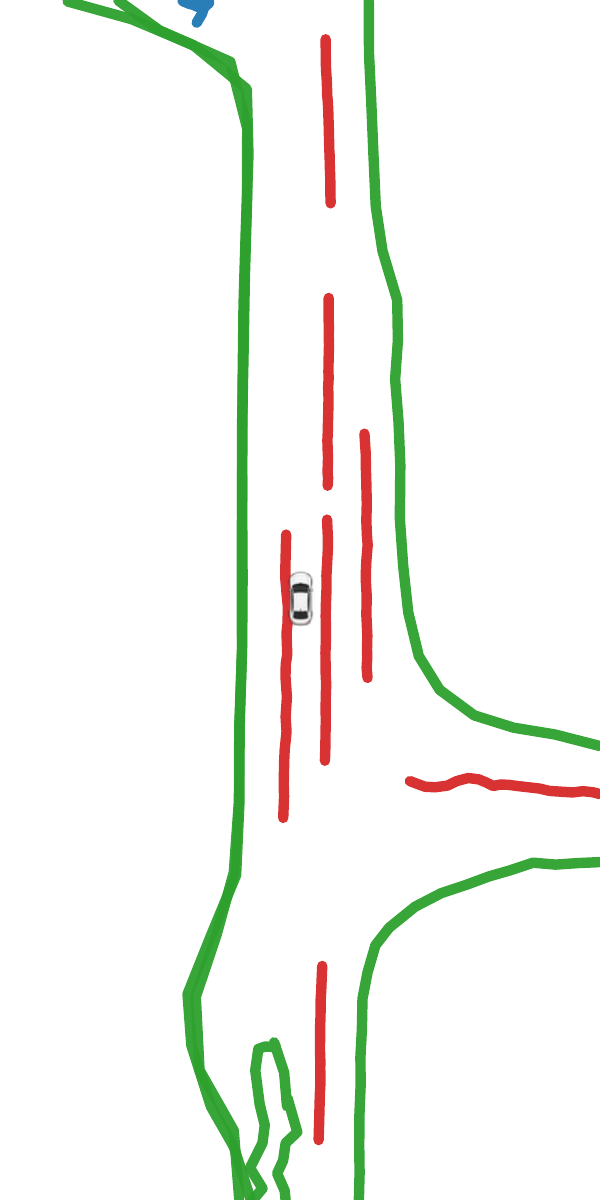} &
\bevcellsixD{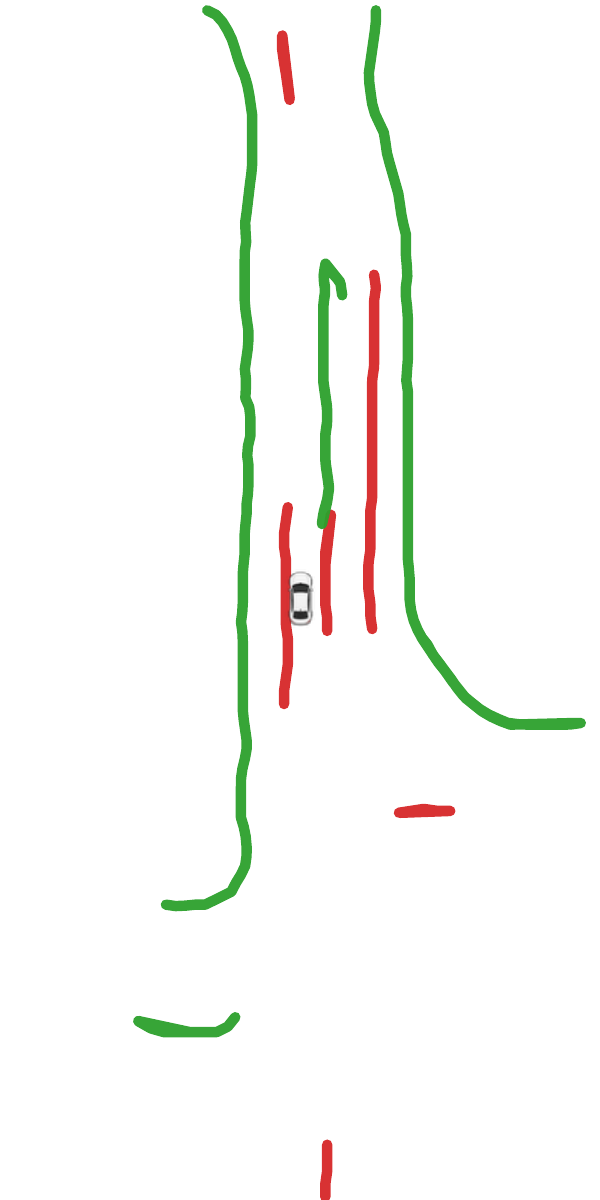} \\
\end{tabular}

\vspace{5pt}
{\footnotesize
\textcolor{clsped}{\rule[-1pt]{8pt}{8pt}}~ped.\,crossing\quad
\textcolor{clsdiv}{\rule[-1pt]{8pt}{8pt}}~divider\quad
\textcolor{clsbnd}{\rule[-1pt]{8pt}{8pt}}~boundary%
}

\caption{Additional qualitative comparison on singapore-onenorth. The extended
$100{\times}50\,\mathrm{m}$ ROI setting illustrates a representative failure case
in the lower-left region, where occlusion and visual ambiguity in the satellite
image caused by nearby buildings may lead to missed map elements.}
\label{fig:qual_appendix_scene0962_5338}
\end{figure*}